\documentclass[10pt,journal,cspaper,compsoc]{IEEEtran}
\usepackage{verbatim}
\usepackage{graphicx}
\usepackage{subfigure}
\usepackage{comment}
\usepackage{array}
\usepackage{mathtools}
\usepackage{amsmath}
\usepackage{amssymb}
\usepackage{color}
\usepackage{multirow}
\usepackage{balance}
\usepackage{threeparttable}
\usepackage{booktabs} % For formal tables
\usepackage{epsfig}
\usepackage{arydshln}
\usepackage{url}
\input{insbox}

\usepackage{epstopdf}
\makeatletter
\newif\if@restonecol
\makeatother

\usepackage{xcolor}
\newcommand{\hide}[1]{} %hide
\newcommand{\vpara}[1]{\vspace{0.05in}\noindent \textbf{#1 }}

\newcommand{\secref}[1]{Section~\ref{#1}} %section reference
\newcommand{\figref}[1]{Figure~\ref{#1}} %section reference
\newcommand{\beq}[1]{\vspace{-0.03in}\begin{equation}#1\end{equation}\vspace{-0.03in}}
\newcommand{\beqn}[1]{\vspace{-0.04in}\begin{eqnarray}#1\end{eqnarray}\vspace{-0.04in}}

\newcommand{\tabincell}[2]{\begin{tabular}{@{}#1@{}}#2\end{tabular}}

\ifCLASSOPTIONcompsoc
\else
\fi
\ifCLASSINFOpdf
\else
\fi
\begin{document}
\title{Neural, Symbolic and Neural-Symbolic Reasoning on Knowledge Graphs}

%\author{Anonymous Author*	}

\author{Jing~Zhang*\normalsize,
	Bo~Chen\normalsize,
	Lingxi Zhang\normalsize, 
	Xirui Ke\normalsize,
	Haipeng Ding\normalsize

	\IEEEcompsocitemizethanks{
		\IEEEcompsocthanksitem Jing Zhang, Bo Chen, Lingxi Zhang, Xirui Ke, Haipeng Ding are with Information School, Renmin University of China, Beijing, China. \protect\\
		E-mail:zhang-jing@ruc.edu.cn, allanchen224@gmail.com, zhanglingxi@ruc.edu.cn, 
		kexirui@ruc.edu.cn, 
		dinghaipeng@ruc.edu.cn, \protect\\
		*Corresponding author 
	}% <-this % stops a space
	\thanks{}}

\IEEEcompsoctitleabstractindextext{%
\begin{abstract}
Knowledge graph reasoning is the fundamental component to support machine learning applications such as information extraction, information retrieval, and recommendation. Since knowledge graphs can be viewed as the discrete symbolic representations of knowledge, reasoning on knowledge graphs can naturally leverage the symbolic techniques. However, symbolic reasoning is intolerant of the ambiguous and noisy data. On the contrary, the recent advances of deep learning promote neural reasoning on knowledge graphs, which is robust to the ambiguous and noisy data, but lacks interpretability compared to symbolic reasoning. Considering the advantages and disadvantages of both methodologies, recent efforts have been made on combining the two reasoning methods. In this survey, we take a thorough look at the development of the symbolic, neural and hybrid reasoning on knowledge graphs. We survey two specific reasoning tasks --- knowledge graph completion and question answering on knowledge graphs, and explain them in a unified reasoning framework. We also briefly discuss the future directions for knowledge graph reasoning.
\end{abstract}
% not want either math or citations to appear in the abstract.

% Note that keywords are not normally used for peer review papers.
}

\maketitle

% is not selected <OR> if conference mode is selected - because compsoc
% conference papers position the abstract like regular (non-compsoc)
\IEEEdisplaynotcompsoctitleabstractindextext

% \fi
\IEEEpeerreviewmaketitle

\section{Introduction}
\label{sec:intro}

\IEEEPARstart{S}{ymbolism} and connectionism are two main paradigms of Artificial Intelligence. Symbolism assumes the basic units which compose the human intelligence are symbols, and the cognitive process is a series of explicit inferences upon symbolic representations~\cite{minsky1969introduction,haugeland1989artificial}. Generally, the models of symbolism have sound readability and interpretability. However, the finite and discrete symbolic representations are insufficient to depict all the intrinsic relationships among data, and also intolerant of ambiguous  and noisy data.
On the contrary, connectionism imitates the process of neuron connections and cooperation in human brains to build models~\cite{rosenblatt1958perceptron,rumelhart1986learning}. Different models of connectionism have been developed. Deep learning is a representative model of connectionism~\cite{bengio2007greedy,hinton2006fast}. Deep learning has reached unprecedented impact across research communities as it achieved superior performances on many tasks in different fields such as image classification in computer vision~\cite{chen2020simple,he2020momentum,he2016deep}, language modeling in natural language processing~\cite{brown2020language,devlin2018bert}, and link prediction in networks~\cite{kipf2016semi,velivckovic2017graph,you2020graph}.This indicates that deep learning models are capable of modeling the implicit correlations inside data. 
However, the models of connectionism cannot provide explicit inference evidence to explain the results, making it look like a black box. 
%So, the concerns about interpretability and accountability of deep learning have been raised by influential thinkers. 
%Neural-symbolic models aim at integrating the strength of connectionism and symbolism in which the knowledge is represented in the symbolic form, whereas learning and reasoning are conducted by the neural network.
The problem of combining symbolism and connectionism has been researched since the 1980s~\cite{gallant1993neural,gallant1988connectionist,shavlik1991symbolic,towell1994refining,towell1994knowledge},
and has attracted much attention recent years~\cite{besold2017neural,de2011neural,garcez2019neural,garcez2012neural,lamb2020graph}.

\begin{figure}[t]
	\centering
	\includegraphics[width=0.48\textwidth]{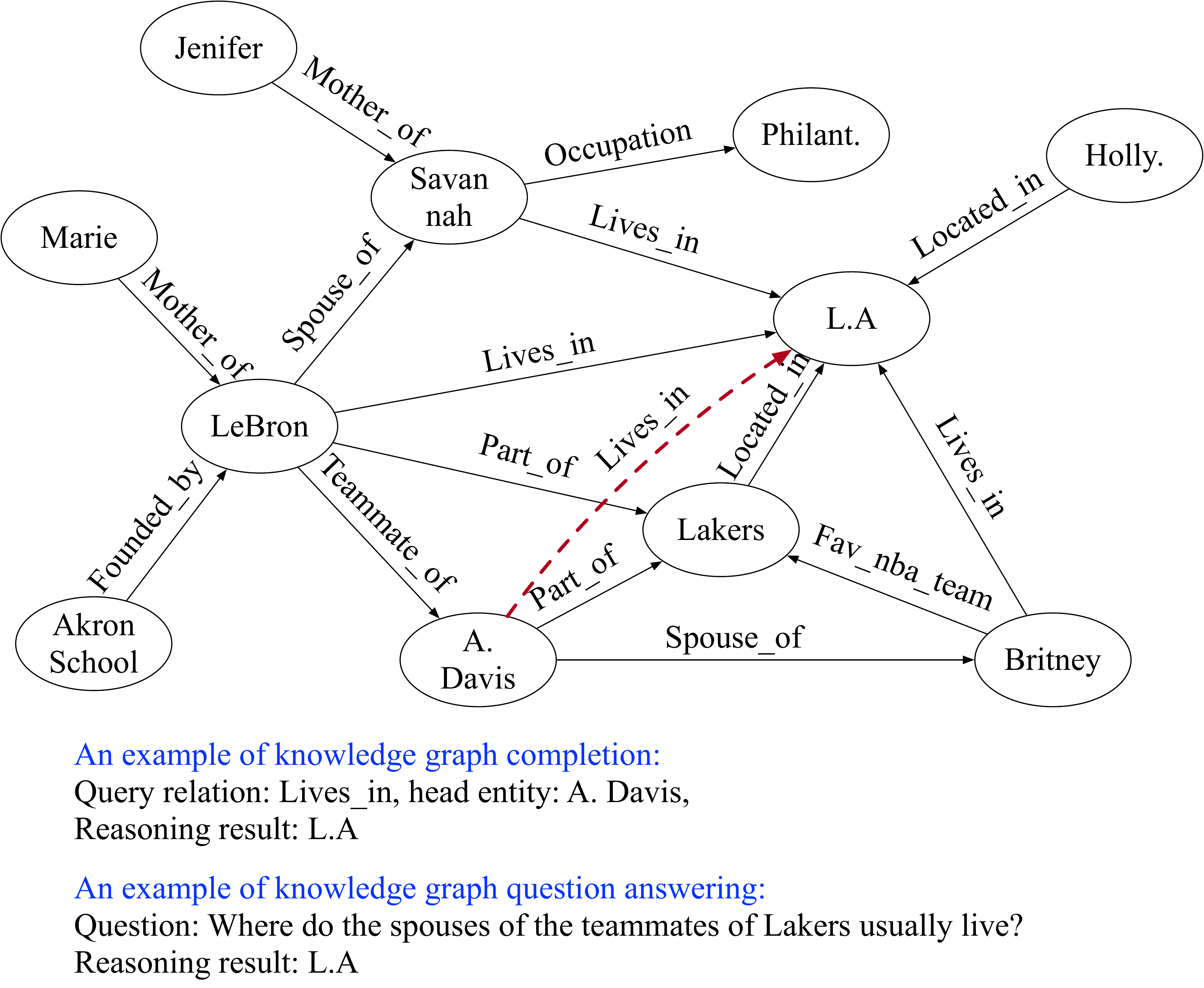}
	\caption{\label{fig:reasoning_example} Illustration of a knowledge graph~\cite{teru2019inductive} and examples of two reasoning tasks.}
\end{figure}

In this survey, we discuss how the above two principles are performed and intertwined for reasoning over knowledge graphs (KGs) such as Freebase~\cite{Bollacker2008}, DBpedia~\cite{Lehmann2015}, YAGO~\cite{Suchanek2007} and NELL~\cite{Carlson2010}, which are composed by a large number of (head, relation, and tail) triplets. Since KGs are naturally discrete symbolic representations to support explicit inferences and can be also represented into the continuous vector space by deep learning models such as the state-of-the-art TransE~\cite{bordes2013translating} to support implicit inferences, it is promising to effectively combine the ideas of symbolism and connectionsim on KG reasoning tasks.
Specifically, we review the mechanisms of two typical KG reasoning tasks --- knowledge graph completion (KGC) and knowledge graph question answering (KGQA).
KGC is to infer the answers, i.e., the tail entities given a head entity and a query relation~\cite{Das2017MINERVA,lao2010relational,Xiong2017deeppath,yang2017differentiable}. It is formulated as $(h, r, ?)$ where $h$ is the head entity, $r$ is the query relation, and $?$ is the tail entity that should satisfy $r$ with $h$. 
KGQA is similar to KGC except that the condition of the head entity $h$ and the query relation $r$ is replaced by a natural language question $q$~\cite{bao2016constraint,berant2013semantic,Saxena2020ImprovingMQ,sun-etal-2018-open}. 
Figure~\ref{fig:reasoning_example} illustrates the two reasoning tasks by examples, where the KGC example is to reason the tail entity ``L.A" given the query relation ``Lives\_in" and the head entity ``A. Davis", and the KGQA example is to reason the answer ``L.A"  given the question ``Where do the spouses of the teammates of Lakers usually live?".

Despite the abundant surveys about knowledge graph embedding~\cite{ji2020survey,rossi2020knowledge,Wang2017},  acquisition and applications~\cite{ji2020survey}, question answering~\cite{fu2020survey,Wu2019} and  reasoning~\cite{CHEN2020}, none of them explicitly sort out the symbolic or the neural methods, or the way to intertwine them. All the aforementioned surveys separate the reasoning tasks of KGC and KGQA. This paper unifies the two tasks in one reasoning framework and categorize the reasoning mechanisms into symbolic, neural, and hybrid. We demonstrate the pros. and cons. of each category and also discuss the future directions for KG reasoning. All the surveyed papers are retrieved by Google Scholar according to the keywords ``knowledge graph reasoning", ``knowledge graph completion", ``knowledge graph question answering", etc., and are chosen to be included if they are published by authoritative conferences or journals or  widely cited.

The rest of the article is organized as follows. \secref{sec:preliminaries} introduces the background knowledge that is closely related to the surveyed techniques. \secref{sec:reasoning_for_KGC} reviews the three kinds of reasoning techniques for knowledge graph completion. \secref{sec:reasoning_4_QA} reviews question answering on KGs under a similar reasoning framework. \secref{sec:con} first summarizes all kinds of reasoning methods in a unified technique development trend and then ends with the discussion of the potential directions in the future.

\section{Background Knowledge and Problem Definition}
\label{sec:preliminaries}
In this section, we first introduce the background information about knowledge graph, inductive logic programming, and Markov network, and then give the definition of knowledge graph reasoning.

\subsection{Knowledge Graphs (KGs)}
\label{sec:kgs}
We denote a knowledge graph $G$ as a set of facts, each represented by a triplet $(h,r,t )$, where $h$ is a head entity, $r$ is a relation and $t$ is a tail entity. In Figure~\ref{fig:reasoning_example}, (LeBron, part\_of Lakers) is an example of a triplet, where LeBron is the head entity, part\_of is the relation and Lakers is the tail entity.

\subsection{Inductive Logic Programming (ILP)}
\label{sec:ILP}

Inductive Logic Programming (ILP) aims at seeking underlying patterns formulated by logic programs/rules/formulas shared in the data. It is one of the rule-based learning methods which derive a set of if-then logic rules to describe the positive instances but not the negative instances. Most ILP works constrain the logic rule to be Horn clause/rule. A Horn rule consists of a head and a body, where the head is a single atom and the body is flat conjunction over several atoms. A Horn rule $\gamma$ can be formulated as:

\beq{
	\label{eq:rule}
	\gamma:A(\alpha_1,\cdots, \alpha_m) \rightarrow \alpha,
}

\noindent where $\alpha$ is called head atom and $\alpha_1$, $\cdots$, $\alpha_m$ ($m \geq 0 $) are body atoms. $A$ is the rule body which is usually defined as a conjunction normal form (CNF) that uses logical operations $\{\land, \lor, \neg \}$ to combine the body atoms together. 
The rule body can also be referred to as a formula. 
If the rule body $A$ on the left is true, then the head atom on the right is also true. An atom $\alpha$ is defined as a predicate symbol $P_i$ that acts as a function to map the set of variables $\{ x_1, x_2, \cdots, x_n \}$ to true or false:

\beq{
	\label{eq:atom}
	\alpha \equiv P_i (x_1, x_2,\cdots, x_n ).
}

\noindent Although there can be multiple variables in a predicate, we usually only consider the simple unary and binary predicates, i.e., $n=1$ and $n=2$. For example, $\text{Person}$ is a unary predicate only applied to a single variable. The atom $\text{Person}(x)$ is true if $x$ is a person. $\text{Mother}$ is a binary predicate applied to two variables. The atom $\text{Mother}(x, y)$ is true if $x$ is the mother of $y$. When all the variables in an atom $\alpha$ are instantiated by constants, $\alpha$ is called a ground atom. 

In a knowledge graph, a relation $r$ can be viewed as a binary predicate; that is, $r(x,y)$ is an atom with two arguments $x$ and $y$. A triplet $(h,r,t) \in G$ can be taken as a ground atom $r(h,t)$ which applies a relation $r$ to a pair of entities $h$ and $t$. 

Given a set of pre-defined predicates $\mathcal{P}$, ground atoms $\mathcal{G}$, positive instances $\mathcal{S}$ and negative instances $\mathcal{N}$, ILP aims at constructing a set of rules to explain the positive instances and reject the negative instances. We take the example of learning which natural numbers are even from \cite{evans2018learning} to explain ILP. The predicate set is defined as:

\beq{
	\mathcal{P} = \{\text{zero}, \text{succ}\}, \nonumber
}

\noindent where $\text{zero}(X)$ is an unary predicate which is true if $X$ is 0, $\text{succ}(X, Y)$ is a binary predicate which is true if $X$ is the successor of $Y$. The ground atoms are:

\beq{
	\mathcal{G} = \{\text{zero}(0),\text{succ}(0,1),\text{succ}(1,2), \cdots\}. \nonumber
} 

The positive and negative instances of the even predicate are:

\beqn{\nonumber
	\mathcal{S} &=& \{\text{even}(0), \text{even}(2), \text{even}(4), \cdots\}, \\ \nonumber
	\mathcal{N} &=& \{\text{even}(1), \text{even}(3), \text{even}(5), \cdots\}. \nonumber
}

The solution of rules for the even predicate is:

\beqn{\nonumber
	\text{even}(X) &\leftarrow& \text{zero}(X), \\\nonumber
	\text{even}(X) &\leftarrow& \text{even}(Y) \land  \text{succ2}(Y,X), \\\nonumber
	\text{succ2}(X,Y) &\leftarrow& \text{succ}(X,Z) \land \text{succ}(Z,Y). \nonumber
}

\noindent where $\text{succ2}(X,Y)$ is an auxiliary predicate which is true when $X$ is the two-hop successor of $Y$. We can see that when the above rules are applied deductively to the ground atoms $\mathcal{G}$, they can produce $\mathcal{S}$ but not $\mathcal{N}$.

Two kinds of approaches are usually used to derive rules, where the top-down approach begins from general rules and adds new atoms to improve the coverage precision of positive instances~\cite{clark1989cn2,michalski1986multi}, while the bottom-up approach begins from the specific rules and deletes atoms to extend the coverage rate of the rules~\cite{plotkin1970note,plotkin1971further}. 

\subsection{Markov Network}
Markov network, also known as Markov random field, tries to model the knowledge graph using the joint distribution of a set of variables $X=(X_1, X_2, ..., X_n)$~\cite{pearl2014probabilistic}. Markov network is an undirected graph where each node represents a variable. For cliques in the graph, a nonnegative real-valued potential function $\phi_c$ is defined and the joint distribution is represented as:

\beq{
	P( X=x) = \frac{1}{Z} \prod_{c}\phi_c( x_{c})
}

\noindent where $x_c$ reflects the state of the variables appearing in the $c$-th clique. $Z = \sum_{x \in \mathcal{X}}\prod_{c}\phi_c(x_c)$ is the partition function for normalization. Markov networks are usually represented as log-linear models, with each clique potential replaced by an exponentiated weighted  sum of  the state's feature:

\beq{
	P(X=x) = \frac{1}{Z}\exp(\sum_{c} w_{c}f_{c}(x_c))
}

\noindent where $f_c$ is the feature function defined on the clique $c$ and $w_c$ is the corresponding vector of weights.

\section{Reasoning for KGC}
\label{sec:reasoning_for_KGC}
This section introduces the reasoning methods for KGC in three main categories. Figure~\ref{fig:kgc-framework} presents the taxonomy of the KGC reasoning methods.

\begin{table*}[t]
	\newcolumntype{?}{!{ \hspace{0.5pt} \vrule width 1pt \hspace{-2pt}}}
	\newcolumntype{k}{!{\hspace{-5pt}}}
	\newcolumntype{C}{>{\centering\arraybackslash}p{2em}}
	\centering \scriptsize
	\renewcommand\arraystretch{1.0}
	
	\caption{
		\label{tb:overall_KGC} A summary of knowledge graph completion and recent advances.	
	}

	\begin{tabular}{cclll}
		
		\toprule
		
		Category & Sub-category &  Model  &       Mechanism \\
		\midrule
		
		\multirow{25}{*}{\tabincell{c}{Neural \\ Reasoning}} 
		& \multirow{7}{*}{\tabincell{c}{Translation\\-based \\ Models}} 
		&  TransE~\cite{bordes2013translating}  &   force $\textbf{h}+\textbf{r}$ to be close to $\textbf{t}$ \\
		&&  TransH~\cite{Wang14}  &   project entities into the relation-specific hyperplane \\
		&&  TransR~\cite{lin2015learning} & project entities and relations in separate spaces\\ 
		&&  TransD~\cite{ji-2015}  & determine the project matrices by both entities and relations  \\
		&&  TransG~\cite{xiao2015transg}  & project a relation into multiple embeddings  \\
		&&  TransAt~\cite{qian2018translating}  & project a relation by entities' categories and the relation's attributes  \\
		&&  RotatE~\cite{sun2018rotate}  & map entities and relations to the complex vector space  \\
		 
		\cmidrule{2-4} 
		& \multirow{7}{*}{\tabincell{c}{Multiplicative \\Models}} 
		&  RESCAL~\cite{NickelICML11}  &  maximize the tensor product between two entities\\
		&& DisMult~\cite{yang2014embedding} & simplify the project matrix into a diagonal matrix\\ 
		&& ComplEx~\cite{pmlr-v48-trouillon16}  & use complex embeddings to handle asymmetry\\
		&& HolE~\cite{NickelAAAI16}  & use the circular correlation to represent entity pairs\\
		&& SimplE~\cite{NIPS2018_7682}  & learn a head and tail embeddings for each entity\\
		&& ANALOGY~\cite{pmlr-v70-liu17d} & constrain relation matrixes to be  normal matrixes \\
		&& DiHEdral~\cite{xu2019relation} & support the non-commutative property of a relation \\
		
		\cmidrule{2-4} 
		& \multirow{11}{*}{\tabincell{c}{Deep Learning \\ Models}} 
		&  ConvE~\cite{dettmers2018convolutional}   &   use the head and relation embeddings as the input of CNN \\
		&&  ConvR~\cite{jiang2019adaptive}  &   extend global filters in ConvE to relation-specific filters \\
		%&&%InteractE~\cite{vashishth2020interacte} & add more convolution operations\\ 
		&&  ConvKB~\cite{nguyen2018novel}  & use the head, relation and tail together as the input of CNN   \\
		&&  CapsE~\cite{Dai-2019}  & use the capsule network as the convolution function  \\
		&&  RSN~\cite{pmlr-v97-guo19c}  & use RNNs to capture long-term relational dependencies \\
		&&  R-GCN~\cite{schlichtkrull2018modeling}  & use relation-aware GCN as the encoder and DisMult as the decoder\\
		%&& SACN~\cite{ShangAAAI2019}     &  use relation-aware GCN as the encoder and ConvE  as the decoder \\
		&& M-GNN~\cite{Wang-ijcai2019}     &  replace the mean aggregator in R-GCN with a MLP \\
		&& CompGCN~\cite{vashishth2019composition}     &  incorporate entity and relation embeddings into the aggregator function\\
		&& KBGAT~\cite{Deepak-ACL2019}     &  incorporate a triplet into the aggregator function\\

		\midrule
		\multirow{4}{*}{\tabincell{c}{Symbolic \\ Reasoning}} 
		& \multirow{4}{*}{\tabincell{c}{ - }} 
		&  AIME~\cite{galarraga2013amie}  &   generate rules by rule extending and rule pruning\\
		&& AIME+~\cite{galarraga2015fast}  &   improve the efficiency of AMIE\\
		&&  RLvLR~\cite{omran2018scalable}  &   reduce the search space by the embedding technique\\
		&&  RuLES~\cite{ho2018rule}  & leverage the embedding technique to measure the quality of the rules  \\
		
		\midrule 
		\multirow{25}{*}{\tabincell{c}{Neural-Symbolic \\ Reasoning}} 
		& \multirow{4}{*}{\tabincell{c}{Symbolic-driven\\ Neural \\Reasoning}} 
		&  KALE~\cite{guo2016kale}   &   learn embeddings on the observed triples and the ground rules \\
		&&  RUGE~\cite{guo2017ruge}  &    change the ground rules to the triplets derived by rules\\
		&&  Wang et al.~\cite{wang2019logic} & transform a triple/ground rule into first-order logic\\ 
		&&  IterE~\cite{zhang2019iere}  & infer rules and update embeddings iteratively  \\

		\cmidrule{2-4} 
		& \multirow{5}{*}{\tabincell{c}{Symbolic-driven\\ Probabilistic\\ Reasoning}} 
		&  MLN~\cite{richardson2006markov}   &   build a probabilistic graphic model for all the rules and learn their weights\\
		&&  pLogicNet~\cite{qu2019policnet}  &  incorporate the embedding technique to infer marginal probabilities in MLN\\
		&&  ProbLog~\cite{de2007problog} & build a local SLD-tree for a relation and learn rules that can support it \\ 
		&&  SLPs~\cite{cussens2001parameter}  & define a randomized procedure for traversing the SLD-tree  in ProbLog \\
		&&  ProPPR~\cite{wang2013programming}  & change the randomized procedure in SLPs to a biased sampling strategy \\
		
		\cmidrule{2-4} 
		& \multirow{16}{*}{\tabincell{c}{Neural-driven\\ Symbolic\\ Reasoning}} 
		&  PRA~\cite{lao2010relational}   &   enumerate the paths between two entities \\
		&&  Lao et al.~\cite{lao-etal-2012-reading}  &  perform PRA on the KG and the extended textural graph\\
		&&  Neelakantan et al.~\cite{Neelakantan2015Compositional} & use RNN to encode the most confident path \\ 
		&&  Chain-of-Reasoning~\cite{Das2017chain}  & change the most confident path to multiple paths \\
		&&  DeepPath~\cite{Xiong2017deeppath}  & use RL to evaluate the sampled paths \\
		&&  AnyBURL~\cite{meilicke2020AnyBURL}  & generalize the sampled paths to abstract rules \\
		&&  MINERVA~\cite{Das2017MINERVA}  & use RL to directly find the answer \\
		&&  MultiHop~\cite{Lin2018multi}  & adopt soft reward and action dropout for MINERVA\\
		&&  CPL~\cite{fu2019CPL}  & leverage the text in addition to the KG when sampling\\
		&&  M-walk~\cite{shen2018mwalk}  & use Monte Carlo Tree Search when sampling\\
		&&  DIVA~\cite{chen2018diva}  & use VAE to unify path sampling and answer reasoning\\
		&&  CogGraph~\cite{Du-2019}  & sample multiple entities at each hop\\
		&&  TensorLog~\cite{cohen2016tensorlog}  & keep all the neighbors at each hop without sampling\\
		&&  Neural LP~\cite{yang2017differentiable}  & learn new rules based on TensorLog\\
		&&  NLIL~\cite{yang2019learn}  & deal with non-chain-like rules\\
		&&  Neural-Num-LP~\cite{wang2019differentiable}  & deal with  numerical operations\\

		\bottomrule
	\end{tabular}
	\normalsize

\end{table*}

\subsection{Neural Reasoning}
\label{sec:neural_reasoning_for_KGC}
Neural reasoning, also known as knowledge graph embedding, aims at learning the distributed embeddings for entities and relations in KGs and inferring the answer entities based on embeddings when given the head entity and relation. Generally, existing neural reasoning methods can be categorized into translation-based models, multiplicative models,  and deep learning models.

\subsubsection{Translation-based Models}
Translation-based models usually learn embeddings by translating a head entity to a tail entity through the relation. For example,  TransE~\cite{bordes2013translating}, a representative translation-based model, maps the entities and relations into the same vector space and forces the added embedding $\textbf{h}+\textbf{r}$ of a head entity $h$ and a relation $r$  to be close to the embedding $\textbf{t}$ of the corresponding tail entity $t$, i.e., minimizes the score of a triple as follows:

 \beq{
 	\label{eq:transe}
 	s(h,r,t) = ||\mathbf{h}+\mathbf{r}-\mathbf{t}||_2^2.
 }

Subsequently, various models have been proposed to improve the capability of TransE. For example, 
TransH deals with the flaws caused by the relations with the properties of reflexive, one-to-many, many-to-one, and many-to-many by projecting the entities into the relation-specific hyperplane, which enables different roles of an entity in different relations/triplets~\cite{Wang14}.
Instead of projecting entities and relations into the same space, TransR builds entity and relation embeddings in separate entity space and relation space~\cite{lin2015learning}. %TransR constructs a mapping matrix for each relation, while 
TransD~\cite{ji-2015} determines the mapping matrices by both the entities and the relations with the hope to capture the diversity of entities and relations simultaneously. 
TransG~\cite{xiao2015transg} further addresses the ambiguity of a relation by incorporating a Bayesian non-parametric infinite mixture model to generate multiple translation components for a relation. 
TransAt~\cite{qian2018translating} determines a relation between two entities in two steps inspired by the human cognitive process. It first checks the categories of entities and then determines a specific relation by relation-related attributes through a relation-aware attention mechanism. 
To model and infer the symmetry/antisymmetry, inversion, and composition patterns together, RotatE~\cite{sun2018rotate} maps the entities and relations to the complex vector space and defines each relation as a rotation from the head entity to the target entity.

\subsubsection{Multiplicative Models}
Multiplicative models produce the entity and relation embeddings via tensor product as follows:

\beq{
	s(h,r,t) = \mathbf{h}^T \mathbf{M}_r \mathbf{t},
}

\noindent where $\mathbf{M}_r$ is an asymmetric $d \times d$ matrix that models the interactions of the latent components in the $r$-th relation. 
Given a relation matrix $M_r$, a feature in the above tensor product is ``on" if and only if the corresponding features of both entities $h$ and $t$ are ``on", which can capture the relational patterns between entities. 
The representative tensor product-based models are RESCAL~\cite{NickelICML11}. However, the tensor product requires a large number of parameters as it models all pairwise interactions. 
To reduce the computational cost, DisMult~\cite{yang2014embedding} is proposed to use the diagonal matrix with the diagonal vector indicating the embedding of the relation $r$ to reduce the number of parameters.
Based on DisMult, ComplEx~\cite{pmlr-v48-trouillon16} further handles asymmetry thanks to the capabilities of complex embeddings.
Instead of tensor product, HolE~\cite{NickelAAAI16} uses the circular correlation of vectors to represent pairs of entities, i.e., $\mathbf{h} \star \mathbf{t}$ where each element $[\mathbf{h} \star \mathbf{t}]_k = \sum_{i=0}^{d-1} h_i t_{(k+i) \text{ mod } d}$.                                                           
SimplE~\cite{NIPS2018_7682} is based on canonical Polyadic decomposition~\cite{Frank1927}, which learns an embedding vector for each relation, and a head embedding plus a tail embedding for each entity. To address the independence between the two embedding vectors of the entities, SimplE introduces the inverse of relations and calculates the average canonical Polyadia score of $(h,r,t)$ and $(t ,r^{-1}, h)$.

In addition, ANALOGY~\cite{pmlr-v70-liu17d} supports analogical inference by constraining the relation matrix $\mathbf{M}_r$ to be the normal matrix in linear mapping, i.e. $\mathbf{M}_r^T\mathbf{M}_r = \mathbf{M}_r\mathbf{M}_r^T$.
DihEdral~\cite{xu2019relation} employs finite non-Abelian group to account for relation compositions, which supports the non-commutative property of a relation. For example, one exchanges the order between \textit{parent\_of} and \textit{spouse\_of} will result in different relations (\textit{parent\_of} as opposed to \textit{parent\_in\_law\_of}), which indicates the non-commutative property of the relation. 

\subsubsection{Deep Learning Models}
The deep learning models, such as the convolutional neural network (CNN), the recurrent neural network (RNN), and the graph neural network (GNN), are also leveraged as encode functions to embed entities and relations in KGs. For example, ConvE~\cite{dettmers2018convolutional} first concatenates a pair of head embedding and relation embedding and then applies 2D convolutions over those embeddings to predict the tail entity. ConvR~\cite{jiang2019adaptive} extends the global filters in ConvE to relation-specific filters, and InteractE~\cite{vashishth2020interacte} captures more interactions by additional convolution operations. Instead of only performing the convolutions on head and relation embeddings, ConvKB~\cite{nguyen2018novel} applies convolutions over the concatenated embeddings of the head entity, relation, and the tail entity together, which captures the translation relationship in a triplet. CapsE~\cite{Dai-2019} applies the capsule network~\cite{Sara-NIPS-2017} as the convolution function to encode the entities. RSN~\cite{pmlr-v97-guo19c} integrates RNNs with residual learning to capture the long-term relational dependencies in knowledge graphs. 

Recently, GNNs are also attempted to encode the neighboring entities and relations together beyond a single triplet. For example, to adapt to the multiple relations in knowledge graphs, R-GCN  extends the transformation weights in GCN to relation-aware weights~\cite{schlichtkrull2018modeling}. To predict the relation between two entities, R-GCN uses relation-aware GCN as the encoder to represent each entity and then leverages DisMult~\cite{yang2014embedding} as the decoder to score a given triplet based on the encoded entities and the introduced diagonal matrix $\mathbf{M}_r$ for each relation. 
Instead of using DisMult as the decoder, SACN~\cite{ShangAAAI2019} uses a variant ConvE~\cite{dettmers2018convolutional} as the decoder, where the encoder is also a relation-aware GCN.
M-GNN~\cite{Wang-ijcai2019} replaces the mean aggregator in each graph convolution layer in R-GCN with a multi-layer perceptron (MLP) to support the injective property, i.e., to map two entities to the same location only if they have identical neighborhood structures with identical embeddings on the corresponding neighbors.
In addition to the relation-aware transformation weights, VR-GCN~\cite{ye2019vectorized} and CompGCN~\cite{vashishth2019composition} explicitly represent relations and further combine entity and relation embeddings by the operations like subtraction and multiplication. KBGAT~\cite{Deepak-ACL2019} enables triplet-aware weights by concatenating the embeddings of the head entity, tail entity, and the relation in a triplet to learn its weight.
More details of knowledge graph embeddings can be found in~\cite{ji2020survey,rossi2020knowledge}.

\vpara{Summary.}
The neural reasoning methods utilize shallow embedding models, such as the translation-based models, the multiplicative-based models, or the deep neural network models including CNN, RNN, and GNN, to embed entities and relations in KGs, based on which they perform the reasoning task. These methods are fault-tolerant, as the reasoning is built on the semantic representations rather than the symbolic representations of entities and relations. However, these simple neural network models cannot infer the answers when the complex logic relations $\{\land, \lor, \neg \}$ exist between the head and the answer entities. Besides, the neural networks lack interpretation, as they cannot provide explicit rules for explaining the reasoning results.

\subsection{Symbolic Reasoning}
\label{sec:symbolic_reasoning_for_KGC}

Symbolic reasoning aims at deducing general logic rules from the knowledge graphs. The entities derived from the given head entity and the query relation following the logic rules are returned as the answers. Existing symbolic reasoning methods are mainly the search-based ILP methods, which usually search and prune rules.
This section will first introduce the state-of-the-art search-based method AMIE~\cite{galarraga2013amie} in detail, and then briefly highlight the improvements of the later methods based on AMIE.

\textit{AMIE}~\cite{galarraga2013amie} explores logic rules following two steps. The first step is Rule Extending, which extends candidate rules by three kinds of operations. The second step is Rule Pruning, which prunes the corrupt rules and outputs the confident rules according to  the predefined evaluation metrics. For implementation, SPARQL on graph databases is adopted to search the proper facts $(h,r,t)$ in KGs which satisfy the rules extended by the first step and exceed the lower bound of the metrics defined by the second step. The details for each part are as follows:

\begin{enumerate}
	\item 
 \textit{Rule Extending} is to generate candidate rules by adding three kinds of new atoms into existing rules iteratively, where the first atom is named as the dangling atom which has a fresh variable as one argument and holds an existing variable that appears in other atoms of the rule as the other argument, the second atom is called the instantiated atom which has an argument instantiated by an entity and shares the other argument with other atoms, and the third atom is the closing atom which shares both of the arguments with other atoms. We present an example of the existing rule and three kinds of atoms as follows:

\beqn{
	\label{eq:amie_op} \nonumber
	  \text{Rule:}   &&  r_h(x, y) \leftarrow r_1(x,z_1)\land ... \land r_n(z_{n-1}, y) \\ 	\nonumber
	  &&\text{Dangling atom:}     r^D(x, k), r^D(k,y), ... \\ \nonumber
	 &&\text{Instantiated atom:}      r^I(x, K), r^I(K,y), ...  \\ \nonumber
	  &&\text{Closing atom:}   r^C(x, z), r^C(z,y), ...  \nonumber
}

\noindent where $r_i(x,y)$ is an atom that indicates the variable $x$ and $y$ satisfy the predicate $r_i$. The new atom $r^D(x,k)$ is a dangling atom that shares one variable $x$ with the atoms in the rule and holds a fresh variable $k$. The new atom $r^I(x,K)$ is an instantiated atom which instantiates one augment with a constant $K$ while sharing the other augment with the rule. The new atom $r^C(x,z)$ shares both $x$ and $z$ with the rule.
 Due to the fact that the time complexity of extending rules grows exponentially with the increase of the new atoms, a maximal length of the rules is adopted to stop the extending procedure early. 

\item  \textit{Rules Pruning } applies two metrics, i.e., the head coverage  ($hc$) and the confidence ($conf$), to prune and output the rules.  The two metrics are defined as follows:

{\small 
	\beqn{
		\label{eq:hc}
	hc(r(x, y)\!\leftarrow\!B)\!\!\!\!\!&:=&\!\!\!\!\!\frac{support(r(x, y)\!\leftarrow\!B)}{\#(x^{'}, y^{'})\!:\! r(x^{'}, y^{'})},\\
	conf(r(x, y)\!\leftarrow\!B)\!\!\!\!\!&:=&\!\!\!\!\!\frac{support(r(x, y)\!\leftarrow\!B)}{\#(x, y)\!: \!\exists z_1, ..., z_m:\!B\!}, \nonumber
}
}

\noindent where $B$ is the abbreviation of the rule body, i.e., $B=r_1\land  \cdots \land r_n$, 
the nominator $support(r(x, y)\!\leftarrow\!B)$ indicates the  number of the distinct pairs of the head and tail entities $(x,y)$ in KGs (i.e., the facts in KGs) which satisfy the relation $r$ and are derived by the rule body $B$.
The denominator $\#(x^{'}, y^{'}): r(x^{'}, y^{'})$ represents the number of the facts in KGs which satisfy the relation $r$, where the facts might be derived by $B$ or not.
The denominator $\#(x, y)\!: \!\exists z_1, ..., z_m:\!B$ with 
$\{z_1, ..., z_m\}$ as the variables in the rule body $B$ excluding the head and tail variables $x$ and $y$, denotes the number of the facts that can be derived by the rule body $B$, where the facts might be observed in KGs or not. 
Given these notations, 
$hc(r(x, y)\!\leftarrow\!B)$ indicates the proportion of the facts in KGs that are covered by the rule $r(x, y)\!\leftarrow\!B$, and $conf(r(x, y)\!\leftarrow\!B)$ denotes the proportion of the facts derived by the rule $r(x, y)\!\leftarrow\!B$ that are included in KGs. 
Note the two metrics are based on the close world assumption that the facts not included in the KGs are false. 
From the definition, we can see that $hc$ represents the coverage/recall of a rule, and $conf$ can reflect the predictive precision of a rule. The two metrics measure the quality of the mined rules in different aspects and are used to prune the rules simultaneously, i.e. the rules whose $hc$ and $conf$ are lower than the predefined thresholds will be discarded and the remaining rules will be outputted.
\end{enumerate}

%Intuitively, the addition of more atoms and constraints to the rule will transform it from general to specific. Due to that $HC$ follows the $Closed World Assumption$ (CWA) which assumes the facts not contained in the KG are false, the rules extending procedure should always decrease its $HC$, which means that $HC$ is a monotonic metric. While $Conf_{pca}$ is based on the $Partial Completeness Assumption$ (PCA) which only considers the facts occurring in the KG, that is, if we know one $y$ for a given $x$ and $r$, then we know all $y$ for that $x$ and $r$. So, $Conf_{pca}$ is a non-monotonic metric. Generally, $minHC$ reflects the temporary quality of the rule, so it is used to prune the candidate rules generated by $Rules~Extending$. While $minConf_{pca}$ catches the incremental quality of the rule, which is commonly used to measure the confidence gain compared with its parent rule after $Rules~Extending$.  Consequently, both of the two metrics measure how good the mined rules in different aspects, so they are all used to prune the rules lower than the metrics bound and output the rules satisfying the settings.

To implement rule extending and pruning efficiently, the authors project the two processes into a SPARQL query and fire it on KGs:

 \beqn{
  	\text{SELECT}~\text{?r},\text{WHERE}~r_h \land r_1 \land ... \land r_{n} \land~?r(X, Y) \\  \nonumber
  	\text{HAVING COUNT} (r_h) \geq K \nonumber
  }
  
\noindent where $?r$ is the new predicate/relation to be added, $X$ and $Y$ represent variables that are either fresh or present in the rule. The above query selects the relation $?r$ such that the result of the query $r_1 \land ... \land r_{n} \land ?r(X, Y) \rightarrow r_h$ is greater than $K$. Through setting the proper 
$K$, the lower bound of the metrics in Eq.\eqref{eq:hc} can be satisfied. For example, if choosing $K$ as $\theta$ --- the lower bound of $hc$ --- times the number of the facts in KGs that satisfy the relation $r_h$, the metric $hc$ of the resulting rules will be greater than $\theta$ --- which is what we want.

 %No matter which operator is applied in particular, \refine{it needs to choose the new proper atoms that fulfill the $minHC$ and $minConf_{pca}$ as much as possible. AMIE designs the count projection queries in the form of}: \zj{What does this paragraph mean?} \cbreply{Fixed, rewrite the paragraph.}

%\noindent where $K:=minHC \times size(H)$ projecting the $HC$ bound into an absolute number that can be expressed as COUNT(*). Such queries select the relation \textit{\textbf{r}} ensuring that the $HC$ of the rule $r_1 \land ... \land r_{n-1} \land~\textit{\textbf{r(X, Y)}} \Rightarrow H$ is bigger or equal than pre-defined $minHC$. Trough setting the proper K, it can select the appropriate candidate atoms. \refine{In practice, AMIE leverages SPARQL and SQL to efficiently conduct the queries in the database.}

\textit{AMIE+}~\cite{galarraga2015fast} improves the implementation efficiency of AMIE~\cite{galarraga2013amie} by revising both the Rule Extending process and the metrics defined in the Rule Pruning process. 

In the process of Rule Extending, AMIE+ extends a rule only if it is possible to close it\footnote{A rule is closed if every variable in the rule appears at least twice.} before exceeding the predefined maximal rule length. In other words, AMIE+ will not add the dangling atoms at the last step, as this will introduce a fresh variable which results in another non-closed rule. Instead, the instantiated atoms and the closing atoms will be added at the last step to close the rule. The SPARQL queries used to search rules are also simplified. For example, suppose we add a dangling atom to a rule $R_p$ (the parent rule) to produce a child rule $R_c$, if the predicate/relation of the new atom has already existed in $R_p$\footnote{AMIE+ allows us to mine the recursive rules, i.e., the head relation occurs in the rule body.}, $support(R_c)$ will be the same as $support(R_p)$. Thus it is not necessary to calculate $support(R_c)$ again, which will speed up the SPARQL query process.

%a simplifying procedure is applied to shorten the projection queries as much as possible. For example, when we apply an $\mathcal{O_D}$ operation to add a new atom to a rule $R_p$ (the parent rule), that will produce a child rule $R_c$. If the relation of the new atom has already existed in the $R_p$\footnote{AMIE allows to mine the recursive rules, that is, the head relation occurs in the rule body.}, the projection query for $supprt(R_c)$ is same as that for $support(R_p)$. This observation can be used to speed up the query process, because the query for $supprt(R_c)$ has one fewer join.

In the process of Rule Pruning, calculating either $hc$ or $conf$ requires reckoning the number of the facts derived by a rule. If the rule body contains the atoms with many variables, the derivation process will be expensive. To speed up this, AMIE+ proposes a method to approximate the metrics based on some pre-computed statistics, such as the size of the joins between two relations. Consequently, AMIE+ achieves 100$\times$ speedup compared with AMIE. 

However, these search-based ILP methods~\cite{chen2016scalekb,galarraga2015fast,galarraga2013amie} are still not scalable for large KGs, because they are based on the projection queries implemented by SQL or SPARQL and the huge search space cannot be easily reduced. 

\textit{RLvLR}~\cite{omran2018scalable}  is one of the early attempts to use the embedding technique to sample the entities and facts that are relevant to the target predicate/relation, which hugely reduces the search space. Specifically, first, RLvLR samples a sub knowledge graph that is relevant to the target predicate; second, it leverages the knowledge graph embedding model RESCAL~\cite{nickel2011three} to generate embeddings for entities and relations in the subgraph, and makes the embedding for an augment of a predicate as the average value of the embeddings of all the entities appearing in the position of this augment; third, it uses a scoring function based on the embeddings to guide and prune the rule search, which turns out to be rather effective for extracting rules; finally, the searched candidate rules are evaluated according to the metrics $hc$ and $conf$ defined in AMIE and are computed by efficient matrix multiplication.  By incorporating the embedding technique, the efficiency of the rule searching process is significantly improved.

Statistical measures~\cite{chen2016scalekb,galarraga2015fast,galarraga2013amie} such as confidence scores may misjudge the rule quality because KGs are inherently incomplete. RuLES~\cite{ho2018rule} leverages the embedding technique to measure the quality of the learned rules. It incorporates the external text information of entities to obtain their embeddings, based on which the confidence score of a fact $r(h,t)$ can be calculated as the dot-product between $h$ and $t$. Then, RuLES defines the external quality of a learned rule as the average confidence score of all the facts derived by the rule. Finally, both the statistical and embedding measures are intertwined to judge the quality of the learned rules more precisely.

\vpara{Summary.}
The traditional search-based ILP methods rely heavily on the search algorithms, various pruning techniques, and efficient database operations, which have several limitations: first, due to the strict matching and the discrete logic operations used during the process of rule searching, the symbolic methods are intolerant of the ambiguous and noisy data; second, the pre-defined evaluation metrics and the rule formations restrict the expressiveness of the learned rules. %Later methods combining the neural network techniques with the symbolic reasoning techniques address the above limitations well, which will be elaborately introduced in the following sections.

\subsection{Neural-Symbolic Reasoning}
\label{sec:neural_symbolic_reasoning_for_KGC}
Although symbolic reasoning is good at logical inference and has powerful interpretability, it has difficulties to deal with the uncertainty of entities and relations and the ambiguity of natural languages, i.e., it is not resilient against data noises.  On the contrary, the neural networks are fault-tolerant, as they learn the abstract semantics, i.e., embeddings, and further compare these embeddings instead of the literal meaning between entities and relations by symbolic representations. The recent advances on reasoning combine these two kinds of reasoning methods\footnote{Henceforth, we also name symbolic reasoning as rule learning, and name neural reasoning as embedding learning.}. Typically, there are three main combination methodologies. The first kind targets at neural reasoning but leverages the logic rules to improve the embeddings in neural reasoning, named as \textit{symbolic-driven neural reasoning}. The second kind replaces the neural reasoning with a probabilistic framework, i.e., builds a probabilistic model to infer the answers, where the logic rules are designed as features in the probabilistic model, which is named as \textit{symbolic-driven probabilistic reasoning}. And the third kind aims to infer rules by symbolic reasoning, but incorporates the neural networks to deal with the uncertainty and ambiguity of data. This kind of methods also reduces the search space in symbolic reasoning, which are named as \textit{neural-driven symbolic reasoning}. We will explain the three types of neural symbolic reasoning methods as follows.

\subsubsection{Symbolic-driven Neural Reasoning}
\label{sec:symbolic_driven_neural_reasoning}
\begin{figure}[t]
	\centering
	\includegraphics[width=0.48\textwidth]{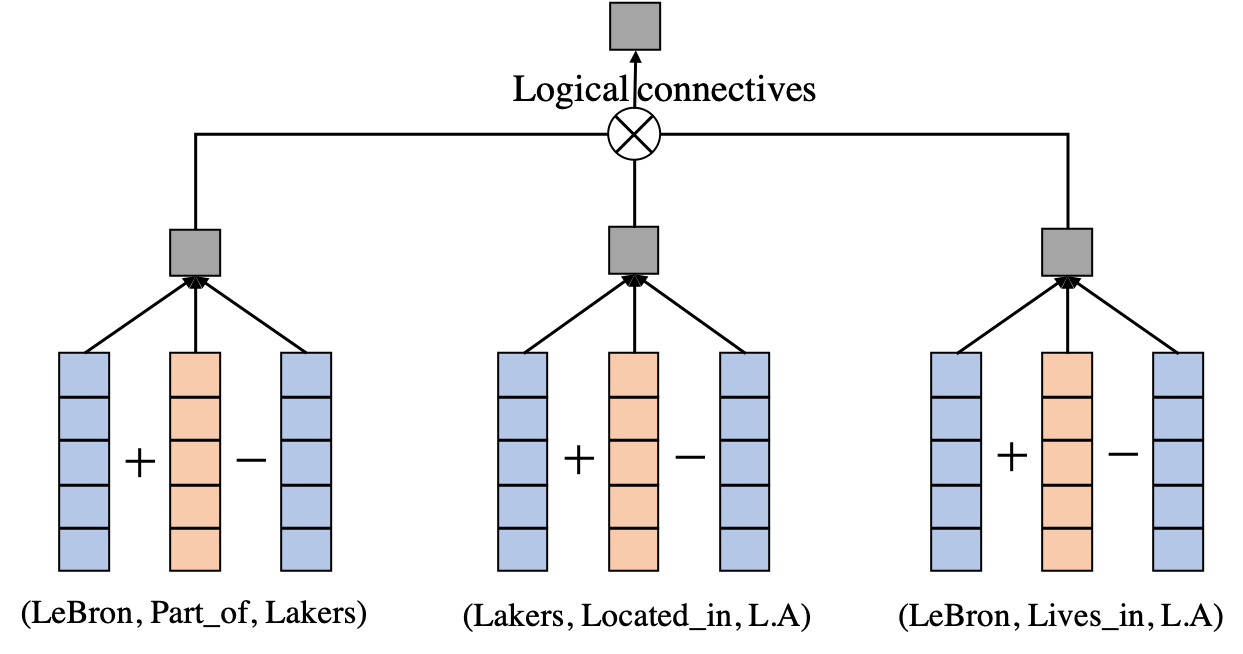}
	\caption{\label{fig:KALE} Illustration of rule scoring in KALE~\cite{guo2016kale}.}
\end{figure}

\begin{table*}[t]
	\centering
	\caption{
		\label{tb:format_first_logic}
		The format of first-order logic~\cite{wang2019logic}. For example, the third line defines the transitivity rule $\left(r_{1}+r_{2}\right) \Rightarrow r_{3}$, following which we can infer a new triple $\left(e_{1}, r_{3}, e_{3}\right)$ from two existing triplets $\left(e_{1}, r_{1}, e_{2}\right)$ and $\left(e_{2}, r_{2}, e_{3}\right)$. }
	
	\begin{tabular}{ll}
		\hline \textbf{Triple and ground rule} & \textbf{The format of first-order logic} \\
		\hline
		$(h, r, t)$ & $r(h) \Rightarrow t$ \\
		$\left(h, r_{1}, t\right) \Rightarrow\left(h, r_{2}, t\right)$ & 
		$\left[(h \in C) \wedge\left[r_{1}(h) \Rightarrow t\right]\right] \Rightarrow\left[r_{2}(h) \Rightarrow t\right]$
		\\
		
		$\left(e_{1}, r_{1}, e_{2}\right)+\left(e_{2}, r_{2}, e_{3}\right) \Rightarrow\left(e_{1}, r_{3}, e_{3}\right)^{\tnote{*}}$
		& 
		$\left[\left[r_{1}\left(e_{1}\right) \Rightarrow e_{2}\right] \wedge\left[r_{2}\left(e_{2}\right) \Rightarrow e_{3}\right]\right] \Rightarrow\left[r_{3}\left(e_{1}\right) \Rightarrow e_{3}\right]$
		\\
		$\left(h, r_{1}, t\right) \Leftrightarrow\left(t, r_{2}, h\right)$ & 
		$\left[\left[r_{1}(h) \Rightarrow t\right] \Rightarrow\left[r_{2}(t) \Rightarrow h\right]\right] \wedge\left[\left[r_{2}(t) \Rightarrow h\right] \Rightarrow\left[r_{1}(h) \Rightarrow t\right]\right]$ \\
		\hline
	\end{tabular}
\end{table*}

%Embedding learning and rule learning can benefit and complement each other. On one hand, logic rules can infer additional triples and add additional reasoning information for representation methods. On the other hand, embeddings encoded with rich semantics can reduce the search space significantly.  

The basic idea of symbolic-driven neural reasoning is to learn the entity and relation embeddings on not only the original observed triplets in KGs but also the triples or ground rules inferred following some pre-defined rules. For example, KALE~\cite{guo2016kale} deals with two types of rules:

\beqn{
	\forall x, y: (x, r_s, y) &\Rightarrow& (x, r_t, y), \\ \nonumber
	\forall x,y,z: (x, r_{s_1}, y) \land (y,r_{s_2},z) &\Rightarrow& (x, r_{t} , z).	
}

They find all the ground rules of the above two types of rules, assign a score to each ground rule indicating how likely a ground rule is satisfied, and finally learn the entity and relation embeddings on the training set of the original triplets and the ground rules.
They employ  t-norm fuzzy logics~\cite{Petr2020}, which define the true value of a rule as a composition of the truth values of its constituents through specific t-norm based logical connectives, to calculate a score for a ground rule $f_1 \Rightarrow f_2$ as:

\beq{
	\label{eq:KALE_score}
	s(f_1 \Rightarrow f_2) = s(f_1) s(f_2) -s(f_1) +1,
}

\noindent where $f$ denotes an atom, i.e., a triplet, or a formula that is composed of multiple atoms associated by logical operations $\{\land, \lor, \neg \}$. If $f$ in the above equation is a triplet, 
its scored is computed by TransE in Eq.\eqref{eq:transe}. 
If $f$ is a formula, its score is defined as a composition of the scores of its constituents:

\beqn{
	\label{eq:conjunction}
	s(f_1 \wedge f_2) &=& s(f_1)\cdot s(f_2), \\\nonumber
	s(f_1 \vee f_2) &=& s(f_1)+s(f_2)-s(f_1)\cdot s(f_2),\\\nonumber
	s(\neg f_1) &=& 1 - s(f_1).\\\nonumber
}

The procedure of calculating the score of a  rule (LeBron, Part\_of, Lakers) $\land$ (Lakers, Located\_in, L.A)  $\Rightarrow$ (LeBron, Lives\_in, L.A) is illustrated in Figure~\ref{fig:KALE}, where the three triplets involved in the rule are scored by TransE, and their scores are combined according to the logical connectives in Eq.~\eqref{eq:KALE_score} and Eq.~\eqref{eq:conjunction} into the score of the rule.
The observed triplets and the rules with high scores are unified as positive instances to learn entity and relation embeddings. 
The initial rules are selected as the top-ranked ones based on the scores calculated by Eq.\eqref{eq:KALE_score}, where the initial entity and relation embeddings are produced by TransE. The rules will be kept the same during the following embedding learning process.

Based on KALE, Guo et al. further propose RUGE\cite{guo2017ruge} to change the one-round injection of rules into an iterative manner. Instead of directly treating a ground rule as the positive instance by KALE, RUGE injects the triplets derived by some rules as the unlabeled triplets to update the entity/relation embeddings. Since the unlabeled triplets are not necessarily true, the authors predict a probability for each unlabeled triplet based on the current embeddings. Then the embeddings are updated based on both the labeled and unlabeled triplets. 
The initial rules are obtained by AMIE~\cite{galarraga2013amie} and also not updated in the following algorithm.
In this way, the unlabeled triplet scoring process and the embedding updating process are iteratively computed.

However, KALE  and RUGE calculate the score of  a rule or a formula as the composition of the scores of its constituents (Cf. Eq.\eqref{eq:KALE_score} and Eq.\eqref{eq:conjunction}), which may result in a high score for a rule or a formula even if the triplets in it are totally irrelevant with each other, as the scores of the triplets are estimated separately. 
To solve this problem, Wang et al.~\cite{wang2019logic} transform a triplet or a ground rule into first-order logic, and then score this first-order logic by performing some vector/matrix operations based on the embeddings of the entities and relationships included in the first-order logic. Table~\ref{tb:format_first_logic} illustrates the format of the first-order logic, and Table \ref{tb:math_logical} presents how to score the first-order logic by the mathematical expression. 
%For example, a triplet $(h,r,t)$ can be expressed by the first-order logic $r(h) \Rightarrow t$, and the first-order logic $a \Rightarrow b$ can be expressed by $\mathbf{a}-\mathbf{b}$.  
In this way, different components, i.e., triplets,  included in the same rule have directly interacted in the vector space, which guarantees both the rule and its encoding format have one-to-one mapping transformations.

%{not sure about this}

%In contrast to using rules as additional information for embedding learning, RLvLR\cite{pouya2018scalable} use embedding models to effectively prune the rule search. RLvLR introduces a new scoring function that allow fast ranking of rules, based on embeddings of entities, predicates and arguments. Argument embedding is defined as the average value of the embeddings of all the entities that related to specified entity. In addition, efficient numerical algorithms for matrices are also employed to speed up the search process.

The above methods infer rules one time in the beginning and keep the rules invariant during the learning process. Thus, the rules will impact embedding learning, while the embeddings will not benefit the inference of rules. On the contrary, while IterE~\cite{zhang2019iere} also infers new rules based on the updated embeddings at each iteration, it specifically infers new rules and derives new triplets from the rules based on the entity and relation embeddings, and then updates these embeddings based on the extended triplet set. The two processes are executed iteratively. The new confident rules are inferred based on their scores which are calculated by performing some matrix operations over the matrices of the relations included in the rules\footnote{IterE use DistMult~\cite{yang2014embedding} to learn the embeddings, thus a relation is represented as a matrix by DistMult.}. To obtain the initial pool of rules, IterE proposes a pruning strategy which has a similar idea  as AMIE~\cite{galarraga2013amie} but combines the operations of traverse and random selection to balance the searching process of potential rules and the convergence of highly possible rules.

\begin{table}[t]
	\centering
	\caption{
		\label{tb:math_logical}
		Mathematical expression of first-order logic~\cite{wang2019logic}.}
	\begin{tabular}{l l}
		\hline
		\textbf{First-order logic} & \textbf{Mathematical expression}  \\ \hline
		$r(h)$ & $\mathbf{r}+\mathbf{h}$ \\ 
		$a \Rightarrow b$  & $\mathbf{a}-\mathbf{b}$ \\ 
		$h \in C$ & $\mathbf{h} \cdot \mathbf{C}$ ($\mathbf{C}$ is a matrix) \\ 
		$a \wedge b$ & $\mathbf{a}$ $\otimes$ $\mathbf{b}$ \\ 
		$a \Leftrightarrow b$ & ($\mathbf{a}-\mathbf{b}) \otimes(\mathbf{a}-\mathbf{b}$) \\ \hline
	\end{tabular}
	
\end{table}

\subsubsection{Symbolic-driven Probabilistic Reasoning}
\label{sec:symbolic_driven_probabilistic_reasoning}
\begin{figure}[t]
	\centering
	\includegraphics[width=0.47\textwidth]{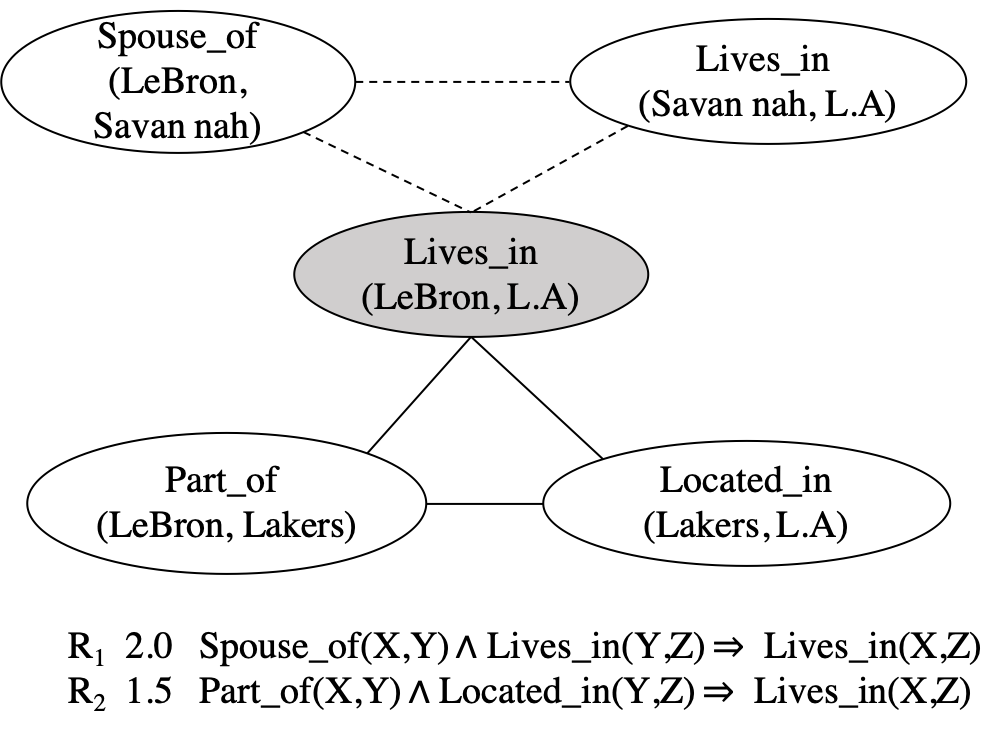}
	\caption{\label{fig:mln} Two examples of rules and the corresponding ground Markov logic network (Dotted lines are clique potentials associated with rule $R_1$, and solid lines are with rule $R_2$. The grey node is the unobserved triplet to be inferred)~\cite{wang2013programming}.}
\end{figure}

Symbolic-driven probabilistic reasoning combines the first-order logic and the probabilistic graphical model to learn the weights of logic rules in a probabilistic framework and thus soundly handles the uncertainty.
This kind of methods usually qualify logic rules by firstly grounding the rules, i.e., iteratively substituting variables in the atoms of the rules with any entities in KG until no new facts/triplets can be deduced, and then attaching probabilities to the ground rules. In a sense, the confidence/quality of a logic rule can be defined as the probability distribution over a pre-constructed probabilistic ground graph as shown in \figref{fig:mln}. Instead of leveraging embedding techniques to qualify the rules, symbolic-driven probabilistic reasoning designs probabilistic models to measure the confidence of rules. In this section, we present two typical probabilistic models, Markov Logic Network (MLN) and ProbLog to explain the characteristics of this category of methods, and then briefly introduce several similar methods.

\textit{Markov Logic Network (MLN)}~\cite{richardson2006markov} is to build a probabilistic graphic model based on the pre-defined rules and the facts in KGs and then learn the weights for different rules. 
Specifically, given a set of rules $\{\gamma_i\}$, each $\gamma_i$ can be grounded by the ground atoms (i.e., triplets) from the KGs. Then based on these ground rules, a Markov logic network can be built as follows:

\begin{enumerate}
	\item A node is built for each ground atom in each ground rule, and the value of the node is set as 1 if the ground atom is observed in KGs, 0 otherwise.
	\item An edge is built between two nodes if and only if the corresponding two ground atoms can simultaneously be used  to instantiate at least one rule. 
	\item All the nodes, i.e., ground atoms, in a ground rule  form a (not necessarily maximal) clique, which corresponds to a feature, with the value as 1 if the ground rule is true, 0 otherwise. A weight $w_i$ is associated with each rule $\gamma_i$.
\end{enumerate}

Take \figref{fig:mln} as an example, given two rules $R_1$ and $R_2$ together with the observed ground atoms Spouse\_of(LeBron, Savan nah), Lives\_in(Savan nah, L.A), Part\_of(LeBron, Lakers), Located\_in(Lakers, L.A) and the unobserved ground atom Lives\_in(LeBron,L.A), a Markov logic network can be derived. For example, an edge is built between Lives\_in(LeBron,L.A) and Spouse\_of(LeBron, Savan nah) as they can simultaneously be used to instantiate $R_1$. Lives\_in(LeBron,L.A),  Spouse\_of(LeBron, Savan nah) and Lives\_in(Savan nah, L.A) form a clique, as they form a ground rule of $R_1$. 
With the built Markov logic network, 
the joint distribution of the values $X$ of all the nodes in the network is defined as:
  
\beq{
	P(X=x) = \frac{1}{Z}\exp(\sum_{i}w_{i}n_{i}(x)),
}

\noindent where $n_i(x)$ is the number of true groundings of the rule $\gamma_i$ and $w_i$ is the weight corresponding to the rule $\gamma_i$. 
Then, MCMC algorithm is applied for inference in MLN and the weights are efficiently learned by optimizing a pseudo-likelihood measure~\cite{richardson2006markov}.
With the learned weights, we can infer the probability of Lives\_in(LeBron,L.A)  given its neighboring ground atoms. 

However, the inference process in MLNs is difficult and inefficient due to the complicated graph structure among triplets. Moreover, the missing triplets in KGs also impact the inference results by rules. Since the recent embedding techniques can effectively predict the missing triplets and can be efficiently trained with stochastic gradient, pLogicNet~\cite{qu2019policnet} proposes to combine the MLN and graph embedding techniques to tackle the above problems. The basic idea is to define the joint distribution of  the triplets/facts in KGs with a MLN and associate each logic rule with a weight, but effectively learn them via variational EM algorithm~\cite{Neal1998}. In the algorithm, E-step infers the plausibility of the unobserved triplets, in
which the variational distribution is parameterized by a knowledge graph embedding model such as TransE, while the M-step updates the weights of logic rules by optimizing the pseudo-likelihood defined on both the observed triplets and those inferred by the knowledge graph embedding model.

\textit{ProbLog}~\cite{de2007problog} is a probabilistic extension of Programming in Logic\footnote{Prolog is a logic programming language associated with artificial intelligence and computational linguistics. The official website is https://www.swi-prolog.org/.} (Prolog). Compared with Prolog, ProbLog adds a probability for each clause $c_i$, which represents either a rule or a ground atom.  An example of clauses that are used to derive the query, i.e. finding all the entities $e$ that satisfy Lives\_in(LeBron, S), are presented as follows:

{\small \beqn{
	\nonumber	
	\text{0.7}:&&\!\!\!\!\!\text{Spouse\_of}(X,Y) \land \text{Lives\_in}(Y,Z) \Rightarrow \text{Lives\_in}(X, Y)\\ \nonumber	
	\text{0.8}:&&\!\!\!\!\!\text{Part\_of}(X,Y) \land \text{Located\_in}(Y,Z) \Rightarrow \text{Lives\_in}(X, Y) \\ 	\nonumber
	\text{1.0}:&&\!\!\!\!\!\text{Spouse\_of}(\text{LeBron}, \text{Savannah})  \\ 	\nonumber
	\text{0.9}:&&\!\!\!\!\!\text{Lives\_in}(\text{Savannah}, \text{L.A})  \\ 	\nonumber
	\text{0.9}:&&\!\!\!\!\!\text{Part\_of}(\text{LeBron}, \text{Lakers})  \\ 	\nonumber
	\text{1.0}:&&\!\!\!\!\!\text{Located\_in}(\text{Lakers}, \text{L.A}) \nonumber
}}

Given a query $q$,  the success probability $P(q|T) $ is defined as:

\beqn{
	\label{eq:p_q_T}
	P(q|T) &=& \sum_{L\subseteq L_T}P(q, L|T), \\
	\label{eq:p_q_L_T}
	P(q, L|T) &=& P(q|L) \cdot P(L|T),\\
	\label{eq:p_q_L}
	P(q|L) &=& \left\{
	\begin{aligned}
		1 \qquad&\exists \theta:  L \models q\theta, \\
		0 \qquad&otherwise,
	\end{aligned}
	\right. \\
	\label{eq:p_L_T}
	P(L|T) &=& \prod_{c_i \in L}p_i \prod_{c_i \in L_T \backslash L}(1-p_i),
}

\noindent where $T = \{p_1:c_1, ..., p_n:c_n\}$ depicts a probability distribution over the causes $L \subseteq L_T = \{c_1, ... c_n\}$. Eq.\eqref{eq:p_q_T} indicates the success probability of query $q$ being decomposed into the summation of all the joint probabilities of the query and each possible cause set $L$. Eq.\eqref{eq:p_q_L_T} further decomposes $P(q, L|T)$ into $P(q|L)$ and $P(L|T)$. $P(q|L)$ represents the probability of the query $q$ given the cause set $L$, whose value equals 1 if there is at least one answer substitution $\theta$ instantiating $L$ and making the query true. Eq.\eqref{eq:p_L_T} explains how to calculate the probability of a cause set $L$.

\begin{figure}[t]
	\centering
	\includegraphics[width=0.47\textwidth]{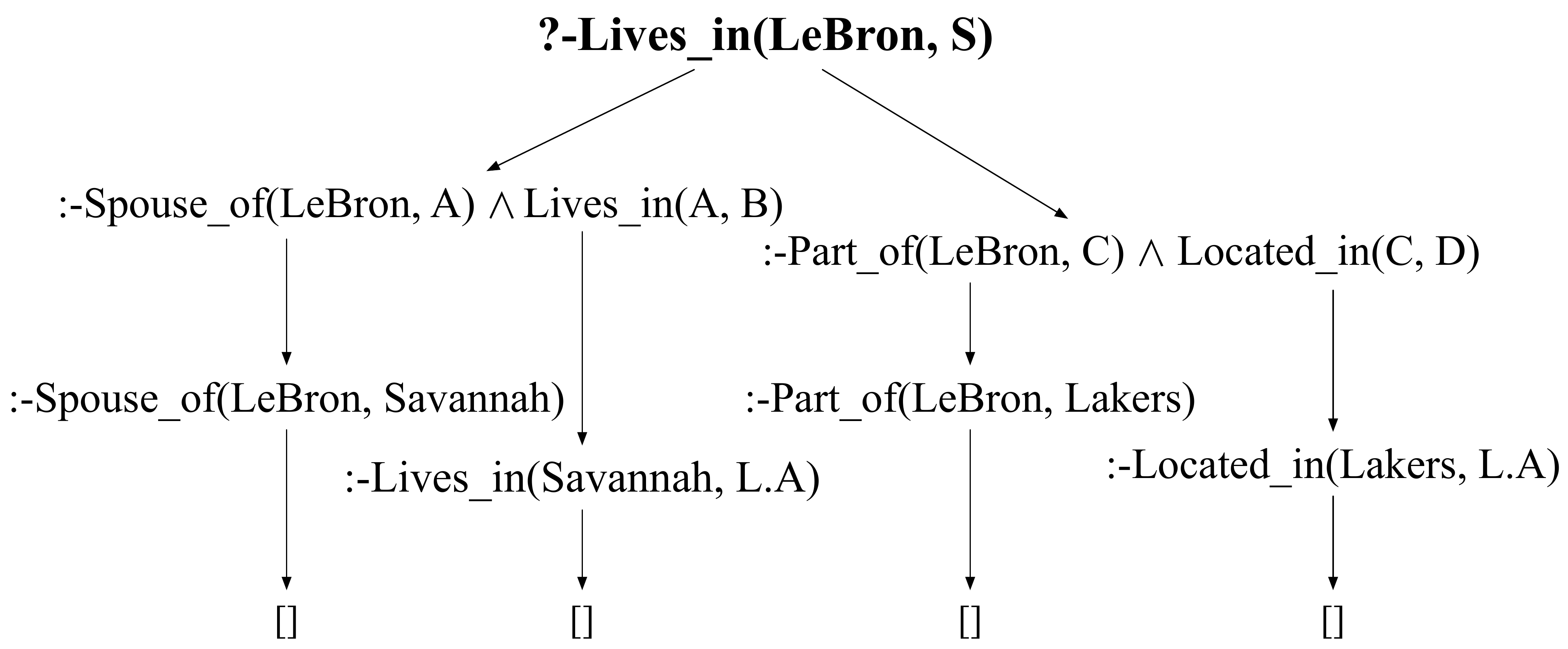}
	\caption{\label{fig:sld}An example of the SLD-tree built for the query Lives\_in(LeBron, S)~\cite{cussens2001parameter}.}
\end{figure}

To calculate the success probability $P(q|L)$, a trivial way is to enumerate all the possible logic rules $L$ with their instantiations. Obviously, it is highly inefficient in real-life applications. ProbLog solves this problem by constructing a proving tree for the target query $q$ according to Prolog's Selective Linear Definite (SLD) resolution. The standard SLD-resolution constructs the SLD-tree in a top-down manner, as shown in~\figref{fig:sld}. It first initializes the root node by the query, and then recursively creates subgoals by applying each clause with its instantiations. The iteration stops when reaching the end conditions, for example, the subgoal is empty, which means a possible answer path is found or the maximal tree depth is reached. As a result, each possible answer path is associated with a set of clauses $\{p_1: c_1, ..., p_n: c_n\} \subseteq T$. 

\hide{
In \figref{fig:sld}, a set of clauses for explaining the given query is:

\beqn{
	\nonumber	
	\text{0.8}:&&\!\!\!\!\!\text{student\_of}(X, Y) \leftarrow \text{follows}(X,Z) \land \text{teachers}(Y,Z)  \\ \nonumber	
	\text{0.5}:&&\!\!\!\!\!\text{teaches}(\text{peter}, \text{computer\_science}) \\ \nonumber
	\text{1.0}:&&\!\!\!\!\!\text{follows}(\text{paul}, \text{computer\_science})	\nonumber
}
}

Then following Eq.\eqref{eq:p_q_T} to Eq.\eqref{eq:p_L_T}, the success probability of a single answer path can be easily computed. A binary decision diagram (BDD)~(\cite{bryant1986graph, flach1994simply}) is further applied to compute the success probability for multi-paths. SLD-tree is also the base of other similar methods such as stochastic logic programs (SLPs)~\cite{cussens2001parameter} and Programming with Personalized PageRank (ProPPR)~\cite{wang2013programming}.

%The standard SLD-resolution constructs the SLD-graph in a top-down manner, as shown in~\figref{fig:sld}. It first initialize the root vertex of the SLD-graph for the query $?-l_1, ..., l_n$ need to be derived, like $likes(john, tom)$ shown in~\figref{fig:sld}, and then recursively create subgoals in the form of $?-b_1\theta, ..., b_m\theta, l_2\theta, ..., l_n\theta$ for each clauses/rules $h\leftarrow b_1, ..., b_m$ in the logic rules. $\theta$ is the substitution unifying $l_1$ and $h$, for example, if $h=friendof(A, tom)$ and $\theta = \{A = mary\}$, then $h\theta = friendof(mary, tom)$. The iteration process terminated until reaching the end-conditions, such as, the subgoal is empty, which means that there has found a successful proof, or exceeding the pre-defined max-depth limitations. As a consequence, each successful proof has a set of associated clauses $\{p_1: h_1, ..., p_n: h_n\} \subseteq T$, then the success probability of single proof is  $\prod_{h_i \in R}p_i$. In ProbLog, a BDDs is further applied to compute the success probability of multi-proofs.

Both MLN and ProbLog learn the probabilities for rules, where MLN builds a global probabilistic graph for all the rules and learns the probabilities for all the rules simultaneously, but ProbLog constructs a local SLD-tree for each query and learns the probabilities for the causes that can support the target query. Other similar methods, such as probabilistic Datalog~\cite{fuhr1995probabilistic}, MarkoViews~\cite{jha2012probabilistic}, stochastic logic programs (SLPs)~\cite{cussens2001parameter}, also attach probabilities to clauses, but having variant optimization frameworks when updating these probabilities. 
%For example, in Datalog, the success probability of each query is also decomposed into the summation of all possible clause sets, where each clause set is attached with a probability/weight. 
For example, SLPs define a randomized procedure for traversing the SLD-tree, where the probability distribution defined over nodes is learned by up weighting the desired answer clauses and down weighting the others.
%In practise, the size of the grounding graph or tree may be extremely large. For example, the number of the grounding atoms in a MLN is always linear with the number of triplets/facts contained in KGs. Specifically, the space complexity is $O(n^k)$, where $n$ is the number of entities in KGs and $k$ is the maximal arity of a predicate. It is infeasible to implement the above methods on real large KGs. To tackle this, ProPPR~\cite{wang2013programming} proposes a local grounding strategy which only grounds a small subgraph from the full grounding graph.
Programming with Personalized PageRank (ProPPR)~\cite{wang2013programming} is an extension to SLPs which changes the randomized sampling to a bias-based strategy based on personalized PageRank (PPR)~\cite{chakrabarti2007dynamic, page1999pagerank}. They use PPR to calculate the probability for each clause based on some pre-defined features, instead of directly setting a probability in Eq.\eqref{eq:p_L_T}.

\subsubsection{Neural-driven Symbolic Reasoning}
\label{sec:neural_driven_symbolic_reasoning}

Neural-driven symbolic reasoning aims to derive the logic rules, where the neural networks are incorporated to deal with the uncertainty and the ambiguity of data, and also reduce the search space in symbolic reasoning. The basic idea is to extend the multi-hop neighbors around the head entity and then predict the answers included in these neighbors. We further divide neural-driven symbolic reasoning methods into path-, graph- and matrix-based framework according to the number of the extended neighbors in each step.

\vpara{Path-based Reasoning.}
\label{sec:path-based-reasoning}
Path-based reasoning extends only one neighbor at each step.
Path-Ranking Algorithm (PRA)~\cite{lao2010relational} is a state-of-the-art path-based method. Given a head entity $h$ and a tail entity $t$, PRA obtains the paths of length $l$ from $h$ to $t$ by performing a random walk with restart algorithm of $l$ steps and then calculates the score $s_p(h, t)$ of the entity pair $(h, t)$ following the path $p$. Finally, PRA estimates the weights of different paths by a linear regression model with $s_p(h, t)$ of different paths as the corresponding feature values. Essentially, PRA is pure symbolic reasoning. We place PRA in this section because most of the following neural-driven symbolic reasonings are based on the idea proposed by PRA. 

PRA is highly dependent on the connectivity of KGs because PRA cannot predict the relation between the nodes that are disconnected from each other. To deal with the problem, Lao et al.~\cite{lao-etal-2012-reading} performs PRA on the graph which consists of both the edge from the original KGs and the syntactic dependency edges extracted from the dependency-parsed web text. Instead of using the above-unlexicalized dependencies, Gardner et al.~\cite{gardner-etal-2013-improving} use the lexicalized words extracted from the corpus to build additional edges, which is more expressive. However, all the above frameworks need to train a model for each relation type and cannot be generalized to other unobserved relations in the test set.

Based on the set of paths connecting the head and tail entities found by PRA, Neelakantan et al.~\cite{Neelakantan2015Compositional} aims to find the most confident path for the entity pair, which is the one with the highest similarity between the embedding of the path and the embedding of the target relation. The embedding of a path is computed by applying the composition function based on the embeddings of the relations included in the path recursively following RNN. Finally, the predictive score of a fact/triplet is represented as the dot product between the embedding of the most confident path and the embedding of the relation. The incorporation of RNN improves the generalization of the model, which can be used to deal with the new relation types that are unobserved in the training data.
Instead of only using the most confident path to predict the score of a triplet, Chain-of-Reasoning~\cite{Das2017chain} combines the similarities of different paths by different strategies such as Top-k, Average, and LogSumExp. 

However, the paths are traversed heuristically, which lacks evaluation in the above methods. 
Meanwhile, when the number of relations in KGs is large, %even the number of paths at most three steps will easily reach millions, 
even with the constraint that paths are at most three steps, the number of finding paths will easily reach millions, 
so it is impractical to enumerate all possible relation paths. 
Thus, DeepPath~\cite{Xiong2017deeppath} proposes a reinforcement learning (RL) method to evaluate the sampled paths, which can reduce the search space. 
The basic idea is to frame the multi-hop reasoning as a Markov Decision Process (MDP) and design a policy function to encode the continuous state of the RL agent. 
The agent reasons in the vector space environment by sampling a relation at each hop to extend the reasoning path.  The state is composed of the embeddings of the current entity and the target entity. The reward function is designed to measure the accuracy, efficiency, and diversity of the path, which will supervise the sampling action at each hop. Meanwhile, to tackle the problem of large action space, DeepPath is initialized by supervised training, where paths traversed by a two-way BFS algorithm are to guide the RL agent.
AnyBRUL~\cite{meilicke2020AnyBURL} also uses RL to sample paths. Additionally, after obtaining the sampled paths, it constructs the ground rules from the sampled paths and generalizes them to abstract rules according to a bottom-up approach.

DeepPath~\cite{Xiong2017deeppath} and  AnyBRUL~\cite{meilicke2020AnyBURL}  require to first sample all the paths between the head and the tail entities, and then leverage them to evaluate whether the tail entity can be the right answer, so, it cannot be directly applied to the scenario where the tail entities are unobserved. Thus, instead of  adopting RL to infer paths, some methods choose to train RL directly to obtain the correct answer entity by the given head entity and the query relation. Among these models, MINERVA~\cite{Das2017MINERVA} is a representative model that samples each neighbor according to a LSTM-based policy function. Different from DeepPath, the state in MINERVA is composed of the embedding of the query relation and the partial path, and the embedding of the answer entity is not needed during the sampling process. These allow MINERVA to derive the answers directly. MINERVA also incorporates a self-relation for each entity to indicate the stop action when the entity itself is sampled. A hard reward 0/1 indicating if the sampled entity at the last step is the correct answer is used to supervise the sampling process.

Instead of using the hard 0/1 reward, Multi-Hop~\cite{Lin2018multi} proposes a soft reward based on the similarity between the true answer entity and the sampled entity at the last step. In addition, inspired by the dropout technique, the model masks some actions by chance during training to avoid choosing lots of repeated paths and alleviates the problem of overfitting. CPL~\cite{fu2019CPL} leverages the text information in addition to the graph structure of the KGs during the sampling process. Specifically, when determining the next-hop entity,  it not only considers the entities in KGs, but also leverages the entities extracted from the text corpus. M-walk~\cite{shen2018mwalk} introduces a value-based RL method and uses Monte Carlo Tree Search (MCTS) to overcome the challenge of sparse positive rewards. Specifically, it adopts an MCTS trajectory-generation step and a policy-improvement step iteratively to refine the policy function, which achieves more positive rewards. DIVA~\cite{chen2018diva} frames the KGC task into a unified model which consists of the components of path finding and answer reasoning, where the paths are modeled as hidden variables and VAE~\cite{Diederik-2013} is adopted to solve the model. Compared with DIVA, PRA, and Chain-of-Reasoning, DeepPath and AnyBRUL emphasize the process of path finding to explicitly derive rules and leave the answer reasoning process to an additional step, while MINERVA, Multi-hop, and M-walk directly reason the answers without explicitly deriving the rules. 

\begin{figure}[t]
	\centering
	\includegraphics[width=0.35\textwidth]{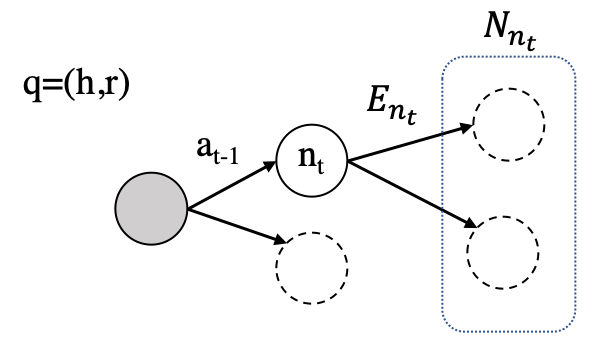}
	\caption{\label{fig:RL} Illustration of the RL-based reasoning~\cite{shen2018mwalk}.}
\end{figure}

Figure~\ref{fig:RL} summarizes the idea of the path-based reasoning process, where $q=(h,r)$ indicates the head entity $h$ and the query relation $r$, $n_t$ represents the current entity, $a_{t-1}$ is the last action that results in the current entity $n_t$, $E_{n_t}$ and $N_{n_t}$ are the set of the relations and the corresponding neighbors of $n_t$, from which the next relation and entity are selected.

\vpara{Graph-based Reasoning.}
\label{sec:graph_based_reasoning}
Graph-based reasoning methods are the extensions to the path-based reasoning methods, which could better explain reasoning in KGs by structuring explanations as a graph rather than a path. 

GraIL~\cite{teru2019inductive} is a graph neural network performed on an extracted subgraph to reason the relation between two entities. Specifically, to validate a triplet $(h,r,t)$, GraIL first extracts the subgraph around $h$ and $t$ as the intersection of the $k$-hop neighbors of the two entities and the relations among them. Then it labels the entities in the extracted subgraph with tuple $(d(i, h), d(i, t))$, where $i$ denotes the $i$-th node in the subgraph and $d(u,v)$ denotes the shortest distance between $u$ and $v$. Next, it adopts an attention-based multi-relational GNN model, i.e., R-GCN~\cite{schlichtkrull2018modeling}, to compute the embedding for each node in the subgraph. And finally, the model concatenates the embeddings of the subgraph (the average of the embeddings of all the entities in the subgraph), the head entity, the tail entity, and the query relation, based on which it scores the correctness of the given triplet $(h,r,t)$. GraIL reasons over local subgraph structures and has a strong inductive bias to learn entity-independent relational semantics.  

Starting from the given head entity and the query relation represented by a pair of head and tail entity, CogGraph~\cite{Du-2019} extends multiple entities at each step by a policy function, and then proposes a GNN model to embed each node in the extended subgraph. CogGraph predicts the answers based on their embeddings.
DPMPN~\cite{Xu2020DPMPN} proposes two GNN models to perform the reasoning. The first GNN model performs the input-invariant message passing globally on the whole KG, which can provide raw and rich representations for entities. The second GNN model performs pruned message passing locally on a subgraph related to the query, which captures input-dependent semantics, disentangled from the full graph.

\begin{figure}[t]
	\centering
	\includegraphics[width=0.47\textwidth]{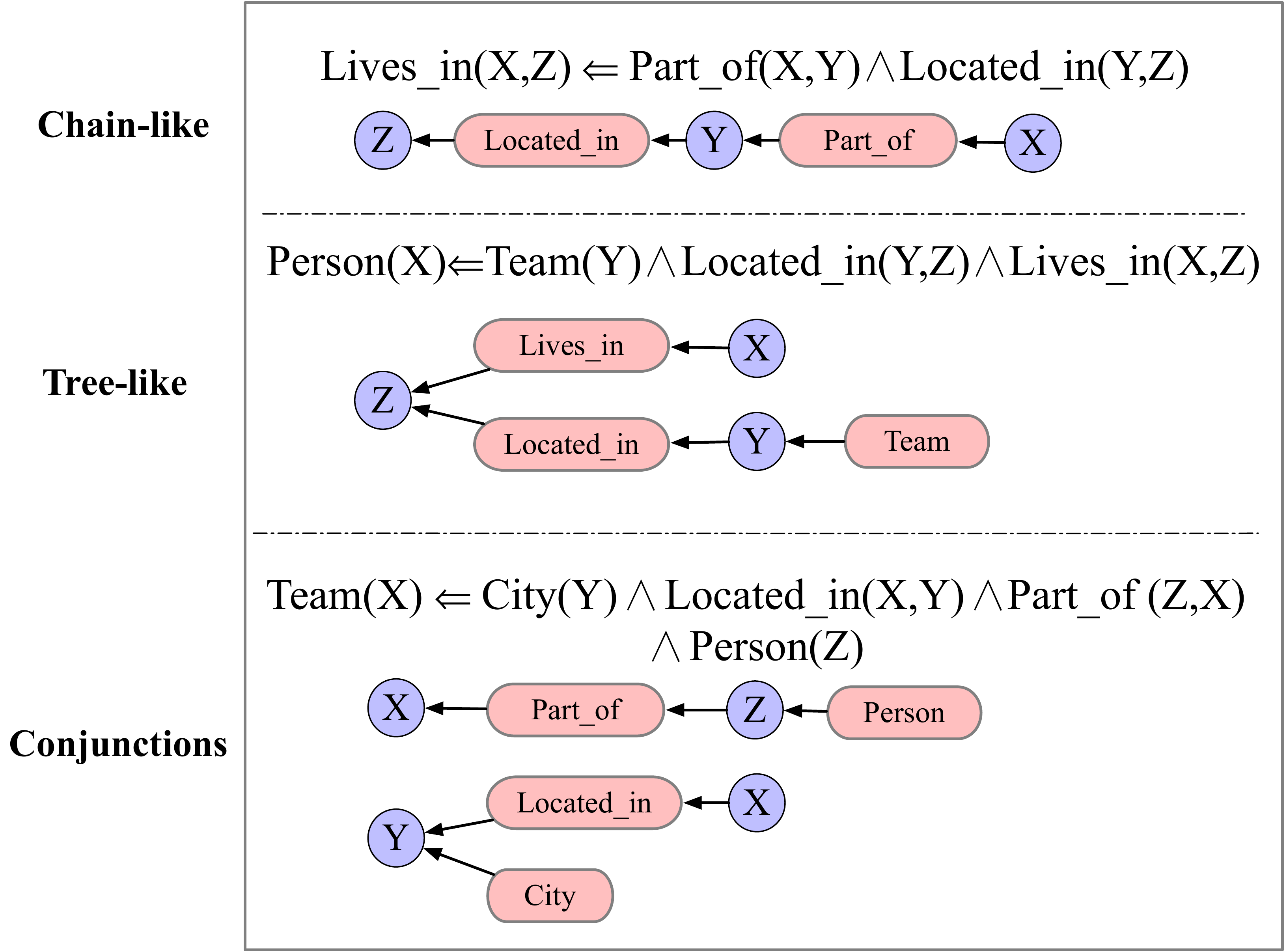}
	\caption{\label{fig:nlil_rules} Examples of chain-like, tree-like and conjunctions of rules. Refer to~\cite{cohen2016tensorlog}.}
\end{figure}

\vpara{Matrix-based Reasoning.}
\label{sec:matrix_based_reasoning}
Matrix-based reasoning can be viewed as an extension to graph-based reasoning, which does not select the neighbors at each hop but incorporates the soft attention scores to indicate the importance of different neighbors. The basic idea is to express the logical relationships between the head and the tail entities by matrix operations. This section firstly introduces the earliest attempt --- TensorLog~\cite{cohen2016tensorlog}, and then describes three typical models, Neural LP~\cite{yang2017differentiable}, Neural Logic Inductive Learning (NLIL)~\cite{yang2019learn} and Neural-Num-LP~\cite{wang2019differentiable}. Based on Tensorlog, Neural LP further learns the new rules, NLIL has the capacity of expressing complex rules shown in \figref{fig:nlil_rules}, and Neural-Num-LP~\cite{wang2019differentiable} particularly deals with the numerical operations such as comparison, aggregation, and negation among entities.

\textit{TensorLog}~\cite{cohen2016tensorlog} infers the weighted chain-like logical rules for explaining each relation $R$ in KGs. Then based on the inferred rules, given a query $R(x,Y)$ with the query relation $R$ and the head entity $x$, the goal is to retrieve a ranked list of entities such that the ground truth answer $y$ is ranked as high as possible.

In TensorLog, each entity in KGs is represented as a one-hot embedding, and each relation 
$R$ in KGs is represented as a matrix $M_R$, where each element $M_R[i,j] =1$ if the fact $R(i,j)$ is in the KGs. Then given a rule $\gamma$ such as $R(X,Y) \rightarrow P(X,Z) \land Q(Z,Y)$ and a head entity $x$, the logical inference of the answers can be implemented by performing matrix multiplications $M_P \cdot M_Q \cdot v_x$. The result is a  vector with each non-zero element $y$ indicating that there exists $z$ such that $P(y,z)$ and $Q(z,x)$ are in the KGs. Since a query relation may be explained by multiple rules, the score of the query relation is calculated by combining all the rules:

\beqn{
	\label{eq:tensorlog_query_relation}
	\sum_{\gamma}\alpha_{\gamma} \prod_{k\in \beta_{\gamma}}\textbf{M}_{R_k},
}

\noindent where $\gamma$ indexes over all possible rules, $\alpha_{\gamma}$ is the confidence associated with the rule $\gamma$, and $\beta_{\gamma}$ is an ordered list of all predicates in the rule $\gamma$. During inference, given a head entity $x$, the score of each retrieved answer equals to the entries in the vector $\textbf{s}$:

\beqn{
	\label{eq:tensorlog_answer}
	 \textbf{s} &=& \sum_{\gamma}(\alpha_{\gamma}(\prod_{k\in \beta_{\gamma}}\textbf{M}_{R_k}\textbf{v}_x)).
}

Then the probability of each true answer $y$ satisfying $R(x,y)$ given the head entity $x$ is maximized:

\beqn{
	\label{eq:tensorlog_objective_function}
	&&\max_{\{\alpha_{\gamma},\beta_{\gamma}\}} \sum_{{x,y}} \text{score}(y|x) = \\ \nonumber
	&&\max_{\{\alpha_{\gamma},\beta_{\gamma}\}} \sum_{{x,y}} \textbf{v}_y^T \left( \sum_{\gamma}\alpha_{\gamma} (\prod_{k\in \beta_{\gamma}}\textbf{M}_{R_k}\textbf{v}_x ) \right)
}

\noindent where $\textbf{v}_x$ and $\textbf{v}_y$ indicate the one-hot vectors of the head entity $x$ and the ground truth answer $y$, respectively. 

To learn the parameters $\alpha$ and $\beta$, 
TensorLog formalizes each rule into a factor graph, 
where each node in the graph represents a variable in the rule and each edge  indicates a predicate/relation in the rule. 
Given a factor graph, or a rule that explains the query $R(x,Y)$, the probability of an answer $y$ to the query, i.e., $R(x,y)$, can be defined as the joint probability of all the possible grounding entities in the graph with $x$ and $y$ being fixed. 
The approximation algorithm belief propagation~\cite{2015Approximation} is adopted to compute the probability. 

TensorLog estimates the rule confidence in a differentiable matrix-production manner. 
However, it is unable to generate new logic rules. 

\textit{Neural Logic Programming (Neural LP)}~\cite{yang2017differentiable} improves the matrix-based reasoning framework based on TensorLog. 
In TensorLog, learning parameters is difficult because each rule is associated with a parameter, and enumerating rules is an inherently discrete task. To overcome this difficulty, Neural LP interchanges the summation and the product  in Eq.\eqref{eq:tensorlog_query_relation} and decomposes the weight of a rule into the weights of the predicates in the rule. The concrete formula to score a query relation is:

\beq{
\label{eq:nerualp}
\prod_{t=1}^{T}\sum_{k}^{|R|}a_t^k\textbf{M}_{R_k},
}

\noindent where $T$ is the maximal length of the rule and $|R|$ is the number of the predicates in the KGs. However, the expressive capability of Eq.\eqref{eq:nerualp} is not sufficient enough, as it assumes that all the rules are of the same length $T$. To address this issue, Neural LP designs a recurrent formulation to model the length of rules dynamically. It is defined as follows:

\beqn{
	\label{eq:neural_LP}
	\textbf{u}_0 &=& \textbf{v}_x \\ \nonumber
	\textbf{u}_t &=& \sum_{k}^{|R|}a_t^k\textbf{M}_{R_k}(\sum_{\tau=0}^{t-1}b_t^{\tau}\textbf{u}_{\tau})\quad \text{for}~1\leq t\leq T \\\nonumber
	\textbf{u}_{T+1}&=&\sum_{\tau = 0}^{T}b_{T+1}^\tau \textbf{u}_{\tau}
}

\noindent where $\textbf{u}$ is an auxiliary memory vector initialized as the input entity $\textbf{v}_x$. At each step, the model first computes a weighted average of previous memory vectors using the memory attention vector $b_t$, and then ``softly" applies the TensorLog operations using  operation attention vector $a_t$. Finally, it computes a weighted average of all the memory vectors and uses attention to control the length of rules.

TensorLog and Neural LP aim at learning some new probabilistic chain-like logic rules for knowledge bases reasoning, but they fail to infer some complex formations of rules such as tree-like, conjunctions, etc. (Refer to the examples in Figure~\ref{fig:nlil_rules}). Moreover, the rules are inferred based on the specific head entity $x$, which deteriorates the generalization of the learned rules.

\textit{Neural Logic Inductive Learning (NLIL)}~\cite{yang2019learn} tackles the non-chain-like rules by incorporating a primitive statement. Specifically, the concept of a primitive statement is an extension to the concept of atom. An atom is defined as a predicate applied to the logic variables, while a primitive statement is defined as a predicate applied to the logic variables or the results of some operators. Here, an operator is defined upon a predicate:

\begin{equation}
\left\{
\begin{array}{lr}
	\varphi_k() = \textbf{M}_k \textbf{1}    &     \text{           if          }  k \in \mathcal{U}, \\
	\varphi_k(\textbf{v}_x) = \textbf{M}_k \textbf{v}_x    &  \text{            if            }  k \in \mathcal{B},
\end{array}
\right.
\end{equation}

\noindent where $\mathcal{U}$ and $\mathcal{B}$ are the sets of unary and binary predicates respectively. The operator of the unary predicates takes no input and is parameterized with a diagonal matrix. For example, for the unary predicate such as Person(X), its operator $\varphi_{Person}() = \textbf{M}_{Person} \textbf{1}$ returns the set of all entities labelled as person. The operator of the binary predicates returns the tail entities that, together with the head entity, i.e., the input of the operator, satisfy the predicate $P_k$. For example, for the binary predicate such as Mother(X,Y), its operator $\varphi_{Mother}(\textbf{v}_x)$ returns the mothers of the input variable $x$.

Based on the definition of the operator, the primitive statement can be instantiated as follows:

\begin{equation}
\label{eq:primitive_statement}
\psi_k(x,y)\!\!\!=\!\!\!
\left\{
\begin{array}{lr}
\!\!\!\sigma((\textbf{M}_k \textbf{1})^{\text{T}} (\prod_{t'=1}^{T'} \textbf{M}^{(t')} \textbf{v}_{y}))    &     \text{           if          }  k \in \mathcal{U}, \\
\!\!\!\sigma( (\prod_{t=1}^{T} \textbf{M}^{(t)} \textbf{v}_{x})^{\text{T}} (\prod_{t'=1}^{T'} \textbf{M}^{(t')} \textbf{v}_{y}))    &  \text{            if            }  k \in \mathcal{B},
\end{array}
\right.
\end{equation}

\noindent where $\sigma$ is the sigmoid function. In Eq.\eqref{eq:primitive_statement}, unary primitive statements and binary primitive statements are instantiated in different ways. Unary primitive statements contain only one relation path starting from variable $y$, while binary primitive statements contain two relation paths starting from variable $x$ and $y$ respectively. Compared with the multiplication of the predicates in a single relation path in Eq.\eqref{eq:tensorlog_objective_function}, Eq.\eqref{eq:primitive_statement} replaces the answer vector $\textbf{v}_y$ with another relation path, which makes it possible to represent ``correlations" between two variables, and the path that starts from the unary operator, e.g., $\varphi_{Person}$. In this way, a primitive statement is capable of representing the tree-like logical rules, as shown in Figure~\ref{fig:nlil_rules}. Then similar to Eq.\eqref{eq:neural_LP}, Eq.\eqref{eq:primitive_statement} can also be relaxed into weighted sums. Instead of assigning a single path attention vector for all the relation paths in Eq.\eqref{eq:tensorlog_objective_function}, NLIL assigns separate attention vectors for each relation path in the $k$-th primitive statement. 

Then the logic combinations of primitive statements, via $\{\land, \lor, \neg\}$, as shown in Figure~\ref{fig:nlil_rules}, are represented by the following equations:

\beqn{
	\label{eq:initial_fomula}
	\mathcal{F}_0 \!\!\!&=&\!\!\! \Phi,\\
	\label{eq:negation} 
	\hat{\mathcal{F}}_{l-1} \!\!\!&=&\!\!\!  \mathcal{F}_{l-1} \cup \{1-f(x,y) : f \in \mathcal{F}_{l-1}\}, \\
	\label{eq:or}
	\mathcal{F} \!\!\!&=&\!\!\!\{f_i(x,y) * f_i'(x,y): f_i, f_i' \in \hat{\mathcal{F}}_{l-1}\}_{i=1}^2,
}

\noindent where each element in the formula set $\{f:f\in\mathcal{F}_l\}$ is called a formula which accepts two entities as input and outputs a scalar. The initial formula set contains only primitive statements $\Psi$. The $(l-1)$-th formula set $\mathcal{F}_{l-1}$ is concatenated with its logic negation to yield $\hat{\mathcal{F}}_{l-1}$. Then each formula in the next level is the logic $\land$ of two formulas from $\hat{\mathcal{F}}_{l-1}$, as the logic $\lor$ can be implicitly represented as $p \lor q = \neg (\neg p \land \neg q)$. Similarly, the formula selection can be parameterized into the weighted-sum form with the attentions. 

In practice, the above different attention vectors are learned by three stacked transformer networks. This complex and compact framework has the capacity of discovering more expressive underlying reasoning patterns.

All reasoning models discussed above only focus on the relational structures of KGs but ignore the numerical values of entities that may be involved in the reasoned rules.

\textit{Neural-Num-LP}~\cite{wang2019differentiable} extends Neural LP~\cite{yang2017differentiable} to learn the numerical rules. It supports the comparison operator by defining it in a matrix format:
	
	\beqn{
		(M_{r_{pq}^\leq})_{ij} &=& \left\{
	\begin{aligned}
		1 \quad&\text{if}~p_i\leq q_j, \\
		0 \quad&\text{otherwise,}
	\end{aligned}
	\right. 
	}

	\noindent where $p,q$ are two numerical features such as ``hasCitation" and ``birthDate". This matrix denotes the binary indicator of the comparison over all pairs of entities in KGs containing the features $p$ and $q$, which can be multiplied with other predicate/relation matrices presented in Eq.\eqref{eq:tensorlog_query_relation}.

\hide{
	\item \textbf{Aggregation operator} defines a general comparisons which is performed not necessarily among two numerical features of a certain entity but rather functions over several features, called aggregations. Formally, the aggregation atom is in the form: 
	
	\beq{
	f\{Y_1, Y_n: p_1(X,Y),p_n(X,Y))\}\circ Z
}

	\noindent where $f$ is an aggregation function that can be arbitrarily complex learnable structures, $\circ \in\{\leq, \ge\}$, $p_1,...,p_n$ are numerical features of an entity $X$, and $Z$ is a numerical value.
	\item \textbf{Negated operator} can be trivially gained by flipping the comparison operator, that is, flipping all zeros to ones and vice versa in the corresponding comparison matrix ${M}_{r_{pq}}$. However, it will result in a dense matrix containing redundant operations when processing matrix productions. To avoid this, Neural-Num-LP employs the local closed-world assumption that only the rows with at least one non-zero element should be flipped, defined as follows:  
	
	\beq{
	\overline{M}_{r_{pq}} = 1_{p}1_{q}^T-{M}_{r_{pq}}
	} 
	
	\noindent E.g., $\overline{M}_{r_{pq}^{\leq}} = 1_{p}1_{q}^T-{M}_{r_{pq}^{\geq}}$.
}

Neural-Num-LP is a good inspiration for fully understanding the reasoning patterns not only from the relations but also from the attributes.

\vpara{Summary.}
The first kind of neural-symbolic reasoning, i.e., the symbolic-driven neural reasoning, aims to learn the entity and relation embeddings. The logic rules are used to increase the number of highly confident triplets to improve embedding performance. Thus the reasoning process is still based on embeddings, which lacks interpretation. 

The second kind, i.e., the symbolic-driven probabilistic reasoning, qualifies the logic rules by grounding the rules in KGs. With the increase of the entities and relations in KGs, the grounding atoms/rules will increase dramatically, making inference and learning computationally expensive. Besides, these methods cannot produce new rules.

The above two  kinds of methods  take the answer prediction as the only target. The difference is that in symbolic-driven probabilistic reasoning, the rules are used as the features to predict the answers, while in  symbolic-driven neural reasoning, the rules are used to  generate more facts for learning high-quality embeddings. 

The third kind, i.e., the neural-driven symbolic reasoning, takes both the answer prediction and the rule learning as the target. To achieve the goal,  it infers the answers following the paths, graphs, or matrixes started from the head entity, which can enhance  the interpretability of the predicted answers. However, with the increase of the hop number, the paths, subgraphs, or the matrix multiplication become more complex, making the predictive performance more sensitive to the sparsity of the  knowledge graphs.

%Specifically, it also defines comparison operators on numerical values in the format of the matrix, like equation*. However, different from the sparse logic-predicates matrix, the matrix defined over numerical comparisons is dense, which will exceed the typical GPU memory. To deal with it,  ~\cite{wang2019differentiable} transfers dense matrix-multiplication to cumulative sum (sumcum) operations, which greatly reduces complexity. 
%The negation operations can be obtained by naively flipping all zeros to ones, but it also results in a dense matrix that is not supported by matrix production. To compute the negated operator, it employs the local closed-world assumption that only the rows with at least one non-zero element should be flipped. 
%What's more, all the rules learned above can be viewed as an instance of learning rules with aggregates in the knowledge bases, like equation [*]. A random neural network can be applied to learn the specific aggregation functions based on the rules contained. \zj{I cannot understand the last two sentences.}

\section{Reasoning for KGQA}
\label{sec:reasoning_4_QA}
%With the rise and development of large-scale knowledge graphs(KG), question answering over knowledge graphs(QA-KG) which uses plain natural language questions to query structured knowledge bases has become a research hotspot, because it provides ordinary users intuitive and easy way to capture the knowledge they exactly want over large-scale KGs without knowing the structure of them. However, understanding the natural language questions as well as the incompleteness of KGs pose great challenges on QA-KG. 

In this section, we will introduce the reasoning methods for KGQA, which can also be categorized into the neural, the symbolic and the neural-symbolic reasoning methods. 
KGQA needs to deal with the natural language questions, which is more complex than reasoning for KGC. 
Thus according to the types of the questions, we can further categorize KGQA into simple-relation questions, multi-hop relation questions and complex-logic questions.

\begin{figure}[t]
	\centering
	\includegraphics[width=0.47\textwidth]{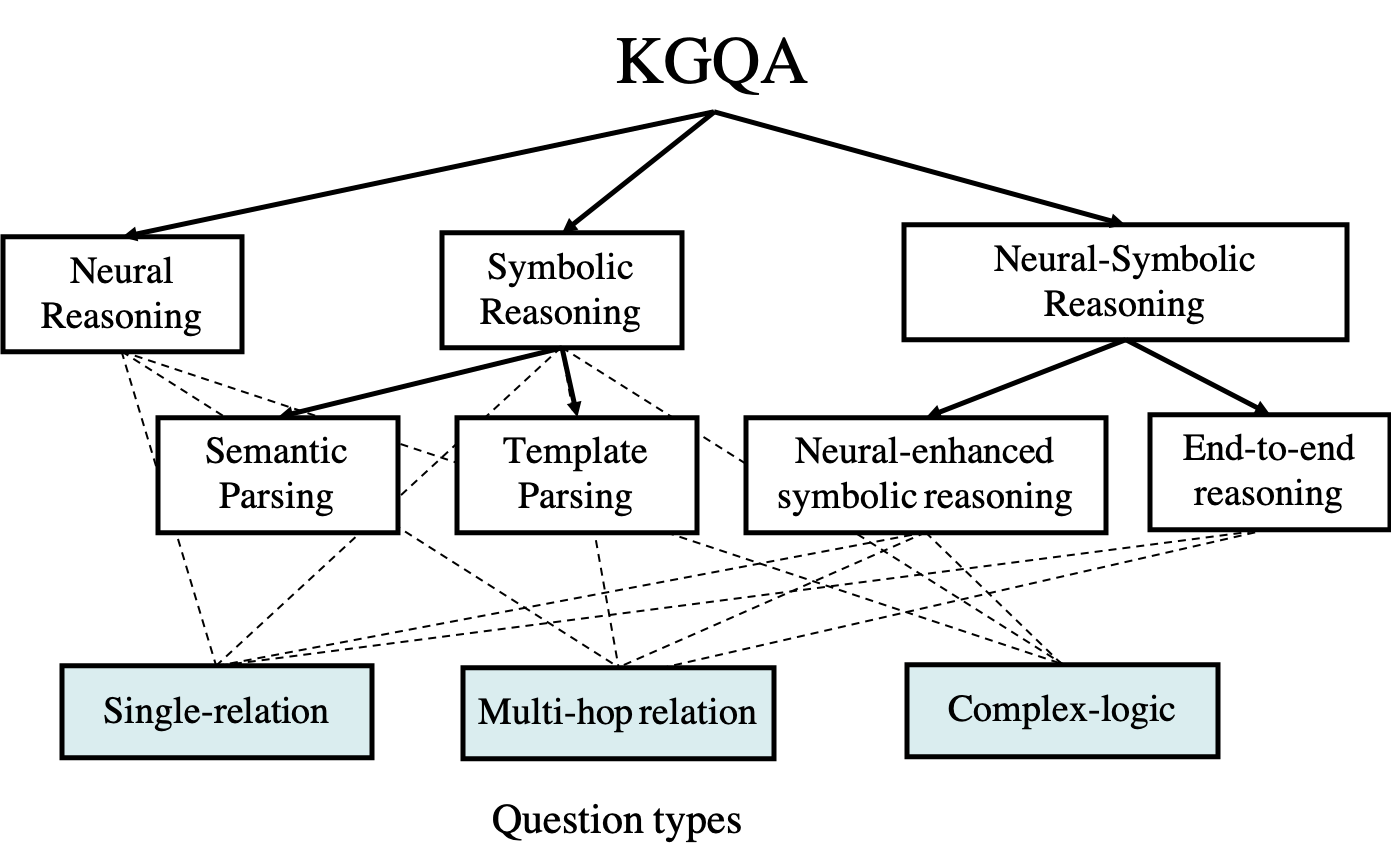} 
	\caption{\label{fig:kgqa-framework} The taxonomy of the KGQA reasoning methods. The dashlines link the question types to the methods that can address them.}
\end{figure}

Single-relation questions refer to questions that only involve a single topic entity and a single relation in KGs. Then the tail entities in KGs corresponding to the topic entity and the relation are extracted as the answers. 
%Thus, generally, the key problem for solving single-relation question answering is to locate the right topic entity and recognise the correct relation. 
Example questions of this type include: ``Who is the wife of Barack Obama" or ``Where is the Forbidden City".
Multi-hop relation questions are path-based which means the answer can be found by walking along a path consisting of multiple intermediate relations and entities starting from the topic entity. 
%The challenges lie in the demand for multi-hop reasoning which indicates there might be multiple missing links because of the incompleteness and sparseness of KGs. A lot of works have been done on this kind of complex questions, which can be roughly classified into two types: template-based methods and embedding-based models.
Complex-logic questions contain several subject entities aggregated by conjunction ($\cap$), disjunction ($\cup$) or logical negation ($\neg$), which means the answer can be obtained by some operations, such as intersection of the results from multiple path queries (multi-hop questions). Questions may also have some complex constraints (Cf. Table~\ref{tb:constraints} for details).
%Similarly, existing studies on complex logical queries can be roughly divided into two groups: template-based methods and embedding-based models. However, unlike the methods of multi-hop QA, the reasoning framework for complex logical queries usually bases on the dependency graph which is a DAG (Directed Acyclic Graph).
Figure~\ref{fig:kgqa-framework} presents the taxonomy of the KGQA reasoning methods and the links between the question types and the methods that can address them.

\begin{table*}[t]
	\newcolumntype{?}{!{ \hspace{0.5pt} \vrule width 1pt \hspace{-2pt}}}
	\newcolumntype{k}{!{\hspace{-5pt}}}
	\newcolumntype{C}{>{\centering\arraybackslash}p{2em}}
	\centering \scriptsize
	\renewcommand\arraystretch{1.0}
	
		\caption{
		\label{tb:overall_KGQA} A summary of knowledge graph question answering and recent advances. S: single-relation question, M: multi-hop relation question, C: complex-logic question.	
	}

	\begin{tabular}{ccclll}
		
		\toprule
		
		Category & Sub-category & Question Type &  Model  &       Mechanism \\
		\midrule
		
		\multirow{6}{*}{\tabincell{c}{Neural \\ Reasoning}} 
		& \multirow{6}{*}{\tabincell{c}{-}} 
		&S &  KEQA~\cite{huang2019knowledge}  &  train a topic entity learning model and  query relation learning model\\
		&&S &  Liu et al.~\cite{lukovnikov2017neural}  &  train an end-to-end topic entity and  query relation learning model\\
		&&M &  EmbedKGQA\cite{Saxena2020ImprovingMQ}  &  use RoBERTa to embed the question\\
		&&C&  GQE~\cite{Hamilton2018EmbeddingLQ}   &  deal with conjunctive logic queries \\
		&&C&  QUERY2BOX~\cite{ren2019query2box}  &   deal with disjunctive queries \\
		&&C&  EmQL~\cite{Sun2020FaithfulEF} & obtain faithful embeddings \\ 
		
		\midrule
		\multirow{6}{*}{\tabincell{c}{Symbolic\\Reasoning}} 
		& \multirow{3}{*}{\tabincell{c}{Semantic\\ Parsing}} 
		&C & Kwiatkowksi et al.~\cite{kwiatkowksi2010inducing}  &  follow CCG to parse questions\\
		&&C& Berant et al.~\cite{berant2013semantic}  &  follow $\lambda$-DCS to parse questions\\
		&&C& Dubey et al.~\cite{dubey2016asknow}  &  follow NLQF to parse questions\\
		\cmidrule{2-5} 
		& \multirow{3}{*}{\tabincell{c}{Template-based \\ Parsing}} 
		&C& UncertainTQA~\cite{Zheng2015HowTB}  &  link query graphs and SPARQLs as templates\\
		&&C& QUINT~\cite{Abujabal2017AutomatedTG}  &  link query graphs and dependency parse trees as templates\\
		&&C& TemplateQA~\cite{Zheng2018QuestionAO}  & link natural language patterns and SPARQLs as templates\\

		\midrule
		\multirow{9}{*}{\tabincell{c}{Neural \\ Symbolic\\Reasoning}} 
		& \multirow{4}{*}{\tabincell{c}{Neural-Symbolic\\ Reasoning}} 
		&S & Yih et al.~\cite{Yih2014}  &  link questions to KG and use CNN to encode the linked entities/relations\\
		&&M& MULTIQUE~\cite{bhutani2020answering}  &  generate a sub query graph and use LSTM to qualify it\\
		&&C& Bao et al.~\cite{bao2016constraint}  &  generate a multi-constraint query graph\\
		&&C& Lan et al.~\cite{lan-jiang-2020-query}  &  incorporate constraints and extend relation paths simultaneously\\

		\cmidrule{2-5} 
		& \multirow{5}{*}{\tabincell{c}{End-to-end \\ Reasoning}} 
		&M& IRN~\cite{zhou2018an}  &  extend a path based on question and entity/relation embeddings\\
		&&M& SRN~\cite{qiu2020stepwise}  &  use RL to learn the path sampling strategy\\
		&&M& Graft-Net~\cite{sun-etal-2018-open}  &extract a subgraph and apply GNN to infer the answer in it\\
		&&M& PullNet~\cite{SunACL2019}  &use RL to learn the subgraph sampling strategy\\
		&&M& VRN~\cite{zhang2018variational}  &model the topic entities as hidden variables\\

		\bottomrule
	\end{tabular}
	\normalsize

\end{table*}

\subsection{Neural Reasoning}
\label{sec:neural_reasoning_for_QA}

Neural reasoning methods for question answering usually encode the entities and relations in KGs as well as the input questions into embeddings of the same space, based on which they infer the answers. Neural reasoning methods can deal with three kinds of questions: the single-relation questions, the multi-hop relation questions and the complex-logic questions.
%Generally, these methods can be divided into question-only methods which directly query the answer by the extracted topic entity and relation from the question, and the end-to-end methods which learn a unified model based on both the questions and the entities in KGs to infer answers. In the following parts, we introduce the two categories of methods which are used for solving single-relation question, multi-hop relation question and conjunctive question respectively. 

\vpara{Single-relation Question.} KEQA~\cite{huang2019knowledge} deals with the single-relation questions by finding several candidate triplets. The tail entity in the candidate triplets is chosen as the answer according to the embeddings of the triplet, the topic entity and the query relation  in the question. Specifically, to find the candidate triplets, KEQA first trains a head entity learning model with Bidirectional LSTM as the key component to predict the entity token and the non-entity token in the question literals. It then extracts the topic entity based on the predicted tokens and searches the entities in KGs which are same to the topic entity or contain the topic entity as the candidate heads. All the triplets with head entity in the candidate heads are named as the candidate triplets. Next, to obtain the embeddings of the topic entity and the query relation, the authors propose another bi-directional LSTM which takes the sequential tokens in the question literals as input to predict the corresponding embedding. Finally, for each candidate triplet, KEQA minimizes the Euclidean distance of the predicted embedding between the candidate and the ground truth, in head entity, predicate, and answer.

Liu et al.~\cite{lukovnikov2017neural} propose a similar end-to-end method to address the single-relation questions. For topic entity and single relation, both word-level information and character-level information from the question are considered. Other researches ~\cite{dai2016cfo,golub2016character,mohammed2017strong,ture2016no,yin2016simple} are the similar neural reasoning methods for single-relation KGQA. 

\hide{

\begin{figure}[t]
	\centering
	\includegraphics[width=0.48\textwidth]{Figures/KEQA}
	\caption{\label{fig:KEQA} Illustration of the neural reasoning method KEQA for solving the single-relation questions~\cite{huang2019knowledge}.}
\end{figure}
}

\begin{figure}[t]
	\centering
	\includegraphics[width=0.48\textwidth]{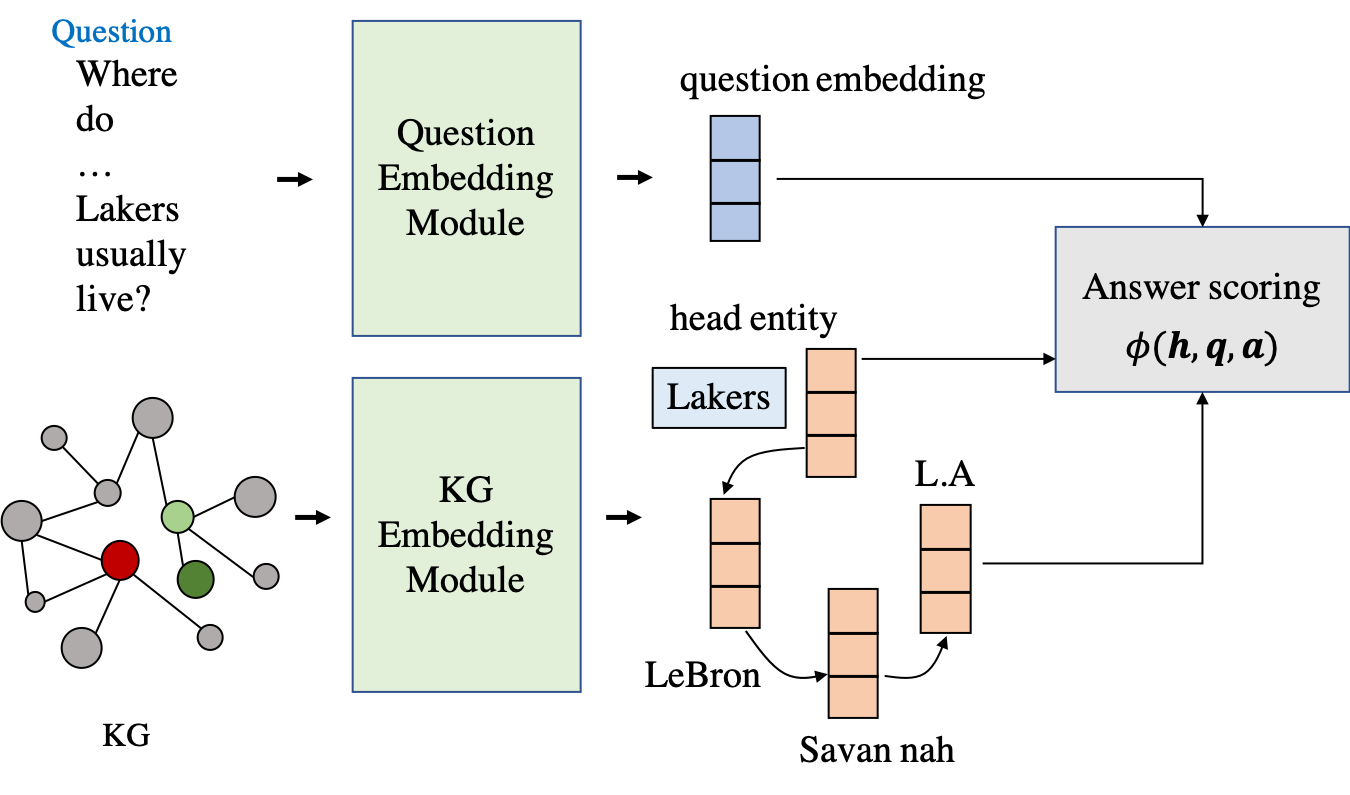}
	\caption{\label{fig:EmbedKGQA} Illustration of the neural reasoning method EmbedKGQA for solving the multi-hop relation questions~\cite{Saxena2020ImprovingMQ}.}
\end{figure}

\vpara{Multi-hop Relation Question.}
In addition to single-relation questions, EmbedKGQA~\cite{Saxena2020ImprovingMQ} is proposed to deal with the multi-hop relation questions. It employs ComplEx~\cite{Trouillon2016ComplexEF} to embed entities and relations in the complex space and applies the same ComplEx scoring function to predict the answers. 
Specifically, the input question $q$ is first embedded by RoBERTa~\cite{Liu-2019}, and then projected by a feed-forward neural network into the complex space. Then a score $\phi(\mathbf{h},\mathbf{q},\mathbf{a})$ is calculated for each triplet of a topic entity, a question and an answer such that $\phi(\mathbf{h},\mathbf{q},\mathbf{a})>0$ if $a$ is an answer entity and $\phi(\mathbf{h},\mathbf{q},\mathbf{a})<0$  for $a$ is not, where $\mathbf{h}$, $\mathbf{q}$ and $\mathbf{a}$ are the embeddings for the topic entity, question and answer respectively. EmbedKGQA selects the entity with the highest score as the answer. Figure~\ref{fig:EmbedKGQA} illustrates the basic idea of EmbedKGQA. Other works such as \cite{Chen2019BidirectionalAM, Dhingra2020DifferentiableRO,Niu2020JointSA} are  similar works for multi-hop KGQA.

\vpara{Complex-logic Question.}
KEQA and EmbedKGQA cannot handle complex logical questions, because these queries involve the logical operations that will result in multiple entities at each hop. To address this defect, some researches represent the logic operations as the learned geometric operations. For example, 
Hamilton et al.~\cite{Hamilton2018EmbeddingLQ} propose a neural reasoning model GQE (i.e., graph query embeddings) to deal with the conjunctive logic queries. On the basis of their work, Q2B (QUERY2BOX) proposed by Ren et al.~\cite{ren2019query2box} fills the gap in disjunctive queries. Both GQE and Q2B assume that a complex logical question can be represented as a DAG. They both start with the embeddings of the topic entities and then iteratively apply their proposed geometric operations to generate the embedding of the query, which are then used to predict the answer entities according to their similarities.

The two proposed geometric operators in GQE are the geometric projection operator $P$ and the geometric intersection operator $I$, where $P$ is responsible for projecting a query embedding $\mathbf{q}$ of the last hop according to the relations of the outgoing neighbors to obtain the embedding $\mathbf{q}'$ of the current hop, and $I$ aggregates the embeddings of all the incoming neighbors of a node in the DAG to simulate the logic conjunction operator. Specifically, $P$ and $I$ are implemented as:

\beqn{
	P(\textbf{q},r) &=& \mathbf{R_rq},\\ \nonumber
	I(\{\textbf{q}_1,...,\textbf{q}_n\}) &=& \mathbf{W_r}\Psi(NN_k(\textbf{q}_i),\forall i=\{1,...,n\}) \nonumber
}

\noindent Where $\mathbf{R_r,W_r}$ are trainable parameter matrices for relation $r$, $NN_k$ is a k-layer feedforward neural network  and $\Psi$ is a symmetric vector function (e.g., an elementwise mean or min of a set over vectors).

However, GQE embeds a question into a single point in the vector space, which is problematic because there are usually multiple immediate entities when traversing the KGs for answering questions. 
And it is not clear how to effectively model a set with a single point. 
Also, it is unnatural to define the logical operation of two points in the vector space. 
Thus, Q2B is proposed to embed a query as a box in which the set of points correspond to the answer entities of the query. 
A box is represented by the embedding of the center and the offset of the box, which models a set of entities whose vectors are inside the box. 
Each relation is associated with a box embedding.
Based on the definition of the box, given an input box embedding $\mathbf{p}$ and a relation embedding $\mathbf{r}$, the geometric projection operator can be simply represented as $\mathbf{p}+\mathbf{r}$, where the centers and the offsets of them are summed respectively. 
Then the intersection of a set of box embeddings is calculated by performing attention over the box centers and shrinking the box offset using the sigmoid function. 
As for the union operation, since boxes can be located anywhere in the vector space, the union of two boxes would no longer be a simple box. 
To rectify this issue, Q2B transforms the query into Disjunctive Normal Form (DNF), i.e. disjunction of conjunctive queries, such that the union operation only appears in the last step. Then a DNF query can be solved by firstly representing each conjunctive query and then simply aggregating their results, for example, taking the nearest point of the boxes of all the conjunctive queries. However, the negation operation is not involved in the above two models. 

Both GQE and Q2B learn geometric operators to simulate logic operators, which, however, are not faithful to deductive reasoning and fail to find entities logically entailed as answers, due to the fact that they casually use spatial operations hoping to achieve logic reasoning.
Thus, EmQL~\cite{Sun2020FaithfulEF} can obtain faithful embeddings by proposing five operators for reasoning, including set intersection, union, difference, relation following and relational filtering.

\vpara{Summary.}
Although the neural reasoning methods can deal with the three kinds of questions, the complex questions with different constraints are not totally solved by the neural reasoning methods (Cf. Table~\ref{tb:constraints} for details). And similar to KG completion, the neural reasoning methods also lack the interpretation for KGQA.

\begin{figure}[t]
	\centering
	\includegraphics[width=0.48\textwidth]{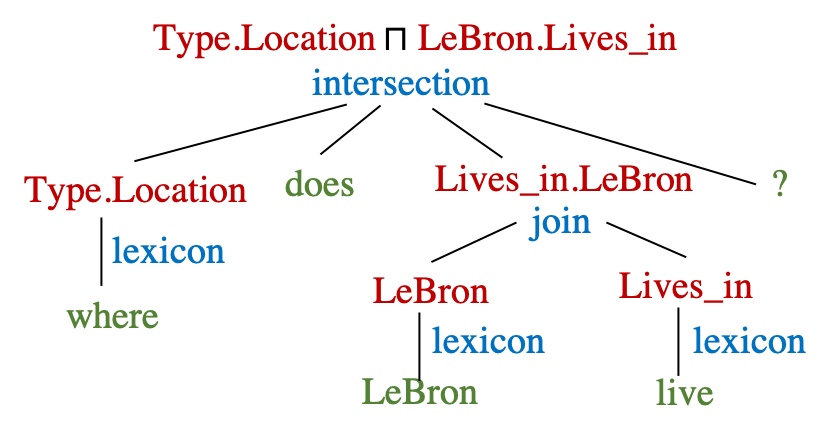}
	\caption{\label{fig:SP} An example of transforming the question ``Where does LeBron live?" into a logic expression following $\lambda$-DCS~\cite{berant2013semantic}.}
\end{figure}

\subsection{Symbolic Reasoning}
\label{sec:symbolic_reasoning_for_QA}

Symbolic reasoning methods for KGQA fall into two major categories --- semantic parsing ones and template-based ones. 
Both aim at generating structured queries for unstructured natural language questions. However, they differ in the way to understand the natural language questions. The semantic parsing methods employ NLP tools to convert the question into the syntactic dependency representation, while the template-based methods use a large number of templates which consist of both the natural language pattern and the corresponding structured query pattern like SPARQL to decompose the complex question.

\vpara{Semantic Parsing.}
Semantic parsing methods target at parsing the input natural language question into a logic expression, based on which the answer to the question can be easily retrieved from the KGs. Different semantic parsing methods transform the natural language questions into different forms of logic expressions. 

For example, Kwiatkowksi et al.~\cite{kwiatkowksi2010inducing} follow combinatory categorial grammar (CCG)~\cite{bos-etal-2004-wide}, a linguistic formalism that couples syntax and semantics, to transform the questions.
For example, for a sentence $x$ ``New York borders Vermont", following the CCG grammar, its corresponding logic expression is Next\_to(ny, vt). 
The target is to learn a function that maps a sentence $x$ to the logic expression $z$, where the function is learned by inducing CCG from the training data of $\{(x,z)\}$.  
%Considering the possible grammars may be too large to enumerate and the ambiguity of possible analyses for each sentence, this work introduce a log-linear CCG parsing model with hidden variable during training to guide the process of grammar refinement and help select the best analysis for each new sentence. 
The whole algorithm consists of two components, where the first one learns the CCG lexicon that is used to define the space of the predicates in the logic expressions, and the second one learns the parameters of the features reflecting the probability of the logic expressions. 

Berant et al.~\cite{berant2013semantic} follow Lambda Dependency-based Compositional Semantics ($\lambda$-DCS)~\cite{Liang2013} to transform the questions, in order to make existential quantification implicit, thus reducing the number of variables. For example, $\lambda x.\exists a.p_1(x,a) \land \exists b.p_2(a,b) \land p_3(b,e) $ is expressed compactly as $p_1.p_2.p_3.e$ by hiding the existential quantifications $\lambda$, $a$ and $b$. 
We give a real example of transforming the question ``Where does LeBron live" into the logic expression following $\lambda$-DCS in Figure~\ref{fig:SP}, where the blue labels indicate the composition rules and the red nodes denote the transformed logic expressions. The whole algorithm also consists of two components, where the first one maps the natural language phrases to the predicates in KGs based on some pre-defined lexicon mappings and a set of composition rules, and the second one is a log-linear model to learn the parameters of the features which reflect the probability of the transformed logic expressions. This method differs from the method presented in~\cite{kwiatkowksi2010inducing} in two ways. First, they propose a  bridging function to deal with the ambiguity of the predicates in natural language questions through generating predicates compatible with the neighboring predicates. Second, the logic expressions $D(x)=\{d\}$ are assumed to be unobserved, and only the question-answer pairs $\{(x_i,y_i)\}$ can be obtained to supervise the logic transformation. Thus, the log-likelihood of the correct answer ($d.z = y_i$), summing over the latent logic expression $d$, is optimized:

\beq{
	\mathcal{O}(\theta)=\sum_{i=1}^n\log\sum_{d\in D(x): d.z=y_i} p_\theta(d|x_i),
}

\noindent where $p_\theta(d|x_i)$ is the probability of generating the logic expression $d$ from the sentence $x_i$ given the parameters $\theta$.

Dubey et al.~\cite{dubey2016asknow} follow natural language query formalization (NLQF) to transform the questions. 
They introduce a chunker-styled pseudo-grammar, named Normalized Query Structure (NQS), to parse each token in the natural language questions. Instead of using static templates, its syntax definition is dynamically fitted to the natural language sentence. For example, the question ``Desserts from which country contain fish." is finally parsed as "[wh = which][R1 = null][D = country][R2 = from][I1 = dessert][R3 = contain][I2 = fish]". Once the NQS instance is obtained, it is fed into the NQS2SPARQL module to generate the SPARQL query, where the SPQRAL is a semantic query language for databases. This module analyzes the NQS instances and then maps the entities in the NQS instances to the entities in KGs. 

Other similar semantic parsing methods can be referred to~\cite{kate2005learning,riedel2010modeling, schmitz2012open,wong2007learning,xu2014answering,yahya2012natural,zettlemoyer2009learning,zou2014natural}.

\hide{
\begin{figure}[t]
	\centering
	\includegraphics[width=0.48\textwidth]{Figures/PA}
	\caption{\label{fig:PA} Pipeline of NLQF AskNow.}
\end{figure}
}

\vpara{Template-based Parsing.}
Template-based parsing usually contains an offline stage and an online stage, where the offline stage targets at generating a collection of templates and the online stage aligns the new arriving question to an existing template, and then aims to retrieve the answer from the KGs based on the matched template. 
The templates can be mainly summarized into two forms: one in natural language pattern and the other in the structured graph pattern. Both of them correspond to the structured queries such as SPARQL, which can be directly executed on KGs to retrieve the answers. Thus there are two main problems in template-based parsing methods: the first one is how to generate the templates and the second one is how to match questions with templates.

\hide{
\begin{figure}[t]
	\centering
	\includegraphics[width=0.48\textwidth]{Figures/UncertainTQA}
	\caption{\label{fig:UncertainTQA}Template generation in UncertainTQA~\cite{Zheng2015HowTB}.}
\end{figure}
}
As one of the pioneering work, Bast and Haussmann~\cite{Bast2015MoreAQ} manually construct three simple question templates without QA corpora. Since it is expensive to manually define these templates, especially for open-domain QA systems over large KGs, later work studies how to generate templates automatically. 
For example, Zheng et al. propose UncertainTQA~\cite{Zheng2015HowTB} to generate templates from the collected natural language questions and SPARQL queries on KGs. It first employs the existing method such as~\cite{Zou2014NaturalLQ} to translate each natural language question into a semantic query graph $g$. Meanwhile, a SPARQL query graph $q$ can be also constructed for each SPARQL query. Then graph edit distance (GED) is used to calculate the graph similarity between the query graph $g$ and the SPARQL query graph $q$. 
A template is defined as a pair of a semantic query graph $g$ and its most similar SPARQL query graph $q$. 
Based on the derived templates, given a new question, its most similar SPARQL query graph will be found and issued to the KGs to retrieve  the answers.
In the above process, the semantic query graph is uncertain, i.e., each node/edge has multiple possible labels with different probabilities, because of the semantic ambiguity. For example, the question ``which actress from the USA is married to Michael Jordan born in a city of NY" may be mapped to three different ``Michael Jordan" and two different ``NY".  Thus unlike the traditional GED algorithm, Uncertain TQA proposes a common structural subgraph(CSS)-based lower bound to avoid exhaustively enumerating every possible match of $g$, which is a uniform bound for GED. 

\hide{

\beqn{
    Pr\{pw(g)\} = \prod_{\forall v \in pw(g)} l(v).p \\
    SimP_\tau (q,g) = \sum_{pw(g) \in PW(g)} Pr\{pw(g)|ged(q,g) \leq \tau\}
}
Where $pw(g)$ is a possible world of uncertain graph $g$ and $Pr\{pw(g)\}$ the appearance probability of $pw(g)$. 
The CSS lower bound for GED is as follows:
\beqn{
    ged(q,g) \geq lb\_ged_{CSS}(q,g) \\\nonumber
    = |V| + |E| - \lambda_E(q,g) + \frac{dif(q,g)}{2} - \lambda_V(q,g),\\
    dif(q,g) = \sum_{i=1}^m (d(u_i,q)\ominus d(v_i,g))
}
Where $|V|$ denotes the larger number of the vertices of $q$ and $g$, $|E|$ denotes the number of edges of the larger one. $\lambda_E$ and $\lambda_V$ denote the common vertex and edge labels between q and g. And $a \ominus b = a-b(a>b)$, $a \ominus b = 0(a\leq b)$. $d(u_i,q)$ denotes i-th largest degree of $q$.
Once the similar pair are found, the mappings between the two graphs will lead to templates. The process is illustrated in Figure ~\ref{fig:UncertainTQA}. 
}

Similarly, QUINT proposed by Abujabal et al.~\cite{Abujabal2017AutomatedTG}  translates the question into a dependency parse tree~\cite{Klein2003AccurateUP}. Instead of matching each question with a SPARQL query graph by UncertainTQA, QUINT retrieves the smallest subgraph connecting the entities in the question and the answers from the KGs, and then maps the graph with the dependency parse tree of the question. The alignment between the two graphs forms a template. QUINT formulates the alignment problem as constrained optimization and finds the best alignment using integer linear programming (ILP). 

\hide{
\begin{figure}[t]
	\centering
	\includegraphics[width=0.4\textwidth]{Figures/Aqqu}
	\caption{\label{fig:Aqqu}The manually constructed auestion templates in Aqqu~\cite{ Bast2015MoreAQ }.}
\end{figure}
}

Both the above two methods generate templates in form of structured graphs. However, unlike a structured graph, a natural language template includes a natural language pattern and a corresponding SPARQL pattern. To generate the natural language template, one should first replace the entities in the question with their types, and then determine the relation type of the template. For example, for the question ``When was Barack Obama born?", a natural language pattern ``When was \$PERSON born?" is derived, which is also mapped to the predicate ``BirthPlace" and then is corresponded to the SPARQL pattern ``$<$Person$>$, birthPlace, ?place". TemplateQA~\cite{Zheng2018QuestionAO} and KBQA~\cite{Cui2017KBQALQ} are both this kinds of natural language templates. TemplateQA assumes that a natural language pattern is likely to be matched to a predicate if they are simultaneously shared by many entity pairs in KGs. KBQA proposes a probabilistic method to capture the matching likelihood between a natural language pattern and a predicate, where each natural language pattern $s$ is modeled as a hidden variable, and the maximum likelihood is adopted to estimate the matching probability $P(r|s)$.

\begin{figure*}[t]
	\centering
	\includegraphics[width=\textwidth]{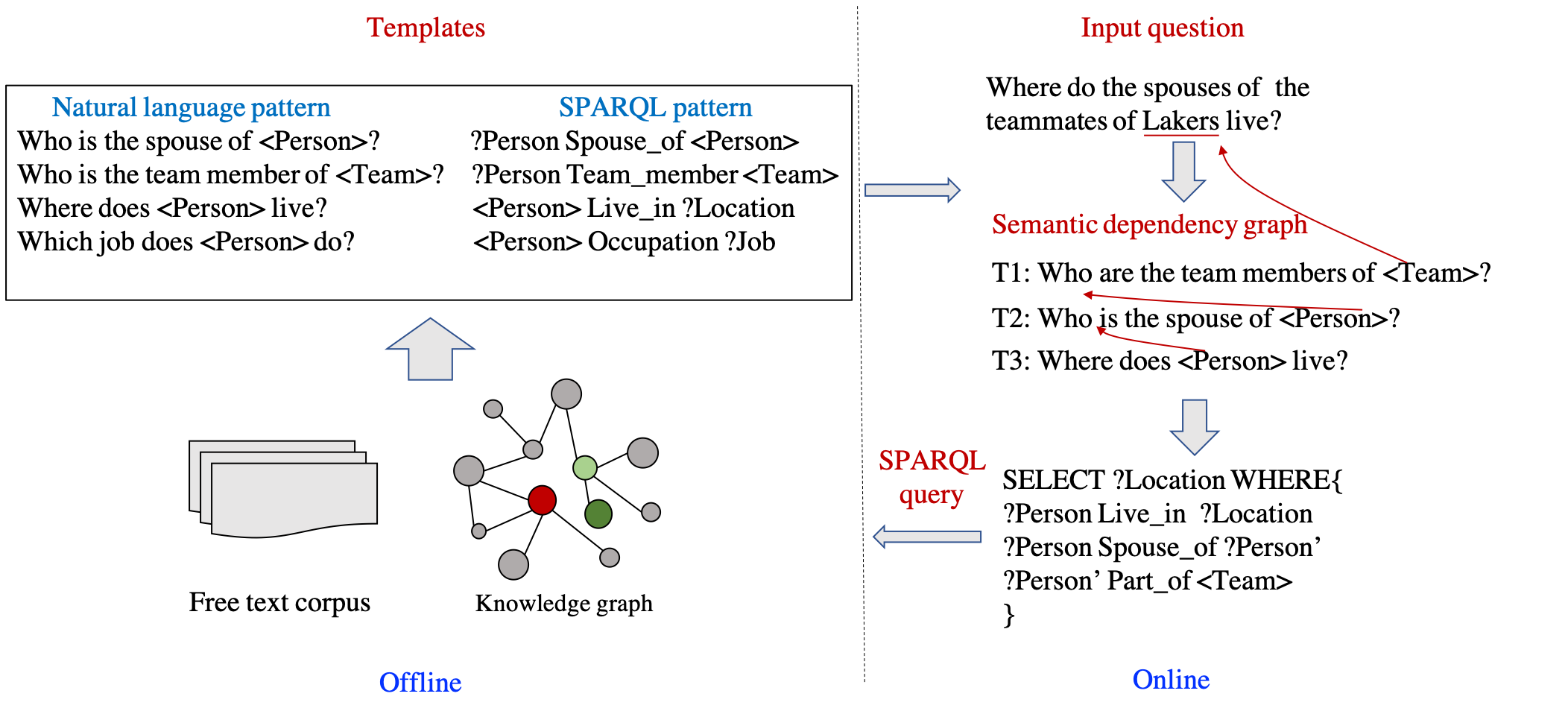}
	\caption{\label{fig:TemplateQA}The framework of TemplateQA~\cite{Zheng2018QuestionAO}.}
\end{figure*}

Given a set of derived templates, TemplateQA computes the Jaccard similarity coefficient between the question $q$ and the natural language template $t$. For complex questions, TemplateQA builds a semantic dependency graph (SDG) through matching each subsequence of $q$ with templates one by one. When one template  is identified, the subsequence is removed and replaced with the answer type of the template. Thus during the decomposition, the type constraint between the two adjacent templates (i.e., the neighboring nodes in SDG) is naturally guaranteed. To reduce the search space, TemplateQA employs both type-based and order-based optimizations. The whole process of TemplateQA is illustrated in Figure~\ref{fig:TemplateQA}, which includes the offline templates generation based on the text corpus and the KGs, and the online SDG parsing for the input question based on the generated templates. Once the SDG is parsed, it is mapped to a SPARQL query, which is issued to the KGs. 
%As for UncertainTQA, it parses questions into syntactic dependency trees, then existing tree edit distance (TED)-based approaches can be used to find the alignment between the question dependency tree and query dependency tree. Bast and Haussmann as well as QUINT use hand-crafted features to train classifiers to choose the best template.

The above methods deal with KGQA with complex questions in quite different approaches. 
For example, QUINT defines some dependency parse rewriting rules to get two separate propositions when facing relative clauses and coordinating conjunctions and then connect the two propositions with for example $conj$ or $pobj$ edges in the dependency parse tree, which enables it to answer conjunctive questions. 
KBQA decomposes the question into a chain of binary sub-questions where the answer of the last sub-question fills in the value of the variable in the next sub-question, which enables it to answer multi-hop questions. 
However, QUINT depends on the pre-defined rewriting rules and KBQA only supports multi-hop questions.
Thus, compared with them, in TemplateQA, the sub-questions form a dependency graph, which is more expressive.
In addition, ~\cite{Mazzeo2016AnsweringCN,TunstallPedoe2010TrueKO,Yahya2012NaturalLQ} are similar works which generate natural language templates while ~\cite{Unger2011PythiaCM,Unger2012TemplatebasedQA} generate structured graph templates.

\hide{
\begin{figure}[t]
	\centering
	\includegraphics[width=0.48\textwidth]{Figures/QUINT}
	\caption{\label{fig:QUINT}Rewriting rules of QUINT ~\cite{ Abujabal2017AutomatedTG }.}
\end{figure}
}

\vpara{Summary.}
Semantic parsing and template-based parsing are both symbolic ways to represent a question as a structured query. Semantic parsing directly learns a function to map the question to a structured query, while template-based parsing first generates a collection of templates and then maps the question to an existing template to obtain the structured query.
These symbolic reasoning methods are good at dealing with complex logic questions, as the structured query can be trees or graphs, which are expressive enough to represent complex logic questions. However, similar to KGC, the symbolic reasoning methods cannot deal with  the ambiguity of natural languages and the uncertainty of entities and relations. 
%Essentially, the structured query is a kind of representation for the question, just as the hand-crafted features or the vector in embedding-based methods. All of the them expect to identify the same questions and distinguish the different ones via the representation. %At the same time, the representation needs to be close to the corresponding predicate(relation). 
%How to generate more accurate templates and how to represent multifarious questions with templates are still the focus of future research on Template-based Parsing.

\hide{
\begin{figure*}[t]
	\centering
	\includegraphics[width=\textwidth]{Figures/multique}
	\caption{\label{fig:MULTIQUE} The sub-query graphs generated sequentially by MULTIQUE for an example question ``What college did the author of `The Hobbit' attend?"~\cite{bhutani2020answering}.}
\end{figure*}
}

\begin{table*}[htbp]
	\newcolumntype{?}{!{\vrule width 1pt}}
	
	\caption{Constraint categories proposed by Bao et al.~\cite{bao2016constraint}.}
	\label{tb:constraints}
	\centering
	\begin{tabular}{c ?l ? l}
		\toprule
		
		\textbf{Constraint Category} & \textbf{Example} & \textbf{Percentage} \\
		\midrule
		Multi-Entity	               & Which films star by \textbf{Forest Whitaker} and are directed by \textbf{Mark Rydell}?  & 30.6\%\\
		Type		                     & Which \textbf{city}  did Bill Clinton born? & 38.8\%\\
		Explicit Temporal 		  & Who is the governor of Kentucky \textbf{2012} ? & 10.4\%\\
		Implicit Temporal          & Who is the us president \textbf{when the Civil War started}? & 3.5\% \\
		Ordinal				            & What is the \textbf{second longest} river in China? & 5.1\%\\
		Aggregation                 &  \textbf{How many} children does Bill Gates have? &  1.2\%\\
		\bottomrule
	\end{tabular}
\end{table*}

\subsection{Neural-Symbolic Reasoning}
\label{sec:neural_symbolic_reasoning_for_QA}

Neural-symbolic reasoning combines the advantages of both the neural reasoning and the symbolic reasoning for KGQA. These  hybrid methods generally fall into two categories: the neural-enhanced symbolic reasoning which only targets at parsing the questions, or the end-to-end reasoning which parses the questions and retrieves the answers simultaneously. 

\subsubsection{Neural-enhanced Symbolic Reasoning}

Neural-enhanced symbolic reasoning still parses the given question into a query graph. After parsing, a neural network is leveraged to evaluate whether the parsed query graph is relevant to the given question.

\vpara{Single-relation Question.}
Yih et al.~\cite{Yih2014} solve the single-relation question by completing two tasks:  linking the mention in the question to an entity in KGs as the topic entity, and mapping the relation pattern described by the question to a relation in KGs.  For example, given a question such as ``\textit{When were DVD players invented?}", the paper first extracts the mention ``\textit{DVD players}" and derives the relation pattern ``when were X invented", and then links ``\textit{DVD players}" to the entity ``\textit{dvd-player}" and also maps ``when were X invented" to the relation ``\textit{be-invent-in}".  Once the topic entity and the relation are detected, the answer to the question can  be directly queried by the relation-entity triple ``\textit{be-invent-in}(\textit{dvd-player},?)" in the KGs. To extract the mention and derive the relation pattern, the authors simply enumerate all the combinations. The key point is to determine the mapping between the extracted mention and the entity as well as the mapping between the derived pattern and the relation. To achieve the goal, the authors propose a CNN model to take the sequential tokens in a mention or a relation pattern as input and output the corresponding embedding, which is then compared with the embedding of the ground truth entity or relation in KGs.

\vpara{Multi-hop Question.}
In comparison with the simple questions, the searching space grows exponentially if the given question involved multi-hop relations. To tackle the multi-hop relation questions, instead of generating the whole query graph at once,  MULTIQUE~\cite{bhutani2020answering} breaks the original question into simple partial queries and builds sub query graphs for partial queries one by one. The search space is shrinked since each time when extending the whole query graph by a new sub-query graph, the model only needs to consider the immediate answers queried by the previous most matched sub-query graph. 
In the graph generation process, how to measure the quality of a sub-query graph is a key problem to be solved. The authors apply an LSTM model to encode the token sequence of the given question, where an attention mechanism is incorporated to emphasize a particular part of the question which is expected to be represented by the current sub-query graph. Meanwhile, the relation and the constraint in each sub-graph are also encoded by the identification and the tokens.
Once the embeddings of the original question and the sub-query graph are encoded, they are concatenated and fed to an MLP, which outputs a scalar to reflect the similarity between the question and the  sub-query graph.

\vpara{Complex-logic Question.}
Bao et al.~\cite{bao2016constraint} further deal with questions with multiple constraints, which are presented in Table~\ref{tb:constraints}. They first transform a question into a multi-constraint query graph, then propose a Siamese convolutional neural networks to calculate the similarity between the query graph and the input natural language question, and finally execute the top-ranked query graph on KGs by instantiating all variable nodes according to the constraints in order.  The multi-constraint query graph contains two types of nodes, where a constant node represents a ground entity in KGs such as ``Barachk Obama" or an attribute value such as ``1961", and a variable node represents an unknown entity or unknown attribute value. The graph also contains two types of edges where a relational edge represents a relation such as ``birthday" in KGs and a functional edge represents a functional predicate of a truth such as $<$ in the truth $\langle 2000,<, 2001\rangle$. To construct the query  graph, an entity linking method proposed by~\cite{yang-chang-2015-mart} is leveraged to detect topic entities from the given question. For each topic entity, one-hop relations or two-hop relations are extended from it to form a basic query graph, and then different types of constraints in Table~\ref{tb:constraints} are detected from the question and binded to the basic query graph. Later, instead of adding constraints only after relation paths have been constructed, Lan et al.~\cite{lan-jiang-2020-query}  propose a method to incorporate constraints and extend relation paths simultaneously, which can effectively reduce the search space.

\subsubsection{End-to-end Reasoning}

End-to-end reasoning parses the questions and retrieves the answers in a unified model. According to the form of the connections between the topic entities and the answers, we categorize the end-to-end reasoning into path-based and graph-based methods.

\vpara{Path-based Reasoning.}
Path-based methods employ a hop-by-hop path search over KGs, which usually contains three stages of dealing with the input question, reasoning over the KGs and predicting the answers.

\begin{figure*}[t]
	\centering
	\includegraphics[width=0.8\textwidth]{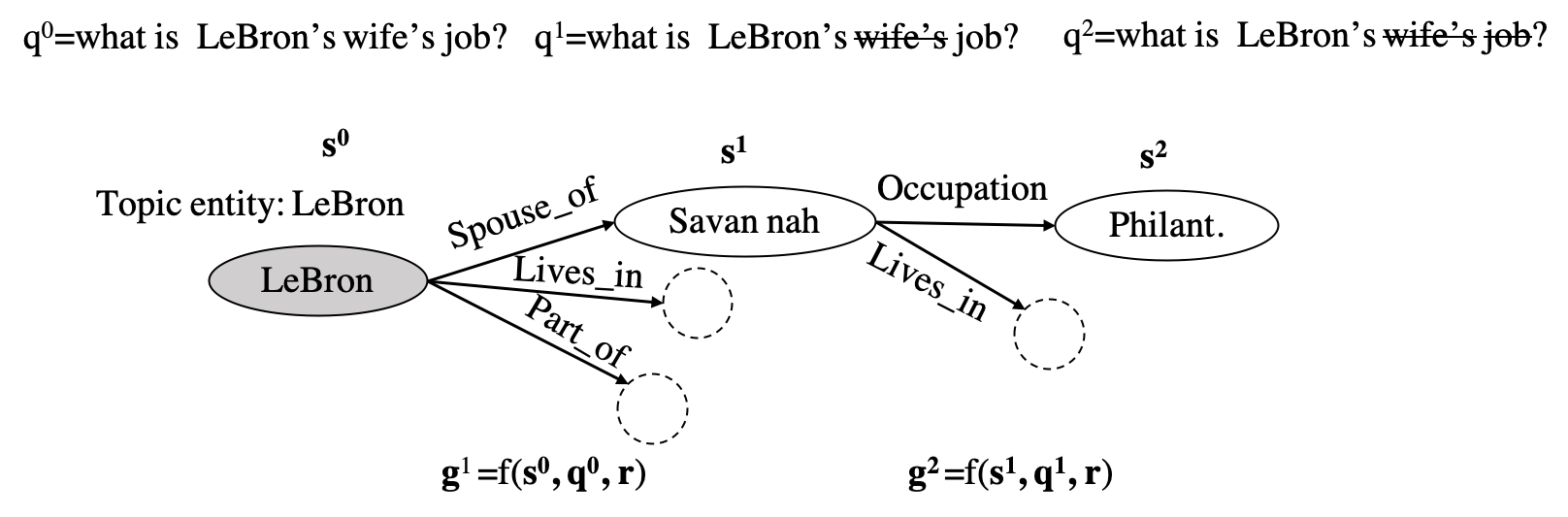}
	\caption{\label{fig:IRN} Illustration of IRN~\cite{zhou2018an}. Notation $\textbf{s}^h$ represents the state vector of the $h$-th hop, which is initialized by the embedding of the topic entity. $\textbf{g}^{h}$ is the vector of the probabilities to select each relation according to the $h$-th state vector, question embedding and each relation embedding. The question embedding is updated after each hop's selection.}
\end{figure*}

For example, IRN (Interpretable Reasoning Network)~\cite{zhou2018an} proposes three modules corresponding to the above three stages in a unified model. The input module initializes the question by the embeddings of the words in the question and updates the question's embedding hop-by-hop according to inference results of the reasoning module. The reasoning module initializes its state by the topic entity of the question and predicts the embedding of the relation $\hat{\mathbf{r}}^h$ at the $h$-th hop based on the question's embedding $\mathbf{q}^{h-1}$ and the state vector $\mathbf{s}^{h-1}$ at the $(h-1)$-th hop, i.e., 

\beqn{
	\label{eq:IRN_predict_relation}
	g_j^h \!\!\!\!\!&=&\!\!\!\!\! 	\text{softmax}((\mathbf{M}_{rq} \mathbf{r}_j)^T \mathbf{q}^{h-1} \!\! + \!\! (\mathbf{M}_{rs} \mathbf{r}_j)^T \mathbf{s}^{h-1}),\\ \nonumber
	\hat{\mathbf{r}}^h  \!\!\!\!\!&=&\!\!\!\!\!  \sum_{j} g_j^h * \mathbf{r}_j,
}

\noindent where $\mathbf{r}_j$ is the embeding of the $j$-th relation in KGs, $g_j^h$ is the probability of selecting the $j$-th relation in KGs, and $\mathbf{M}_{rs},\mathbf{M}_{rq}$ are the project matrices mapping $\mathbf{r}$ from the relation space to the state space and to the question space respectively. Then the predicted relation $\hat{\mathbf{r}}^h$ is utilized to update the state vector and the question embedding by:

\beqn{
	\label{eq:IRN_update_state}
    \mathbf{s}^h  &=&  \mathbf{s}^{h-1} + \mathbf{M}_{rs} \hat{\mathbf{r}}^h,\\\nonumber
    \label{eq:IRN_update_question}
    \mathbf{q}^h  &=&  \mathbf{q}^{h-1} - \mathbf{M}_{rq} \hat{\mathbf{r}}^h. 
}

In the above equation, in order to pay attention to different parts of the question at each hop, the predicted relations from previous hops are removed from the question at each hop. The answer module predicts an entity conditioned on the similarity between its pre-trained embedding  and the reasoning state at the last hop. In addition to the answer entity, the entities at the intermediate hops are also predicted and evaluated to improve the answer prediction performance, i.e.,  

\beqn{
    \mathbf{e}^h &=& \mathbf{M}_{se} \mathbf{s}^h,\\\nonumber
    o_j^h &=& P(a^h = e_j | \mathbf{s}^h) = softmax(\mathbf{e}_j^T \mathbf{e}^h),
}

\noindent where $\mathbf{M}_{se}$ is the projection matrix mapping $\mathbf{s}^h$ from the state space to the entity space and $\mathbf{e}_j$ is the pre-trained embedding of the $j$-th entity in KGs. Figure~\ref{fig:IRN} illustrates the three modules in IRN.

IRN assumes that the ground truth paths for answering the questions are also observed, thus the intermediate entities and the answer entities can both supervise the model training process. However, in most cases, we can only obtain the answers to the questions, without knowing the reasoning paths. To deal with the challenge, SRN (Stepwise Reasoning Network)~\cite{qiu2020stepwise} formulates the reasoning process as a Markov decision process where the answer entity can provide the delayed reward to the decision of the relation at each hop. In addition to the delayed reward, SRN also incorporates the intermediate reward to overcome the delayed and sparse rewards. To emphasize different parts of a question, instead of removing the predicted relations from the question by IRN, SRN employs the attention mechanism to decide which part should be focused on at present. ~\cite{Neelakantan2015Compositional} ~\cite{Das2017MINERVA} ~\cite{Das2017chain} are similar path-based works.

%In input stage, SRN applies a bidirectional GRU network to obtain the representation of the question. In reasoning stage, at each hop, to use different semantic information of the question, SRN employs single-layer perceptron to get the representation transformed at each hop which makes it step-aware, and attention mechanism to make each potential action (relation and entity pair) interact with the representation which makes it relation-aware. Moreover, SRN encodes the decision history through another GRU network to emphasize the precedence order of triple selection. In answer stage, SRN incorporates relation embeddings, relation-aware question representations, and encoded decision history to calculate the semantic scores, which helps to predict the answer entity. 

The path-based reasoning methods can obtain explicit paths, i.e., logic rules for solving multi-hop QA. These methods usually employ deliberate analysis on the input question in order to focus on different parts of the question at each hop. Meanwhile, since there is a need for reasoning over several hops, intermediate reward can be powerful signals to supervise the whole training process.

\vpara{Graph-based Reasoning.}
Instead of extending a single path from the topic entity to the answer, the graph-based reasoning methods extend a subgraph around the topic entity, which is more expressive than a single path.
The general idea of graph-based reasoning is to extend a subgraph and then reason the answer in it by the recent technique of graph representation learning.

\begin{figure}[t]
	\centering
	\includegraphics[width=0.5\textwidth]{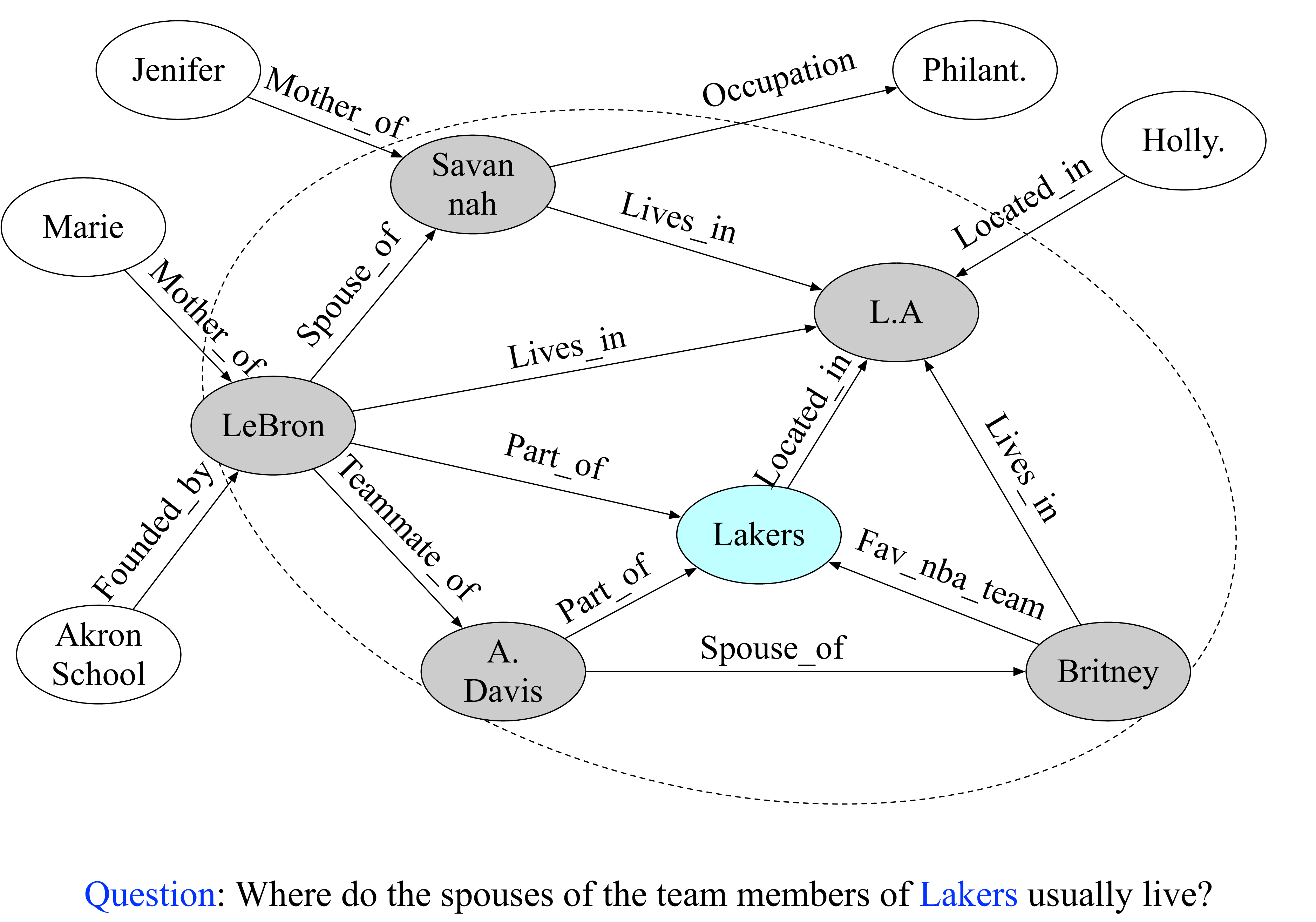}
	\caption{\label{fig:graphnet} Illustration of Graft-Net ~\cite{sun-etal-2018-open}. Given the topic entity ``Lakers" in a question, Graft-Net first extracts the subgraph around the topic entity and then performs a variant GNN model to represent each node in the subgraph.  }
\end{figure}

Graft-Net~\cite{sun-etal-2018-open} is a representative graph-based reasoning model. It first performs a Personalized PageRank (PPR)~\cite{Haveliwala2002TopicsensitiveP} around the topic entities in the question and then incorporates other entities that might be an answer to the question. In PPR, the relations in KGs which are more relevant to the question are weighted higher. After running PPR, they retain the top entities by PPR score, along with all the relations between them. Graft-Net not only leverages the original KGs, but also incorporates Wikipedia text copus as the additional source to infer the answers. When extracting a subgraph from the text corpus, both the most relevant sentences to the question and any entities linked to these sentences are extracted as the nodes, and then the relations from the KGs among these entities, plus the mention links between the sentences and the entities are extracted as the edges in the subgraph.
The subgraphs extracted from the KGs and the text corpus are merged together to construct the final subgraph. 

Since the final subgraph is a heterogeneous graph that includes both the entities and the sentences as the nodes, Graft-Net performs a variant GNN model to represent different types of nodes by different update rules, i.e., the embedding of an entity $v$ at the $l$-th step is updated as follows:

\beq{
	h_v^{(l)}=\text{FFN}
	\begin{pmatrix}
	\begin{bmatrix}
	h_v^{(l-1)} \\
	h_q^{(l-1)} \\
	\sum_{r} \sum_{v' \in N_r(v)} \alpha_r^{v'} \psi_r(h_{v'}^{(l-1)}) \\
	\sum_{(d,p) \in M(v)} H_{d,p}^{(l-1)} \\
	\end{bmatrix}
	\end{pmatrix}
}

\noindent where the first two terms correspond to the entity embedding and the question embedding, respectively, from the previous layer. The third term aggregates the states from the entity neighbors of any relations of the current entity, i.e. $N_r(v)$, after scaling with an attention weight $\alpha_r^{v'}$ and applying relation specific transformation $\psi_r$ proposed in R-GCN~\cite{schlichtkrull2018modeling}. The last term aggregates the states of all tokens that correspond to the mentions of the entity $v$ among the documents in the subgraph. In the above equation,  $M(v) = \{(d,p)\}$ is the set of document-position pairs, where each pair $(d,p)$ indicates $d$ mentions entity $v$ at the position $p$. The embedding of a document $d$ in a subgraph at the $l$-th step is updated by:

\beqn{
	\tilde{H}_{d,p}^{(l)} &=& \text{FFN}(H_{d,p}^{(l-1)},\sum_{v \in L(d,p)} h_v^{(l-1)}) \\\nonumber
	H_d^{(l)} &=& \text{LSTM}(\tilde{H}_{d,p}^{(l)})
}

\noindent where $L(v) = \{(d,p)\}$ is the set of entities linked to entity $v$ at position $p$ at document $d$. The above equations first aggregate over the entity states coming in at each position separately and then aggregate states within the document using an LSTM network.
Graft-Net predicts whether one entity in the subgraph is the answer based on the entity's embedding of the last step. Figure~\ref{fig:graphnet} illustrates the basic idea of Graft-Net without the text information.

However, the question-specific subgraphs the Graft-Net  builds heuristically are far from optimal, i.e., they are often much larger than necessary, and sometimes do not contain the correct answer. 
Thus, PullNet~\cite{SunACL2019} proposes a policy function to learn how to construct the subgraph, rather than using an ad-hoc subgraph-building strategy. The classification model used to predict the answer in the subgraph is the same as Graft-Net. Since the intermediate entities in the subgraph are latent, PullNet proposes a weak supervision method to train the policy function. The general idea is to find all shortest paths between topic entities and the answer entities and mark the entities in such shortest paths as the ground truth intermediate entities.

Graft-Net and PullNet assume the topic entities are given, however, in many cases, only the questions are provided. 
Thus, VRN (Variational Reasoning Network)~\cite{zhang2018variational} models the topic entities as hidden variables and proposes two probabilistic modules in a unified architecture, one for topic entity recognition ($P(y | q)$) and the other for logic reasoning ($P(a | y,q)$), where $y$ indicates a hidden topic entity, $q$ refers to the query and $a$ refers to the answer. The two modules are jointly trained such that they can coordinate with and benefit from each other. However, the reasoning-graph is also an ad-hoc subgraph. Instead of performing PPR to build the subgraph as Graft-net does, VRN performs topological sort within $T$ hops starting from a topic entity $y$ to build the scope $\mathcal{G}_y$. For each potential answer $a \in \mathcal{G}_y$, the reasoning graph $\mathcal{G}_{y \rightarrow a}$ is represented as the minimum subgraph that contains all the paths from $y$ to $a$ in $\mathcal{G}_y$. Then VRN proposes a ``forward graph embedding" method to embed the reasoning-graph for each answer $a$ recursively using its parents' embeddings. Finally, the similarity between the reasoning-graph embedding and the question representation is calculated as the score to predict the answer.
Other similar graph-based works include~\cite{Bordes2014QuestionAW} ~\cite{Hu2018ASF}.

\vpara{Summary.}
The first kind of the neural-symbolic reasoning for KGQA, i.e., the  neural-enhanced symbolic reasoning, leverages the neural networks to evaluate the similarity between the parsed query graph and the natural language question. These methods usually need the ground truth of the query graphs for the questions, which are not easy to be annotated. 

The second kind of  the neural-symbolic reasoning for KGQA, i.e., the end-to-end reasoning, directly aligns the questions with the answers, where the answers are easily obtained and can be used as the ground truth to supervise the alignment. The objective of these methods is to represent the inferred path or graph into an embedding vector, based on which the answer can be determined. The reasoned paths/graphs can be viewed as the explanation for the reasoning results. However, the operations over the embeddings can only support the single-relation and multi-hop relation questions. It is still unknown how to address the  complex logic questions by the end-to-end reasoning methods.

\begin{figure*}
	\centering
	
	\subfigure[KGC]{\label{subfig:KGC}
		\includegraphics[width=0.9\textwidth]{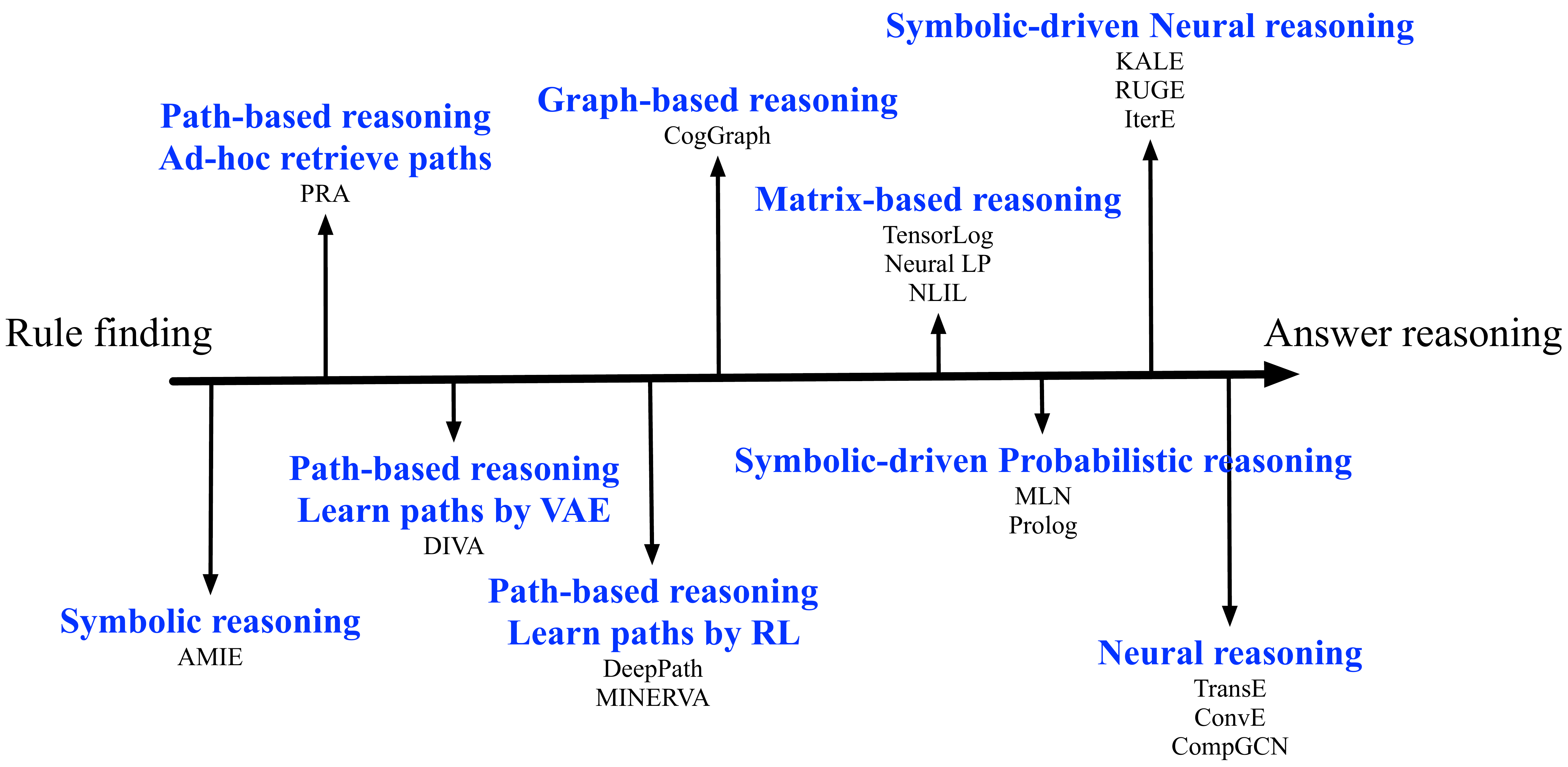}
	}
	\subfigure[KGQA]{\label{subfig:KGQA}
		\includegraphics[width=0.9\textwidth]{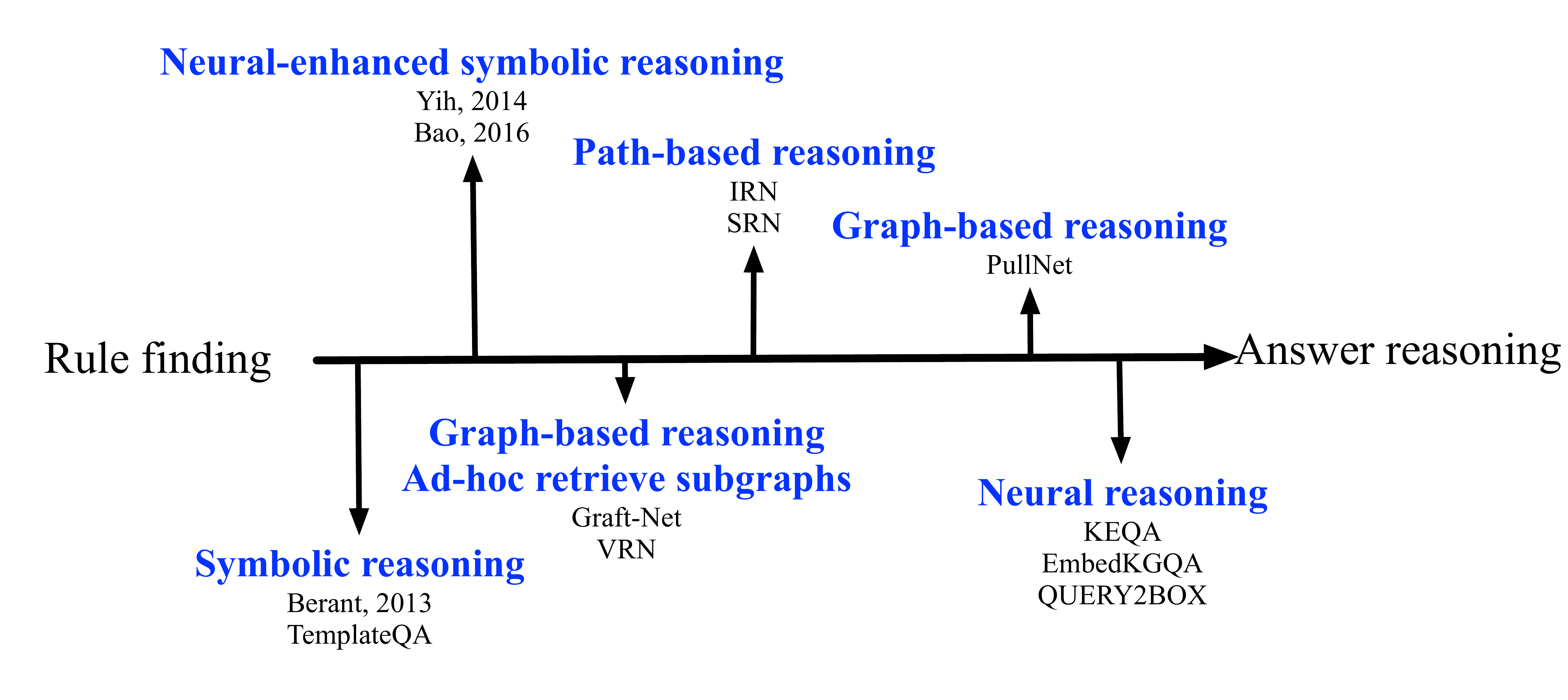}
	}
	\caption{\label{fig:unified_reasoning_framework} Summary of KGC and KGQA in a unified reasoning framework.}
\end{figure*}

\section{Conclusion and Future Directions}
\label{sec:con} 

In this section, we first conclude this survey by casting KQC and KGQA in a unified reasoning framework and then discuss  some potential future directions.

\subsection{Conclusion}

Generally, most of the reasoning methods for completion can be decomposed into two key components: rule finding and answer reasoning, where rule finding targets at inferring the rules from the observed triplets in KGs, and answer reasoning aims to predict the answer for the given head entity and the query relation. 
Different methods for KGC emphasize different components. In Figure~\ref{subfig:KGC}, going from left to right, the methods pay more and more attention to answer reasoning and less and less attention on rule finding. 
For example, on the far left, the pure symbolic reasoning methods such as AMIE only explain how to find rules from the data but without discussing the usage of the rules to reason answers. 
Then the path-based reasoning method PRA not only ad-hocly retrieves the paths by PageRank, but also trains a simple linear regression model based on the paths to reason answers. However, how to retrieve the paths is still the core problem to be solved in PRA. 
DIVA treats the two components equally in a unified model through formulating paths as hidden variables. 
Subsequently, DeepPath and MINERVA directly reason the answers by RL, without explicitly deriving the paths. 
Graph-based reasoning such as CogGraph extends a single path to a subgraph and matrix-based reasoning such as TensorLog and Neural LP extends the subgraph to the whole graph with attentions on different nodes at each hop. 
Although the explicit rule finding is missing, the path-based, graph-based, and matrix-based reasoning can still derive the paths according to the selections or attentions at each hop.
At the same time, symbolic-driven probabilistic reasoning and symbolic-driven neural reasoning thoroughly take the answer reasoning as the target. The difference is that in the symbolic-driven probabilistic reasoning, the rules are used as the features to predict the answers, while in the symbolic-driven neural reasoning, the rules are used to  generate more facts for learning high-quality embeddings.
Finally, the neural reasoning methods get rid of the rules and reason answers entirely based on the embedding techniques. 

Most of the reasoning methods for KGQA can also be decomposed into rule finding and answer reasoning, where the former targets at parsing the rule, i.e., a path or a subgraph from the given question, and the latter  aims to predict the answer for the given question.
In Figure~\ref{subfig:KGQA}, the methods from the left to the right gradually pay more attention to answer reasoning and less attention to rule finding. Starting from the left, the pure symbolic methods and the neural-enhanced symbolic reasoning methods only parse the questions to obtain the executable queries, query paths, or subgraphs, where the neural-enhanced symbolic reasoning methods incorporate additional embedding techniques to evaluate the parsed paths or subgraphs. Then the graph-based reasoning methods Graft-Net and VRN retrieve the subgraphs heuristically and then reason the answers from the subgraphs by neural networks or graph convolution networks. Later, the path-based methods IRN and SRN and the graph-based method PullNet reason the paths/subgraphs and the answers in an end-to-end manner by RL or weak supervision techniques. The far right  neural reasoning methods get rid of the paths/subgraphs entirely and directly measure the relationships between the answers and the given question. The pure symbolic methods can parse quite complex questions such as those with multiple constraints as they only parse the questions. However, the path/graph-based end-to-end methods need to extend the paths/subgraphs along with the relations in KGs, making it not easy to incorporate the complex logic questions.

\subsection{Future Directions}

Despite the existing researches on reasoning methods for KGC and KGQA, there are still unsolved challenges on these tasks such as reasoning for the few-shot relations in KGC and reasoning for the complex questions in KGQA. In addition, existing works mainly focus on leveraging the knowledge graph structures for reasoning, but only a few works investigate how to leverage the side information of the entities and the relations to benefit the reasoning task~\cite{fu2019CPL,sun-etal-2018-open,toutanova2015representing}, the performance of which can be still improved. Moreover, some other kinds of reasoning tasks on KGs such as dynamic reasoning and analogical reasoning are also worth studying. Finally, existing reasoning methods lack the transferability from one KG to others.
We explain these future directions in detail as below.

\vpara{Few-shot Reasoning.} Most of the neural-symbolic models heavily rely on a huge amount of training instances. However, the relationships between entities in KGs are far from complete, especially for the rare relations, making it extremely difficult to capture the underlying patterns of these rare relations.

Few-shot learning is a paradigm proposed for learning in the scenario of lacking training instances, which has first shown the significant performance in computer vision~\cite{fei2006one}. Then, few-shot learning in KGs, which aims to discover the underlying patterns of a relation with which only a few triplets is associated, has been studied recently~\cite{Du-2019, xiong2018one}. Although these are good attempts, the poor performance of reasoning tasks reported by them indicates that few-shot reasoning is still an unsolved challenge.

\vpara{Answering Complex Questions.}
Existing neural-symbolic reasoning models are criticized as most of them can only answer the single-relation questions, or limited-hop relations (e.g., three-hop relations), let alone the questions with additional constraints. Ding et al.~\cite{ding2019cognitive} showed that the performance of the traditional reasoning models on question answering task decreases dramatically with the increase of the hops, as the search space increases exponentially with the hops, making it difficult to reason the correct answer from such big search space.

Recently for question answering upon text corpus, neural-symbolic models incorporating human cognition have shown their superiority on reasoning capacity~\cite{ding2019cognitive}, as they consistently perform well with the increase of the hops in questions. Thus, a new way is called for to incorporate human cognition into neural-symbolic reasoning in KGs. Humans usually answer multi-hop questions  following two reasoning systems, where the first system makes fast and intuitive thinking for collecting enough  raw evidence, and the second system makes slow and logical thinking for reasoning among the collected raw data. How to model the two systems in a neural-symbolic reasoning framework in KGs effectively is a promising direction.

\vpara{Reasoning upon Multi-sources.}
Since the structure information comprised of entities and relations in a knowledge graph is far from complete, incorporating additional information from unstructured text data for reasoning is encouraged. 

Although some models such as Graft-Net~\cite{sun-etal-2018-open} and PullNet~\cite{SunACL2019} are the state-of-the-art reasoning models on both the structural and the textual information, it is still challenging to determine the correct evidence to be linked to the incomplete graph structure from the large textual data. Meanwhile, although the textual information can enrich the KGs, it contains much useless and redundant information, which may result in side effects on the reasoning tasks.

\vpara{Dynamic Reasoning.} Dynamic reasoning aims at learning new logic rules and inferring new facts evolving with time. Existing reasoning methods are all devoted to reasoning in the static KGs, but they ignore the temporal information contained in knowledge. However, as we all know, the facts contained in KGs such as (Steve Jobs, CEO of, Apple inc.) are not always true over time. Besides, new knowledge is produced by humans continually, which may be injected into KGs dynamically. Thus, dynamic reasoning upon the dynamic KGs is demanded to self-correct KGs and mine new logic rules continually.

\vpara{Analogical Reasoning.} Learning quickly is a hallmark of human intelligence, which involves figuring out the underlying patterns in a new domain by adopting the experience in old domains. It will be appreciated that if the reasoning models on KGs are able to perform the same adaptive learning by comparing the similarities between the new KGs and old KGs. We take academic knowledge graphs as examples to explain analogical reasoning. Suppose in these KGs for different academic fields such as computer vision (CV) and  natural language processing (NLP), science problems and techniques are added as the nodes, and multiple relationships such as techniques solving problems and techniques citing techniques are added as the edges. Some deep learning techniques such as CNN~\cite{krizhevsky2017imagenet} and pre-training~\cite{liu2020} techniques originally proposed in the problems of CV are further adapted to the problems of NLP, and have shown their superior performance, which can be treated as typical analogical reasoning on the problems between CV and NLP.

\vpara{Knowledge Graph Pre-training.} Neural reasoning methods for KGC such as TransE and ConvE reviewed in \secref{sec:neural_reasoning_for_KGC} produce entity embeddings and relation embeddings, which can be incorporated into the symbolic reasoning process to improve the capacity of fault-tolerance (Cf. ~\secref{sec:neural_driven_symbolic_reasoning} for details). However, these neural reasoning models treat the embeddings for all entities and relations in the given knowledge graph as parameters to be learned, which cannot be transferred to other knowledge graphs. 

Recently, the transferable pre-training graph neural networks have proved to be able to capture the graph structure characteristics across different graph data~\cite{hu2020gpt-gnn,qiu2020gcc}. Inspired by the success of the graph pre-training models, a knowledge graph pre-training model that can capture the transferable semantics of the entities and relations across different knowledge graphs is worth studying.

\balance
\bibliographystyle{abbrv}
\bibliography{reference}

\begin{thebibliography}{100}

\bibitem{Abujabal2017AutomatedTG}
A.~Abujabal, M.~Yahya, M.~Riedewald, and G.~Weikum.
\newblock Automated template generation for question answering over knowledge
  graphs.
\newblock {\em Proceedings of the 26th International Conference on World Wide
  Web}, page 1191–1200, 2017.

\bibitem{bao2016constraint}
J.~Bao, N.~Duan, Z.~Yan, M.~Zhou, and T.~Zhao.
\newblock Constraint-based question answering with knowledge graph.
\newblock In {\em Proceedings of COLING 2016, the 26th International Conference
  on Computational Linguistics: Technical Papers}, pages 2503--2514, 2016.

\bibitem{Bast2015MoreAQ}
H.~Bast and E.~Haussmann.
\newblock More accurate question answering on freebase.
\newblock In {\em Proceedings of the 24th {ACM} International Conference on
  Information and Knowledge Management}, pages 1431--1440, 2015.

\bibitem{bengio2007greedy}
Y.~Bengio, P.~Lamblin, D.~Popovici, and H.~Larochelle.
\newblock Greedy layer-wise training of deep networks.
\newblock In {\em Advances in neural information processing systems}, pages
  153--160, 2007.

\bibitem{berant2013semantic}
J.~Berant, A.~Chou, R.~Frostig, and P.~Liang.
\newblock Semantic parsing on freebase from question-answer pairs.
\newblock In {\em Proceedings of the 2013 Conference on Empirical Methods in
  Natural Language Processing}, pages 1533--1544, 2013.

\bibitem{besold2017neural}
T.~R. Besold, A.~d. Garcez, S.~Bader, H.~Bowman, P.~Domingos, P.~Hitzler, K.-U.
  K{\"u}hnberger, L.~C. Lamb, D.~Lowd, P.~M.~V. Lima, et~al.
\newblock Neural-symbolic learning and reasoning: A survey and interpretation.
\newblock {\em arXiv preprint arXiv:1711.03902}, 2017.

\bibitem{bhutani2020answering}
N.~Bhutani, X.~Zheng, K.~Qian, Y.~Li, and H.~Jagadish.
\newblock Answering complex questions by combining information from curated and
  extracted knowledge bases.
\newblock In {\em Proceedings of the First Workshop on Natural Language
  Interfaces}, pages 1--10, 2020.

\bibitem{Bollacker2008}
K.~Bollacker, C.~Evans, P.~Paritosh, T.~Sturge, and J.~Taylor.
\newblock Freebase: A collaboratively created graph database for structuring
  human knowledge.
\newblock In {\em Proceedings of the 2008 ACM SIGMOD International Conference
  on Management of Data}, pages 1247--1250, 2008.

\bibitem{Bordes2014QuestionAW}
A.~Bordes, S.~Chopra, and J.~Weston.
\newblock Question answering with subgraph embeddings.
\newblock In {\em Proceedings of the 2014 Conference on Empirical Methods in
  Natural Language Processing}, pages 615--620, 2014.

\bibitem{bordes2013translating}
A.~Bordes, N.~Usunier, A.~Garcia-Duran, J.~Weston, and O.~Yakhnenko.
\newblock Translating embeddings for modeling multi-relational data.
\newblock In {\em Proceedings of the 27th Annual Conference on Neural
  Information Processing Systems 2013}, pages 2787--2795, 2013.

\bibitem{bos-etal-2004-wide}
J.~Bos, S.~Clark, M.~Steedman, J.~R. Curran, and J.~Hockenmaier.
\newblock Wide-coverage semantic representations from a {CCG} parser.
\newblock In {\em Proceedings of the 20th International Conference on
  Computational Linguistics}, pages 1240--1246, 2004.

\bibitem{brown2020language}
T.~B. Brown, B.~Mann, N.~Ryder, M.~Subbiah, J.~Kaplan, P.~Dhariwal,
  A.~Neelakantan, P.~Shyam, G.~Sastry, A.~Askell, et~al.
\newblock Language models are few-shot learners.
\newblock {\em arXiv preprint arXiv:2005.14165}, 2020.

\bibitem{bryant1986graph}
R.~E. Bryant.
\newblock Graph-based algorithms for boolean function manipulation.
\newblock {\em Computers, IEEE Transactions on}, 100(8):677--691, 1986.

\bibitem{Carlson2010}
A.~Carlson, J.~Betteridge, B.~Kisiel, B.~Settles, E.~R. Hruschka, and T.~M.
  Mitchell.
\newblock Toward an architecture for never-ending language learning.
\newblock In {\em Proceedings of the Twenty-Fourth AAAI Conference on
  Artificial Intelligence}, page 1306–1313, 2010.

\bibitem{chakrabarti2007dynamic}
S.~Chakrabarti.
\newblock Dynamic personalized pagerank in entity-relation graphs.
\newblock In {\em Proceedings of the 16th International Conference on World
  Wide Web}, pages 571--580, 2007.

\bibitem{chen2020simple}
T.~Chen, S.~Kornblith, M.~Norouzi, and G.~Hinton.
\newblock A simple framework for contrastive learning of visual
  representations.
\newblock {\em arXiv preprint arXiv:2002.05709}, 2020.

\bibitem{chen2018diva}
W.~Chen, W.~Xiong, X.~Yan, and W.~Wang.
\newblock Variational knowledge graph reasoning.
\newblock In {\em Proceedings of the 2018 Conference of the North American
  Chapter of the Association for Computational Linguistics: Human Language
  Technologies}, pages 1823--1832, 2018.

\bibitem{CHEN2020}
X.~Chen, S.~Jia, and Y.~Xiang.
\newblock A review: Knowledge reasoning over knowledge graph.
\newblock {\em Expert Systems with Applications}, 141:112948, 2020.

\bibitem{chen2016scalekb}
Y.~Chen, D.~Z. Wang, and S.~Goldberg.
\newblock Scalekb: scalable learning and inference over large knowledge bases.
\newblock {\em The VLDB Journal}, 25(6):893--918, 2016.

\bibitem{Chen2019BidirectionalAM}
Y.~Chen, L.~Wu, and M.~J. Zaki.
\newblock Bidirectional attentive memory networks for question answering over
  knowledge bases.
\newblock {\em ArXiv}, abs/1903.02188, 2019.

\bibitem{clark1989cn2}
P.~Clark and T.~Niblett.
\newblock The cn2 induction algorithm.
\newblock {\em Machine learning}, 3(4):261--283, 1989.

\bibitem{cohen2016tensorlog}
W.~W. Cohen.
\newblock Tensorlog: A differentiable deductive database.
\newblock {\em arXiv preprint arXiv:1605.06523}, 2016.

\bibitem{Cui2017KBQALQ}
W.~Cui, Y.~Xiao, H.~Wang, Y.~Song, S.-w. Hwang, and W.~Wang.
\newblock Kbqa: Learning question answering over qa corpora and knowledge
  bases.
\newblock {\em Proc. VLDB Endow.}, 10(5):565--576, 2017.

\bibitem{cussens2001parameter}
J.~Cussens.
\newblock Parameter estimation in stochastic logic programs.
\newblock {\em Machine Learning}, 44(3):245--271, 2001.

\bibitem{dai2016cfo}
Z.~Dai, L.~Li, and W.~Xu.
\newblock Cfo: Conditional focused neural question answering with large-scale
  knowledge bases.
\newblock In {\em Proceedings of the 54th Annual Meeting of the Association for
  Computational Linguistics}, 2016.

\bibitem{Das2017MINERVA}
R.~Das, S.~Dhuliawala, M.~Zaheer, L.~Vilnis, I.~Durugkar, A.~Krishnamurthy,
  A.~Smola, and A.~McCallum.
\newblock Go for a walk and arrive at the answer: Reasoning over knowledge
  bases with reinforcement learning.
\newblock In {\em Proceedings of the 6th Workshop on Automated Knowledge Base
  Construction}, pages 1--18, 2018.

\bibitem{Das2017chain}
R.~Das, A.~Neelakantan, D.~Belanger, and A.~McCallum.
\newblock Chains of reasoning over entities, relations, and text using
  recurrent neural networks.
\newblock In {\em Proceedings of the 15th Conference of the European Chapter of
  the Association for Computational Linguistics}, pages 132--141, 2017.

\bibitem{de2011neural}
L.~de~Penning, A.~Garcez, L.~C. Lamb, and J.~Meyer.
\newblock A neural-symbolic cognitive agent for online learning and reasoning.
\newblock In {\em Proceedings of the Twenty-Second International Joint
  Conference on Artificial Intelligence}, volume~2, pages 1653--1658, 2011.

\bibitem{de2007problog}
L.~De~Raedt, A.~Kimmig, and H.~Toivonen.
\newblock Problog: A probabilistic prolog and its application in link
  discovery.
\newblock In {\em Proceedings of the 20th International Joint Conference on
  Artifical Intelligence}, pages 2462--2467. Hyderabad, 2007.

\bibitem{dettmers2018convolutional}
T.~Dettmers, P.~Minervini, P.~Stenetorp, and S.~Riedel.
\newblock Convolutional 2d knowledge graph embeddings.
\newblock In {\em Thirty-Second AAAI Conference on Artificial Intelligence},
  2018.

\bibitem{devlin2018bert}
J.~Devlin, M.-W. Chang, K.~Lee, and K.~Toutanova.
\newblock Bert: Pre-training of deep bidirectional transformers for language
  understanding.
\newblock {\em arXiv preprint arXiv:1810.04805}, 2018.

\bibitem{Dhingra2020DifferentiableRO}
B.~Dhingra, M.~Zaheer, V.~Balachandran, G.~Neubig, R.~Salakhutdinov, and
  W.~Cohen.
\newblock Differentiable reasoning over a virtual knowledge base.
\newblock {\em ArXiv}, abs/2002.10640, 2020.

\bibitem{ding2019cognitive}
M.~Ding, C.~Zhou, Q.~Chen, H.~Yang, and J.~Tang.
\newblock Cognitive graph for multi-hop reading comprehension at scale.
\newblock In {\em Proceedings of the 57th Annual Meeting of the Association for
  Computational Linguistics}, pages 2694--2703, 2019.

\bibitem{Du-2019}
Z.~Du, C.~Zhou, M.~Ding, H.~Yang, and J.~Tang.
\newblock Cognitive knowledge graph reasoning for one-shot relational learning.
\newblock In {\em https://arxiv.org/abs/1906.05489}, 2019.

\bibitem{dubey2016asknow}
M.~Dubey, S.~Dasgupta, A.~Sharma, K.~H{\"o}ffner, and J.~Lehmann.
\newblock Asknow: A framework for natural language query formalization in
  sparql.
\newblock In {\em The Semantic Web. Latest Advances and New Domains - 13th
  International Conference}, pages 300--316, 2016.

\bibitem{evans2018learning}
R.~Evans and E.~Grefenstette.
\newblock Learning explanatory rules from noisy data.
\newblock {\em Journal of Artificial Intelligence Research}, 61:1--64, 2018.

\bibitem{flach1994simply}
P.~A. Flach.
\newblock {\em Simply logical - intelligent reasoning by example}.
\newblock Wiley professional computing. Wiley, 1994.

\bibitem{fu2020survey}
B.~Fu, Y.~Qiu, C.~Tang, Y.~Li, H.~Yu, and J.~Sun.
\newblock A survey on complex question answering over knowledge base: Recent
  advances and challenges.
\newblock {\em ArXiv}, abs/2007.13069, 2020.

\bibitem{fu2019CPL}
C.~Fu, T.~Chen, M.~Qu, W.~Jin, and X.~Ren.
\newblock Collaborative policy learning for open knowledge graph reasoning.
\newblock In {\em Proceedings of the 2019 Conference on Empirical Methods in
  Natural Language Processing}, pages 2672--2681, 2019.

\bibitem{fuhr1995probabilistic}
N.~Fuhr.
\newblock Probabilistic datalog—a logic for powerful retrieval methods.
\newblock In {\em Proceedings of the 18th Annual International ACM SIGIR
  Conference on Research and Development in Information Retrieval}, pages
  282--290, 1995.

\bibitem{galarraga2015fast}
L.~Gal{\'a}rraga, C.~Teflioudi, K.~Hose, and F.~M. Suchanek.
\newblock Fast rule mining in ontological knowledge bases with amie.
\newblock {\em The VLDB Journal}, 24(6):707--730, 2015.

\bibitem{galarraga2013amie}
L.~A. Gal{\'a}rraga, C.~Teflioudi, K.~Hose, and F.~Suchanek.
\newblock Amie: association rule mining under incomplete evidence in
  ontological knowledge bases.
\newblock In {\em Proceedings of the 22nd International Conference on World
  Wide Web}, pages 413--422, 2013.

\bibitem{gallant1988connectionist}
S.~I. Gallant.
\newblock Connectionist expert systems.
\newblock {\em Communications of the ACM}, 31(2):152--169, 1988.

\bibitem{gallant1993neural}
S.~I. Gallant.
\newblock {\em Neural network learning and expert systems}.
\newblock MIT press, 1993.

\bibitem{garcez2019neural}
A.~d. Garcez, M.~Gori, L.~C. Lamb, L.~Serafini, M.~Spranger, and S.~N. Tran.
\newblock Neural-symbolic computing: An effective methodology for principled
  integration of machine learning and reasoning.
\newblock {\em arXiv preprint arXiv:1905.06088}, 2019.

\bibitem{garcez2012neural}
A.~S.~d. Garcez, K.~B. Broda, and D.~M. Gabbay.
\newblock {\em Neural-symbolic learning systems: foundations and applications}.
\newblock Springer Science \& Business Media, 2012.

\bibitem{gardner-etal-2013-improving}
M.~Gardner, P.~P. Talukdar, B.~Kisiel, and T.~Mitchell.
\newblock Improving learning and inference in a large knowledge-base using
  latent syntactic cues.
\newblock In {\em Proceedings of the 2013 Conference on Empirical Methods in
  Natural Language Processing}, pages 833--838. Association for Computational
  Linguistics, 2013.

\bibitem{golub2016character}
D.~Golub and X.~He.
\newblock Character-level question answering with attention.
\newblock In {\em Proceedings of the 2016 Conference on Empirical Methods in
  Natural Language Processing}, pages 1598--1607, 2016.

\bibitem{2015Approximation}
M.~R. Gormley, M.~Dredze, and J.~Eisner.
\newblock Approximation-aware dependency parsing by belief propagation.
\newblock {\em Transactions of the Association for Computational Linguistics
  (TACL)}, 3:489--501, 2015.

\bibitem{pmlr-v97-guo19c}
L.~Guo, Z.~Sun, and W.~Hu.
\newblock Learning to exploit long-term relational dependencies in knowledge
  graphs.
\newblock In {\em Proceedings of the 36th International Conference on Machine
  Learning}, pages 2505--2514, 2019.

\bibitem{guo2016kale}
S.~Guo, Q.~Wang, L.~Wang, B.~Wang, and L.~Guo.
\newblock Jointly embedding knowledge graphs and logical rules.
\newblock In {\em Proceedings of the 2016 Conference on Empirical Methods in
  Natural Language Processing}, pages 192--202, 2016.

\bibitem{guo2017ruge}
S.~Guo, Q.~Wang, L.~Wang, B.~Wang, and L.~Guo.
\newblock Knowledge graph embedding with iterative guidance from soft rules.
\newblock In {\em Proceedings of the Thirty-Second {AAAI} Conference on
  Artificial Intelligence}, pages 4816--4823, 2018.

\bibitem{Hamilton2018EmbeddingLQ}
W.~L. Hamilton, P.~Bajaj, M.~Zitnik, D.~Jurafsky, and J.~Leskovec.
\newblock Embedding logical queries on knowledge graphs.
\newblock In {\em Advances in Neural Information Processing Systems 31}, pages
  2030--2041, 2018.

\bibitem{haugeland1989artificial}
J.~Haugeland.
\newblock {\em Artificial intelligence: The very idea}.
\newblock MIT press, 1989.

\bibitem{Haveliwala2002TopicsensitiveP}
T.~H. Haveliwala.
\newblock Topic-sensitive pagerank.
\newblock In {\em Proceedings of the Eleventh International World Wide Web
  Conference}, pages 517--526, 2002.

\bibitem{he2020momentum}
K.~He, H.~Fan, Y.~Wu, S.~Xie, and R.~Girshick.
\newblock Momentum contrast for unsupervised visual representation learning.
\newblock In {\em Proceedings of the IEEE/CVF Conference on Computer Vision and
  Pattern Recognition}, pages 9729--9738, 2020.

\bibitem{he2016deep}
K.~He, X.~Zhang, S.~Ren, and J.~Sun.
\newblock Deep residual learning for image recognition.
\newblock In {\em Proceedings of the IEEE Conference on Computer Vision and
  Pattern Recognition}, pages 770--778, 2016.

\bibitem{hinton2006fast}
G.~E. Hinton, S.~Osindero, and Y.-W. Teh.
\newblock A fast learning algorithm for deep belief nets.
\newblock {\em Neural computation}, 18(7):1527--1554, 2006.

\bibitem{Frank1927}
F.~L. Hitchcock.
\newblock The expression of a tensor or a polyadic as a sum of products.
\newblock {\em Studies in Applied Mathematics}, 6:164–189, 1927.

\bibitem{ho2018rule}
V.~T. Ho, D.~Stepanova, M.~H. Gad-Elrab, E.~Kharlamov, and G.~Weikum.
\newblock Rule learning from knowledge graphs guided by embedding models.
\newblock In {\em Proceedings of the 17th International Semantic Web
  Conference}, pages 72--90. Springer, 2018.

\bibitem{Hu2018ASF}
S.~Hu, L.~Zou, and X.~Zhang.
\newblock A state-transition framework to answer complex questions over
  knowledge base.
\newblock In {\em Proceedings of the 2018 Conference on Empirical Methods in
  Natural Language Processing}, pages 2098--2108, 2018.

\bibitem{hu2020gpt-gnn}
Z.~Hu, Y.~Dong, K.~Wang, K.-W. Chang, and Y.~Sun.
\newblock Gpt-gnn: Generative pre-training of graph neural networks.
\newblock In {\em Proceedings of the 26th ACM SIGKDD International Conference
  on Knowledge Discovery and Data Mining}, pages 1857--1867, 2020.

\bibitem{huang2019knowledge}
X.~Huang, J.~Zhang, D.~Li, and P.~Li.
\newblock Knowledge graph embedding based question answering.
\newblock In {\em Proceedings of the Twelfth ACM International Conference on
  Web Search and Data Mining}, pages 105--113, 2019.

\bibitem{Petr2020}
P.~H´ajek.
\newblock The metamathematics of fuzzy logic.
\newblock 1998.

\bibitem{jha2012probabilistic}
A.~Jha and D.~Suciu.
\newblock Probabilistic databases with markoviews.
\newblock {\em Proceedings of the VLDB Endowment}, 5(11), 2012.

\bibitem{ji-2015}
G.~Ji, S.~He, L.~Xu, K.~Liu, and J.~Zhao.
\newblock Knowledge graph embedding via dynamic mapping matrix.
\newblock In {\em Proceedings of the 53rd Annual Meeting of the Association for
  Computational Linguistics and the 7th International Joint Conference on
  Natural Language Processing (Volume 1: Long Papers)}, pages 687--696, 2015.

\bibitem{ji2020survey}
S.~Ji, S.~Pan, E.~Cambria, P.~Marttinen, and P.~S. Yu.
\newblock A survey on knowledge graphs: Representation, acquisition and
  applications.
\newblock In {\em arXiv preprint arXiv:2002.00388}, 2020.

\bibitem{jiang2019adaptive}
X.~Jiang, Q.~Wang, and B.~Wang.
\newblock Adaptive convolution for multi-relational learning.
\newblock In {\em Proceedings of the 2019 Conference of the North American
  Chapter of the Association for Computational Linguistics: Human Language
  Technologies, Volume 1 (Long and Short Papers)}, pages 978--987, 2019.

\bibitem{kate2005learning}
R.~J. Kate, Y.~W. Wong, and R.~J. Mooney.
\newblock Learning to transform natural to formal languages.
\newblock In {\em Proceedings of the Twentieth National Conference on
  Artificial Intelligence and the Seventeenth Innovative Applications of
  Artificial Intelligence Conference}, pages 1062--1068, 2005.

\bibitem{NIPS2018_7682}
S.~M. Kazemi and D.~Poole.
\newblock Simple embedding for link prediction in knowledge graphs.
\newblock In {\em Proceedings of the 32nd International Conference on Neural
  Information Processing Systems}, pages 4284--4295. 2018.

\bibitem{Diederik-2013}
D.~P. Kingma and M.~Welling.
\newblock Auto-encoding variational bayes.
\newblock In {\em Proceedings of the 2nd International Conference on Learning
  Representations}, 2014.

\bibitem{kipf2016semi}
T.~N. Kipf and M.~Welling.
\newblock Semi-supervised classification with graph convolutional networks.
\newblock {\em arXiv preprint arXiv:1609.02907}, 2016.

\bibitem{Klein2003AccurateUP}
D.~Klein and C.~D. Manning.
\newblock Accurate unlexicalized parsing.
\newblock In {\em Proceedings of the 41st Annual Meeting on Association for
  Computational Linguistics - Volume 1}, page 423–430, 2003.

\bibitem{krizhevsky2017imagenet}
A.~Krizhevsky, I.~Sutskever, and E.~G. Hinton.
\newblock Imagenet classification with deep convolutional neural networks.
\newblock {\em Commun. ACM}, pages 84--90, 2017.

\bibitem{kwiatkowksi2010inducing}
T.~Kwiatkowksi, L.~Zettlemoyer, S.~Goldwater, and M.~Steedman.
\newblock Inducing probabilistic ccg grammars from logical form with
  higher-order unification.
\newblock In {\em Proceedings of the 2010 Conference on Empirical Methods in
  Natural Language Processing}, pages 1223--1233, 2010.

\bibitem{lamb2020graph}
L.~Lamb, A.~Garcez, M.~Gori, M.~Prates, P.~Avelar, and M.~Vardi.
\newblock Graph neural networks meet neural-symbolic computing: A survey and
  perspective.
\newblock {\em arXiv preprint arXiv:2003.00330}, 2020.

\bibitem{lan-jiang-2020-query}
Y.~Lan and J.~Jiang.
\newblock Query graph generation for answering multi-hop complex questions from
  knowledge bases.
\newblock In {\em Proceedings of the 58th Annual Meeting of the Association for
  Computational Linguistics}, pages 969--974, July 2020.

\bibitem{lao2010relational}
N.~Lao and W.~W. Cohen.
\newblock Relational retrieval using a combination of path-constrained random
  walks.
\newblock {\em Machine learning}, 81(1):53--67, 2010.

\bibitem{lao-etal-2012-reading}
N.~Lao, A.~Subramanya, F.~Pereira, and W.~W. Cohen.
\newblock Reading the web with learned syntactic-semantic inference rules.
\newblock In {\em Proceedings of the 2012 Joint Conference on Empirical Methods
  in Natural Language Processing and Computational Natural Language Learning},
  pages 1017--1026, 2012.

\bibitem{Lehmann2015}
J.~Lehmann, R.~Isele, M.~Jakob, A.~Jentzsch, D.~Kontokostas, P.~N. Mendes,
  S.~Hellmann, M.~Morsey, P.~van Kleef, S.~Auer, and C.~Bizer.
\newblock Dbpedia: A large-scale, multilingual knowledge base extracted from
  wikipedia.
\newblock 6:167--195, 2015.

\bibitem{fei2006one}
F.~Li, R.~Fergus, and P.~Perona.
\newblock One-shot learning of object categories.
\newblock {\em IEEE transactions on pattern analysis and machine intelligence},
  28(4):594--611, 2006.

\bibitem{Liang2013}
P.~Liang.
\newblock Lambda dependency-based compositional semantics.
\newblock {\em Technical report}, 2013.

\bibitem{Lin2018multi}
X.~V. Lin, R.~Socher, and C.~Xiong.
\newblock Multi-hop knowledge graph reasoning with reward shaping.
\newblock In {\em Proceedings of the 2018 Conference on Empirical Methods in
  Natural Language Processing}, pages 3243–--3253, 2018.

\bibitem{lin2015learning}
Y.~Lin, Z.~Liu, M.~Sun, Y.~Liu, and X.~Zhu.
\newblock Learning entity and relation embeddings for knowledge graph
  completion.
\newblock In {\em Proceedings of the Twenty-Ninth {AAAI} Conference on
  Artificial Intelligence}, pages 2181--2187, 2015.

\bibitem{pmlr-v70-liu17d}
H.~Liu, Y.~Wu, and Y.~Yang.
\newblock Analogical inference for multi-relational embeddings.
\newblock In {\em Proceedings of the 34th International Conference on Machine
  Learning}, volume~70, pages 2168--2178, 2017.

\bibitem{liu2020}
X.~Liu, F.~Zhang, Z.~Hou, Z.~Wang, L.~Min, J.~Zhang, and J.~Tang.
\newblock Self-supervised learning: Generative or contrastive.
\newblock In {\em https://arxiv.org/abs/2006.08218}, 2020.

\bibitem{Liu-2019}
Y.~Liu, M.~Ott, N.~Goyal, J.~Du, M.~Joshi, D.~Chen, O.~Levy, M.~Lewis,
  L.~Zettlemoyer, and V.~Stoyanov.
\newblock Roberta: A robustly optimized bert pretraining approach.
\newblock In {\em arXiv preprint arXiv:1907.11692}, 2019.

\bibitem{lukovnikov2017neural}
D.~Lukovnikov, A.~Fischer, J.~Lehmann, and S.~Auer.
\newblock Neural network-based question answering over knowledge graphs on word
  and character level.
\newblock In {\em Proceedings of the 26th International Conference on World
  Wide Web}, pages 1211--1220, 2017.

\bibitem{Mazzeo2016AnsweringCN}
G.~M. Mazzeo and C.~Zaniolo.
\newblock Answering controlled natural language questions on rdf knowledge
  bases.
\newblock In {\em Proceedings of the 19th International Conference on Extending
  Database Technology}, pages 608--611, 2016.

\bibitem{meilicke2020AnyBURL}
C.~Meilicke, M.~W. Chekol, M.~Fink, and H.~Stuckenschmidt.
\newblock Reinforced anytime bottom up rule learning for knowledge graph
  completion.
\newblock {\em arXiv preprint arXiv:2004.04412}, 2020.

\bibitem{michalski1986multi}
R.~S. Michalski, I.~Mozetic, J.~Hong, and N.~Lavrac.
\newblock The multi-purpose incremental learning system aq15 and its testing
  application to three medical domains.
\newblock In {\em Proceedings of the 5th National Conference on Artificial
  Intelligence}, pages 1041--1045, 1986.

\bibitem{minsky1969introduction}
M.~Minsky and S.~Papert.
\newblock An introduction to computational geometry.
\newblock {\em Cambridge tiass., HIT}, 1969.

\bibitem{mohammed2017strong}
S.~Mohammed, P.~Shi, and J.~Lin.
\newblock Strong baselines for simple question answering over knowledge graphs
  with and without neural networks.
\newblock In {\em Proceedings of the 2018 Conference of the North American
  Chapter of the Association for Computational Linguistics: Human Language
  Technologies}, pages 291--296, 2018.

\bibitem{Deepak-ACL2019}
D.~Nathani, J.~Chauhan, C.~Sharma, and M.~Kaul.
\newblock Learning attention-based embeddings for relation prediction in
  knowledge graphs.
\newblock In {\em Proceedings of the 57th Annual Meeting of the Association for
  Computational Linguistics}, page 4710–4723, 2019.

\bibitem{Neal1998}
R.~M. Neal and G.~E. Hinton.
\newblock A view of the em algorithm that justifies incremental, sparse, and
  other variants.
\newblock {\em Learning in graphical models}, page 355–368, 1999.

\bibitem{Neelakantan2015Compositional}
A.~Neelakantan, B.~Roth, and A.~McCallum.
\newblock Compositional vector space models for knowledge base inference.
\newblock In {\em Proceedings of the 53rd Annual Meeting of the Association for
  Computational Linguistics and the 7th International Joint Conference on
  Natural Language Processing of the Asian Federation of Natural Language
  Processing}, volume~1, pages 156--166, 2015.

\bibitem{nguyen2018novel}
T.~D. Nguyen, D.~Q. Nguyen, D.~Phung, et~al.
\newblock A novel embedding model for knowledge base completion based on
  convolutional neural network.
\newblock In {\em Proceedings of the 2018 Conference of the North American
  Chapter of the Association for Computational Linguistics: Human Language
  Technologies, Volume 2 (Short Papers)}, pages 327--333, 2018.

\bibitem{NickelAAAI16}
M.~Nickel, L.~Rosasco, and T.~Poggio.
\newblock Holographic embeddings of knowledge graphs.
\newblock In {\em Proceedings of the Thirtieth AAAI Conference on Artificial
  Intelligence}, page 1955–1961, 2016.

\bibitem{NickelICML11}
M.~Nickel, V.~Tresp, and H.~P. Kriegel.
\newblock A three-way model for collective learning on multi-relational data.
\newblock In {\em Proceedings of the 28th International Conference on
  International Conference on Machine Learning}, page 809–816, 2011.

\bibitem{nickel2011three}
M.~Nickel, V.~Tresp, and H.-P. Kriegel.
\newblock A three-way model for collective learning on multi-relational data.
\newblock In {\em Proceedings of the 28th International Conference on Machine
  Learning}, volume~11, pages 809--816, 2011.

\bibitem{Niu2020JointSA}
G.~Niu, B.~Li, Y.~Zhang, Y.~Sheng, C.~Shi, J.~Li, and S.~Pu.
\newblock Joint semantics and data-driven path representation for knowledge
  graph inference.
\newblock {\em ArXiv}, abs/2010.02602, 2020.

\bibitem{omran2018scalable}
P.~G. Omran, K.~Wang, and Z.~Wang.
\newblock Scalable rule learning via learning representation.
\newblock In {\em Proceedings of the Twenty-Seventh International Joint
  Conference on Artificial Intelligence}, pages 2149--2155, 2018.

\bibitem{page1999pagerank}
L.~Page, S.~Brin, R.~Motwani, and T.~Winograd.
\newblock The pagerank citation ranking: Bringing order to the web.
\newblock Technical report, Stanford InfoLab, 1999.

\bibitem{pearl2014probabilistic}
J.~Pearl.
\newblock {\em Probabilistic reasoning in intelligent systems: networks of
  plausible inference}.
\newblock Elsevier, 2014.

\bibitem{TunstallPedoe2010TrueKO}
W.~T. Pedoe.
\newblock True knowledge: Open-domain question answering using structured
  knowledge and inference.
\newblock {\em AI Mag.}, 31(3):80--92, 2010.

\bibitem{plotkin1970note}
G.~D. Plotkin.
\newblock A note on inductive generalization.
\newblock {\em Machine Intelligence}, 5(1):153--163, 1970.

\bibitem{plotkin1971further}
G.~D. Plotkin.
\newblock A further note on inductive generalization.
\newblock {\em Machine Intelligence}, 6(101-124):248, 1971.

\bibitem{qian2018translating}
W.~Qian, C.~Fu, Y.~Zhu, D.~Cai, and X.~He.
\newblock Translating embeddings for knowledge graph completion with relation
  attention mechanism.
\newblock In {\em Proceedings of the Twenty-Seventh International Joint
  Conference on Artificial Intelligence}, pages 4286--4292, 2018.

\bibitem{qiu2020gcc}
J.~Qiu, Q.~Chen, Y.~Dong, J.~Zhang, H.~Yang, M.~Ding, K.~Wang, and J.~Tang.
\newblock Gcc: Graph contrastive coding for graph neural network pre-training.
\newblock In {\em Proceedings of the 26th ACM SIGKDD International Conference
  on Knowledge Discovery and Data Mining}, 2020.

\bibitem{qiu2020stepwise}
Y.~Qiu, Y.~Wang, X.~Jin, and K.~Zhang.
\newblock Stepwise reasoning for multi-relation question answering over
  knowledge graph with weak supervision.
\newblock In {\em Proceedings of The Thirteenth {ACM} International Conference
  on Web Search and Data Mining}, pages 474--482, 2020.

\bibitem{qu2019policnet}
M.~Qu and J.~Tang.
\newblock Probabilistic logic neural networks for reasoning.
\newblock In {\em Advances in Neural Information Processing Systems 32: Annual
  Conference on Neural Information Processing Systems 2019}, pages 7712--7722,
  2019.

\bibitem{ren2019query2box}
H.~Ren, W.~Hu, and J.~Leskovec.
\newblock Query2box: Reasoning over knowledge graphs in vector space using box
  embeddings.
\newblock In {\em Proceedings of the 8th International Conference on Learning
  Representations}, 2020.

\bibitem{richardson2006markov}
M.~Richardson and P.~Domingos.
\newblock Markov logic networks.
\newblock {\em Machine learning}, 62(1-2):107--136, 2006.

\bibitem{riedel2010modeling}
S.~Riedel, L.~Yao, and A.~McCallum.
\newblock Modeling relations and their mentions without labeled text.
\newblock In {\em Joint European Conference on Machine Learning and Knowledge
  Discovery in Databases}, pages 148--163. Springer, 2010.

\bibitem{rosenblatt1958perceptron}
F.~Rosenblatt.
\newblock The perceptron: a probabilistic model for information storage and
  organization in the brain.
\newblock {\em Psychological review}, 65(6):386, 1958.

\bibitem{rossi2020knowledge}
A.~Rossi, D.~Firmani, A.~Matinata, P.~Merialdo, and D.~Barbosa.
\newblock Knowledge graph embedding for link prediction: A comparative
  analysis.
\newblock In {\em arXiv preprint arXiv:2002.00819}, 2020.

\bibitem{rumelhart1986learning}
D.~E. Rumelhart, G.~E. Hinton, and R.~J. Williams.
\newblock Learning representations by back-propagating errors.
\newblock {\em nature}, 323(6088):533--536, 1986.

\bibitem{Sara-NIPS-2017}
S.~Sabour, N.~Frosst, and G.~E. Hinton.
\newblock Dynamic routing between capsules.
\newblock In {\em Proceedings of the 31st International Conference on Neural
  Information Processing Systems}, page 3859–3869, 2017.

\bibitem{Saxena2020ImprovingMQ}
A.~Saxena, A.~Tripathi, and P.~Talukdar.
\newblock Improving multi-hop question answering over knowledge graphs using
  knowledge base embeddings.
\newblock In {\em Proceedings of the 58th Annual Meeting of the Association for
  Computational Linguistics}, pages 4498--4507, 2020.

\bibitem{schlichtkrull2018modeling}
M.~Schlichtkrull, T.~N. Kipf, P.~Bloem, R.~Van Den~Berg, I.~Titov, and
  M.~Welling.
\newblock Modeling relational data with graph convolutional networks.
\newblock In {\em European Semantic Web Conference}, pages 593--607, 2018.

\bibitem{schmitz2012open}
M.~Schmitz, S.~Soderland, R.~Bart, O.~Etzioni, et~al.
\newblock Open language learning for information extraction.
\newblock In {\em Proceedings of the 2012 Joint Conference on Empirical Methods
  in Natural Language Processing and Computational Natural Language Learning},
  pages 523--534, 2012.

\bibitem{ShangAAAI2019}
C.~Shang, Y.~Tang, J.~Huang, J.~Bi, X.~He, and B.~Zhou.
\newblock End-to-end structure-aware convolutional networks for knowledge base
  completion.
\newblock In {\em Proceedings of the Thirty-third AAAI Conference on Artificial
  Intelligence}, 2019.

\bibitem{shavlik1991symbolic}
J.~W. Shavlik, R.~J. Mooney, and G.~G. Towell.
\newblock Symbolic and neural learning algorithms: An experimental comparison.
\newblock {\em Machine learning}, 6(2):111--143, 1991.

\bibitem{shen2018mwalk}
Y.~Shen, J.~Chen, P.-S. Huang, Y.~Guo, and J.~Gao.
\newblock M-walk: Learning to walk in graph with monte carlo tree search.
\newblock In {\em Proceedings of 6th International Conference on Learning
  Representations}, pages 6686--6797, 2018.

\bibitem{Suchanek2007}
F.~M. Suchanek, G.~Kasneci, and G.~Weikum.
\newblock Yago: A core of semantic knowledge.
\newblock In {\em Proceedings of the 16th International Conference World Wide
  Web}, pages 697--706, 2007.

\bibitem{Sun2020FaithfulEF}
H.~Sun, A.~O. Arnold, T.~Bedrax-Weiss, F.~Pereira, and W.~Cohen.
\newblock Faithful embeddings for knowledge base queries.
\newblock In {\em Advances in Neural Information Processing Systems}, 2020.

\bibitem{sun-etal-2018-open}
H.~Sun, B.~Dhingra, M.~Zaheer, K.~Mazaitis, R.~Salakhutdinov, and W.~Cohen.
\newblock Open domain question answering using early fusion of knowledge bases
  and text.
\newblock In {\em Proceedings of the 2018 Conference on Empirical Methods in
  Natural Language Processing}, pages 4231--4242, 2018.

\bibitem{SunACL2019}
H.~Sun, T.~B. Weiss, and W.~W. Cohen.
\newblock Pullnet: Open domain question answering with iterative retrieval on
  knowledge bases and text.
\newblock In {\em Proceedings of the 2019 Conference on Empirical Methods in
  Natural Language Processing}, pages 2380--2390, 2019.

\bibitem{sun2018rotate}
Z.~Sun, Z.~Deng, J.~Nie, and J.~Tang.
\newblock Rotate: Knowledge graph embedding by relational rotation in complex
  space.
\newblock In {\em Proceedings of the 7th International Conference on Learning
  Representations}, 2019.

\bibitem{teru2019inductive}
K.~K. Teru, E.~Denis, and W.~L. Hamilton.
\newblock Inductive relation prediction by subgraph reasoning.
\newblock {\em arXiv}, 2019.

\bibitem{toutanova2015representing}
K.~Toutanova, D.~Chen, P.~Pantel, H.~Poon, P.~Choudhury, and M.~Gamon.
\newblock Representing text for joint embedding of text and knowledge bases.
\newblock In {\em Proceedings of the 2015 Conference on Empirical Methods in
  Natural Language Processing}, pages 1499--1509, 2015.

\bibitem{towell1994knowledge}
G.~G. Towell and J.~W. Shavlik.
\newblock Knowledge-based artificial neural networks.
\newblock {\em Artificial intelligence}, 70(1-2):119--165, 1994.

\bibitem{towell1994refining}
G.~G. Towell and J.~W. Shavlik.
\newblock Refining symbolic knowledge using neural networks.
\newblock {\em Machine learning: A multistrategy approach}, 4:405--429, 1994.

\bibitem{pmlr-v48-trouillon16}
T.~Trouillon, J.~Welbl, S.~Riedel, E.~Gaussier, and G.~Bouchard.
\newblock Complex embeddings for simple link prediction.
\newblock In {\em Proceedings of the 33rd International Conference on
  International Conference on Machine Learning}, pages 2071--2080, 2016.

\bibitem{Trouillon2016ComplexEF}
T.~Trouillon, J.~Welbl, S.~Riedel, {\'E}.~Gaussier, and G.~Bouchard.
\newblock Complex embeddings for simple link prediction.
\newblock {\em Proceedings of the 33nd International Conference on Machine
  Learning}, 48:2071--2080, 2016.

\bibitem{ture2016no}
F.~T{\"{u}}re and O.~Jojic.
\newblock No need to pay attention: Simple recurrent neural networks work!
\newblock In {\em Proceedings of the 2017 Conference on Empirical Methods in
  Natural Language Processing}, pages 2866--2872. Association for Computational
  Linguistics, 2017.

\bibitem{Unger2012TemplatebasedQA}
C.~Unger, L.~B\"{u}hmann, J.~Lehmann, A.-C. Ngonga~Ngomo, D.~Gerber, and
  P.~Cimiano.
\newblock Template-based question answering over rdf data.
\newblock In {\em Proceedings of the 21st International Conference on World
  Wide Web}, page 639–648, 2012.

\bibitem{Unger2011PythiaCM}
C.~Unger and P.~Cimiano.
\newblock Pythia: Compositional meaning construction for ontology-based
  question answering on the semantic web.
\newblock In {\em Proceedings of the 16th International Conference on Natural
  Language Processing and Information Systems}, page 153–160, 2011.

\bibitem{vashishth2020interacte}
S.~Vashishth, S.~Sanyal, V.~Nitin, N.~Agrawal, and P.~P. Talukdar.
\newblock Interacte: Improving convolution-based knowledge graph embeddings by
  increasing feature interactions.
\newblock In {\em AAAI}, pages 3009--3016, 2020.

\bibitem{vashishth2019composition}
S.~Vashishth, S.~Sanyal, V.~Nitin, and P.~Talukdar.
\newblock Composition-based multi-relational graph convolutional networks.
\newblock {\em Proceedings of the 8th International Conference on Learning
  Representations}, 2020.

\bibitem{velivckovic2017graph}
P.~Veli{\v{c}}kovi{\'c}, G.~Cucurull, A.~Casanova, A.~Romero, P.~Lio, and
  Y.~Bengio.
\newblock Graph attention networks.
\newblock {\em arXiv preprint arXiv:1710.10903}, 2017.

\bibitem{Dai-2019}
T.~Vu, T.~D. Nguyen, D.~Q. Nguyen, D.~Phung, et~al.
\newblock A capsule network-based embedding model for knowledge graph
  completion and search personalization.
\newblock In {\em Proceedings of the 2019 Conference of the North American
  Chapter of the Association for Computational Linguistics: Human Language
  Technologies, Volume 1 (Long and Short Papers)}, pages 2180--2189, 2019.

\bibitem{wang2019logic}
P.~Wang, D.~Dou, F.~Wu, N.~d. Silva, and L.~Jin.
\newblock Logic rules powered knowledge graph embedding.
\newblock {\em ArXiv}, abs/1903.03772, 2019.

\bibitem{wang2019differentiable}
P.~Wang, D.~Stepanova, C.~Domokos, and J.~Z. Kolter.
\newblock Differentiable learning of numerical rules in knowledge graphs.
\newblock In {\em Proceedings of the 8th International Conference on Learning
  Representations}, 2020.

\bibitem{Wang2017}
Q.~Wang, Z.~Mao, B.~Wang, and L.~Guo.
\newblock Knowledge graph embedding: A survey of approaches and applications.
\newblock {\em IEEE Transactions on Knowledge and Data Engineering},
  29:2724--2743, 2017.

\bibitem{wang2013programming}
W.~Y. Wang, K.~Mazaitis, and W.~W. Cohen.
\newblock Programming with personalized pagerank: a locally groundable
  first-order probabilistic logic.
\newblock In {\em Proceedings of the 22nd ACM International Conference on
  Information \& Knowledge Management}, pages 2129--2138, 2013.

\bibitem{Wang-ijcai2019}
Z.~Wang, Z.~Ren, C.~He, P.~Zhang, and Y.~Hu.
\newblock Robust embedding with multi-level structures for link prediction.
\newblock In {\em Proceedings of the Twenty-Eighth International Joint
  Conference on Artificial Intelligence, {IJCAI-19}}, pages 5240--5246, 2019.

\bibitem{Wang14}
Z.~Wang, J.~Zhang, J.~Feng, and Z.~Chen.
\newblock Knowledge graph embedding by translating on hyperplanes.
\newblock In {\em Proceedings of the Twenty-Eighth AAAI Conference on
  Artificial Intelligence}, page 1112–1119, 2014.

\bibitem{wong2007learning}
Y.~W. Wong and R.~Mooney.
\newblock Learning synchronous grammars for semantic parsing with lambda
  calculus.
\newblock In {\em Proceedings of the 45th Annual Meeting of the Association of
  Computational Linguistics}, pages 960--967, 2007.

\bibitem{Wu2019}
P.~Wu, X.~Zhang, and Z.~Feng.
\newblock A survey of question answering over knowledge base.
\newblock In {\em Knowledge Graph and Semantic Computing: Knowledge Computing
  and Language Understanding}, pages 86--97, 2019.

\bibitem{xiao2015transg}
H.~Xiao, M.~Huang, Y.~Hao, and X.~Zhu.
\newblock Transg: A generative mixture model for knowledge graph embedding.
\newblock In {\em Proceedings of the 54th Annual Meeting of the Association for
  Computational Linguistics}, pages 2316--2325, 2016.

\bibitem{Xu2020DPMPN}
X.~Xiaoran, F.~Wei, J.~Yunsheng, X.~Xiaohui, S.~Zhiqing, and D.~Zhi-Hong.
\newblock Dynamically pruned message passing networks for large-scale knowledge
  graph reasoning.
\newblock In {\em In Eighth International Conference on Learning
  Representations}, 2020.

\bibitem{Xiong2017deeppath}
W.~Xiong, T.~Hoang, and W.~Y. Wang.
\newblock Deeppath: A reinforcement learning method for knowledge graph
  reasoning.
\newblock In {\em Proceedings of the 2017 Conference on Empirical Methods in
  Natural Language Processing}, pages 564--573, 2017.

\bibitem{xiong2018one}
W.~Xiong, M.~Yu, S.~Chang, X.~Guo, and W.~Y. Wang.
\newblock One-shot relational learning for knowledge graphs.
\newblock In {\em Proceedings of the 2018 Conference on Empirical Methods in
  Natural Language Processing}, pages 1980--1990, 2018.

\bibitem{xu2019relation}
C.~Xu and R.~Li.
\newblock Relation embedding with dihedral group in knowledge graph.
\newblock In {\em Proceedings of the 57th Conference of the Association for
  Computational Linguistics}, pages 263--272, 2019.

\bibitem{xu2014answering}
K.~Xu, S.~Zhang, Y.~Feng, and D.~Zhao.
\newblock Answering natural language questions via phrasal semantic parsing.
\newblock In {\em CCF International Conference on Natural Language Processing
  and Chinese Computing}, pages 333--344. Springer, 2014.

\bibitem{yahya2012natural}
M.~Yahya, K.~Berberich, S.~Elbassuoni, M.~Ramanath, V.~Tresp, and G.~Weikum.
\newblock Natural language questions for the web of data.
\newblock In {\em Proceedings of the 2012 Joint Conference on Empirical Methods
  in Natural Language Processing and Computational Natural Language Learning},
  pages 379--390, 2012.

\bibitem{Yahya2012NaturalLQ}
M.~Yahya, K.~Berberich, S.~Elbassuoni, M.~Ramanath, V.~Tresp, and G.~Weikum.
\newblock Natural language questions for the web of data.
\newblock In {\em Proceedings of the 2012 Joint Conference on Empirical Methods
  in Natural Language Processing and Computational Natural Language Learning},
  page 379–390, 2012.

\bibitem{yang2014embedding}
B.~Yang, W.-t. Yih, X.~He, J.~Gao, and L.~Deng.
\newblock Embedding entities and relations for learning and inference in
  knowledge bases.
\newblock In {\em Proceedings of the 3rd International Conference on Learning
  Representations}, 2015.

\bibitem{yang2017differentiable}
F.~Yang, Z.~Yang, and W.~W. Cohen.
\newblock Differentiable learning of logical rules for knowledge base
  reasoning.
\newblock In {\em Advances in Neural Information Processing Systems 30}, pages
  2319--2328, 2017.

\bibitem{yang-chang-2015-mart}
Y.~Yang and M.~Chang.
\newblock S-{MART}: Novel tree-based structured learning algorithms applied to
  tweet entity linking.
\newblock In {\em Proceedings of the 53rd Annual Meeting of the Association for
  Computational Linguistics and the 7th International Joint Conference on
  Natural Language Processing (Volume 1: Long Papers)}, pages 504--513, 2015.

\bibitem{yang2019learn}
Y.~Yang and L.~Song.
\newblock Learn to explain efficiently via neural logic inductive learning.
\newblock In {\em Proceedings of the 8th International Conference on Learning
  Representations}, 2020.

\bibitem{ye2019vectorized}
R.~Ye, X.~Li, Y.~Fang, H.~Zang, and M.~Wang.
\newblock A vectorized relational graph convolutional network for
  multi-relational network alignment.
\newblock In {\em Proceedings of the Twenty-Eighth International Joint
  Conference on Artificial Intelligence}, pages 4135--4141, 2019.

\bibitem{Yih2014}
W.~Yih, X.~He, and C.~Meek.
\newblock Semantic parsing for single-relation question answering.
\newblock In {\em Proceedings of the 52nd Annual Meeting of the Association for
  Computational Linguistics}, pages 643--648, 2014.

\bibitem{yin2016simple}
W.~Yin, M.~Yu, B.~Xiang, B.~Zhou, and H.~Sch{\"u}tze.
\newblock Simple question answering by attentive convolutional neural network.
\newblock In {\em Proceedings of the 26th International Conference on
  Computational Linguistics}, pages 1746--1756, 2016.

\bibitem{you2020graph}
Y.~You, T.~Chen, Y.~Sui, T.~Chen, Z.~Wang, and Y.~Shen.
\newblock Graph contrastive learning with augmentations.
\newblock {\em Advances in Neural Information Processing Systems}, 33, 2020.

\bibitem{zettlemoyer2009learning}
L.~S. Zettlemoyer and M.~Collins.
\newblock Learning context-dependent mappings from sentences to logical form.
\newblock In {\em Proceedings of the 47th Annual Meeting of the Association for
  Computational Linguistics}, pages 976--984, 2009.

\bibitem{zhang2019iere}
W.~Zhang, B.~Paudel, L.~Wang, J.~Chen, H.~Zhu, W.~Zhang, A.~Bernstein, and
  H.~Chen.
\newblock Iteratively learning embeddings and rules for knowledge graph
  reasoning.
\newblock In {\em Proceedings of The World Wide Web Conference, 2019}, pages
  2366--2377, 2019.

\bibitem{zhang2018variational}
Y.~Zhang, H.~Dai, Z.~Kozareva, A.~J. Smola, and L.~Song.
\newblock Variational reasoning for question answering with knowledge graph.
\newblock In {\em Proceedings of the Thirty-Second {AAAI} Conference on
  Artificial Intelligence}, pages 6069--6076, 2018.

\bibitem{Zheng2018QuestionAO}
W.~Zheng, J.~X. Yu, L.~Zou, and H.~Cheng.
\newblock Question answering over knowledge graphs: Question understanding via
  template decomposition.
\newblock {\em Proceedings of the VLDB Endowment}, 11(11):1373--1386, 2018.

\bibitem{Zheng2015HowTB}
W.~Zheng, L.~Zou, X.~Lian, J.~X. Yu, S.~Song, and D.~Zhao.
\newblock How to build templates for rdf question/answering: An uncertain graph
  similarity join approach.
\newblock In {\em Proceedings of the 2015 {ACM} {SIGMOD} International
  Conference on Management of Data}, pages 1809--1824, 2015.

\bibitem{zhou2018an}
M.~Zhou, M.~Huang, and X.~Zhu.
\newblock An interpretable reasoning network for multi-relation question
  answering.
\newblock In {\em Proceedings of the 27th International Conference on
  Computational Linguistics}, pages 2010--2022, 2018.

\bibitem{zou2014natural}
L.~Zou, R.~Huang, H.~Wang, J.~X. Yu, W.~He, and D.~Zhao.
\newblock Natural language question answering over rdf: a graph data driven
  approach.
\newblock In {\em Proceedings of the 2014 ACM SIGMOD International Conference
  on Management of Data}, pages 313--324, 2014.

\bibitem{Zou2014NaturalLQ}
L.~Zou, R.~Huang, H.~Wang, J.~X. Yu, W.~He, and D.~Zhao.
\newblock Natural language question answering over rdf: a graph data driven
  approach.
\newblock In {\em Proceedings of the 2014 {ACM} {SIGMOD} International
  Conference on Management of Data}, pages 313--324, 2014.

\end{thebibliography}

%\input{appendix.tex}


%\vfill


\end{document}